\documentclass[preprint,10pt,3p]{elsarticle}
\usepackage{graphics} 
\usepackage{epsfig} 
\usepackage{times} 
\usepackage{amsmath} 
\usepackage{amssymb}  
\usepackage{epstopdf}
\usepackage{caption}
\usepackage{verbatim}
\usepackage{cases}
\allowdisplaybreaks[4]
\usepackage{enumerate}
\usepackage{booktabs}
\usepackage{balance}
\usepackage{mathbbol}

\usepackage{amsmath}
\usepackage[linesnumbered,ruled,vlined]{algorithm2e}
\usepackage{algpseudocode}

\usepackage{mathrsfs}
\usepackage{threeparttable}
\usepackage[draft]{todonotes}
\usepackage{makecell}

\usepackage{float}

\usepackage{multirow}
\usepackage[caption=false,font=normalsize,labelfont=sf,textfont=sf]{subfig}
\usepackage{textcomp}
\usepackage{stfloats}

\usepackage{wrapfig}
\usepackage{flushend} 

\usepackage{underscore}

\usepackage{enumitem}

\usepackage{lineno}

\usepackage[colorlinks=true,      
linkcolor=black,      
citecolor=black,      
filecolor=black,      
urlcolor=blue]{hyperref}

\newdefinition{remark}{Remark}

\def\x{{\bf x}}

\def\0{{\bf 0}}
\def\1{{\bf 1}}

 \biboptions{authoryear}

\begin{document}

	\begin{frontmatter}
		\title{HAIM-DRL: Enhanced Human-in-the-loop Reinforcement Learning \\ for Safe and Efficient Autonomous Driving}
		\author{Zilin Huang}
		\ead{zilin.huang@wisc.edu}
		\author{Zihao Sheng}
		\ead{zihao.sheng@wisc.edu}
            \author{Chengyuan Ma}
		\ead{cma97@wisc.edu}
		\author{Sikai Chen\corref{cor1}}
		\ead{sikai.chen@wisc.edu}
		
		\cortext[cor1]{Corresponding author: Sikai Chen}
		
		\address{Department of Civil and Environmental Engineering, University of Wisconsin-Madison, Madison, WI, 53706, USA}
		
\begin{abstract}

Despite significant progress in autonomous vehicles (AVs), the development of driving policies that ensure both the safety of AVs and traffic flow efficiency has not yet been fully explored. In this paper, we propose an enhanced human-in-the-loop reinforcement learning method, termed the \textbf{Human as AI mentor-based deep reinforcement learning (HAIM-DRL)} framework, which facilitates safe and efficient autonomous driving in mixed traffic platoon. Drawing inspiration from the human learning process, we first introduce an innovative learning paradigm that effectively injects human intelligence into AI, termed \textbf{Human as AI mentor (HAIM)}. In this paradigm, the human expert serves as a mentor to the AI agent. While allowing the agent to sufficiently explore uncertain environments, the human expert can take control in dangerous situations and demonstrate correct actions to avoid potential accidents. On the other hand, the agent could be guided to minimize traffic flow disturbance, thereby optimizing traffic flow efficiency. In detail, HAIM-DRL leverages data collected from free exploration and partial human demonstrations as its two training sources. Remarkably, we circumvent the intricate process of manually designing reward functions; instead, we directly derive proxy state-action values from partial human demonstrations to guide the agents' policy learning. Additionally, we employ a minimal intervention technique to reduce the human mentor's cognitive load. Comparative results show that HAIM-DRL outperforms traditional methods in driving safety, sampling efficiency, mitigation of traffic flow disturbance, and generalizability to unseen traffic scenarios. The code and demo videos for this paper can be accessed at: \href{https://zilin-huang.github.io/HAIM-DRL-website/}{https://zilin-huang.github.io/HAIM-DRL-website/}.

\end{abstract}
		
		
	\end{frontmatter}

\section{Introduction}

Autonomous vehicles (AVs), with their potential to revolutionize the transportation industry, have garnered significant attention in recent years. This revolution promises enhanced road safety, optimized traffic flow, and improved fuel economy, as highlighted by recent studies~\citep{feng2023dense,dong2022development,shi2021effect,shi2023deep,jiang2022reinforcement,huang2023cv2x,HAN2023100104,zhu2020safe,wang2023gops}. Yet, fully realizing these promises is fraught with challenges. Among these, one of the most critical is finding an appropriate \textbf{driving policy} that ensures AVs' safety while optimizing traffic flow efficiency~\citep{zhu2020safe,zhu2021survey,di2021survey}. In general, after undergoing extensive offline training, AVs are capable of generating their own online driving policies based on perception results from their sensors. These driving policies serve as the strategy or set of rules to which AVs adhere when navigating, including the operational (e.g., pedal and brake control, turn signals), tactical (e.g., lane-changing, lane-keeping), and strategic levels (e.g., routing)~\citep{di2021survey}. However, crafting a universally applicable driving policy for AVs, ensuring safe, efficient, and harmonious driving, is exceptionally challenging due to the dynamic and unpredictable nature of road environments. This complexity arises not only from the numerous possible traffic scenarios~\citep{wu2023toward,peng2022safe,li2022efficient}, but also from the intricate interactions between human-driven vehicles (HVs) and AVs on the roads~\citep{shi2023deep,chen2020traffic}.

\subsection{IL and RL Methods for Driving Policy Learning}
Driving policy learning is the main focus of this paper. Within the domain of learning-based driving policy learning, there are two main branches: Imitation Learning (IL) and Reinforcement Learning (RL)~\citep{aradi2020survey}. IL aims to emulate human driving behavior by replicating demonstrated actions in specific situations, such as behavior cloning (BC)~\citep{le2022survey}, inverse RL (IRL)~\citep{huang2023conditional}, and generative adversarial imitation learning (GAIL)~\citep{ho2016generative}. During the demonstration phase, the novice agent avoids interacting with potentially risky environments, thereby ensuring training safety. However, two primary inherent problems are exposed: Firstly, distributional shift~\citep{codevilla2019exploring}, where accumulated errors over time cause a deviation from the training distribution, leading to control failure. Secondly, the limitations of asymptotic performance~\citep{silver2018general}, which are constrained by the proficiency of the demonstration source and unlikely to outperform it. In this context, conservative Q-learning (CQL)~\citep{kumar2020conservative} draws wide attention as it addresses the impact of out-of-distribution (OOD) actions in distributional shift problem. Conversely, RL employs a self-optimizing approach with the potential to address these imitation-related pitfalls~\citep{aradi2020survey}. RL-based methods, including proximal policy optimization (PPO) and soft actor-critic (SAC), have demonstrated successful applications across numerous challenging driving scenarios in AVs~\citep{kiran2021deep}. 

In RL, human intentions are encoded into a reward function, enabling the agent to freely explore and learn from its environment. Moreover, reward-shaping (RS) methods, such as PPO-RS~\citep{haarnoja2018soft} and SAC-RS~\citep{schulman2017proximal}, have been proposed to address the issue of potentially diminished learning efficiency when the reward signals generated by the environment are sparse. However, safety during training and testing phase still presents a crucial challenge~\citep{peng2022safe,li2022efficient,kiran2021deep,muhammad2020deep}. Due to the exploratory nature of RL, agents might inevitably encounter risky situations before learning how to avoid them. Constraint optimization-based methods have been developed to improve safety, such as trust region methods (e.g., constraint policy optimization, CPO~\citep{achiam2017constrained}) and Lagrangian methods (e.g., SAC-Lag~\citep{ha2021learning}, PPO-Lag~\citep{stooke2020responsive}). Nevertheless, these methods can only assure the upper bound of the agent's failure probability, lacking an explicit mechanism to prevent the occurrence of critical failures~\citep{li2022efficient}. Moreover, the sampling efficiency of interaction between the agent and its environment is often low, leading to significant consumption of computational resources and extended training times~\citep{wu2023toward}.

\subsection{Human-in-the-loop Learning Methods}
Recently, some studies have attempted to incorporate human intelligence into the training loop of traditional RL or IL paradigms, leading to the development of the \textbf{human-in-the-loop learning (HL)} method~\citep{christiano2017deep}. This method, drawing upon the robustness and high adaptability of human beings in context understanding and knowledge-based reasoning, has been proven to enhance the safety and sampling efficiency of the conventional RL or IL training process. This integration has led to tangible successes in various real-world applications, such as robotic motion control~\citep{krishna2022socially} and the development of large language models~\citep{ouyang2022training,qu2023envisioning}, with \textbf{ChatGPT} being a notable example. Generally, human intelligence in these contexts can manifest as human evaluation, human demonstration, and human intervention~\citep{christiano2017deep,wu2022safe,wu2023toward,li2022efficient}. For example, in the robotics community, three representative methods of HL include dataset aggregation (DAgger)~\citep{ross2011reduction}, human-gated DAgger (HG-DAgger)~\citep{kelly2019hg}, and intervention weighted regression (IWR)~\citep{mandlekar2020human}. DAgger rectifies the compounding error of BC by intermittently requesting additional demonstrations from human experts. Instead of providing demonstrations upon request, HG-DAgger and IWR allow the human expert to intercede during exploration and guide the agent back to secure states.

While HL has made numerous achievements in the robotics community, its application in the realm of driving policy learning still faces several limitations: (a) Previous studies have mostly relied on human input in passive roles, such as suggesting which actions were favorable~\citep{mandel2017add} or evaluating collected trajectories according to human preference~\citep{christiano2017deep}. This passive human involvement can pose a risk to the human-AI collaborative system, as the agent explores the environment without sufficient safeguards. (b) Some studies train a human expert policy by utilizing offline human demonstrations~\citep{nair2018overcoming}. Nevertheless, the failure to make use of data from the agent's free exploration can still result in OOD issues. (c) Many existing studies adhere to the traditional RL paradigm, which necessitates the design of a reward function~\citep{ibarz2018reward}. However, manually crafting a reward function that effectively encompasses all driving behaviors for a trained agent presents a significant challenge. (d) Although online takeover and demonstrations by humans could be more effective, long-term supervision can lead to exhaustion among human mentor~\citep{wu2023toward,li2022efficient}. To accommodate the physical response of human drivers in the real-world, careful design must be taken on when and how human experts should be involved in the learning process.

\begin{figure*}[t]
\centering
  \includegraphics[width=0.95\textwidth]{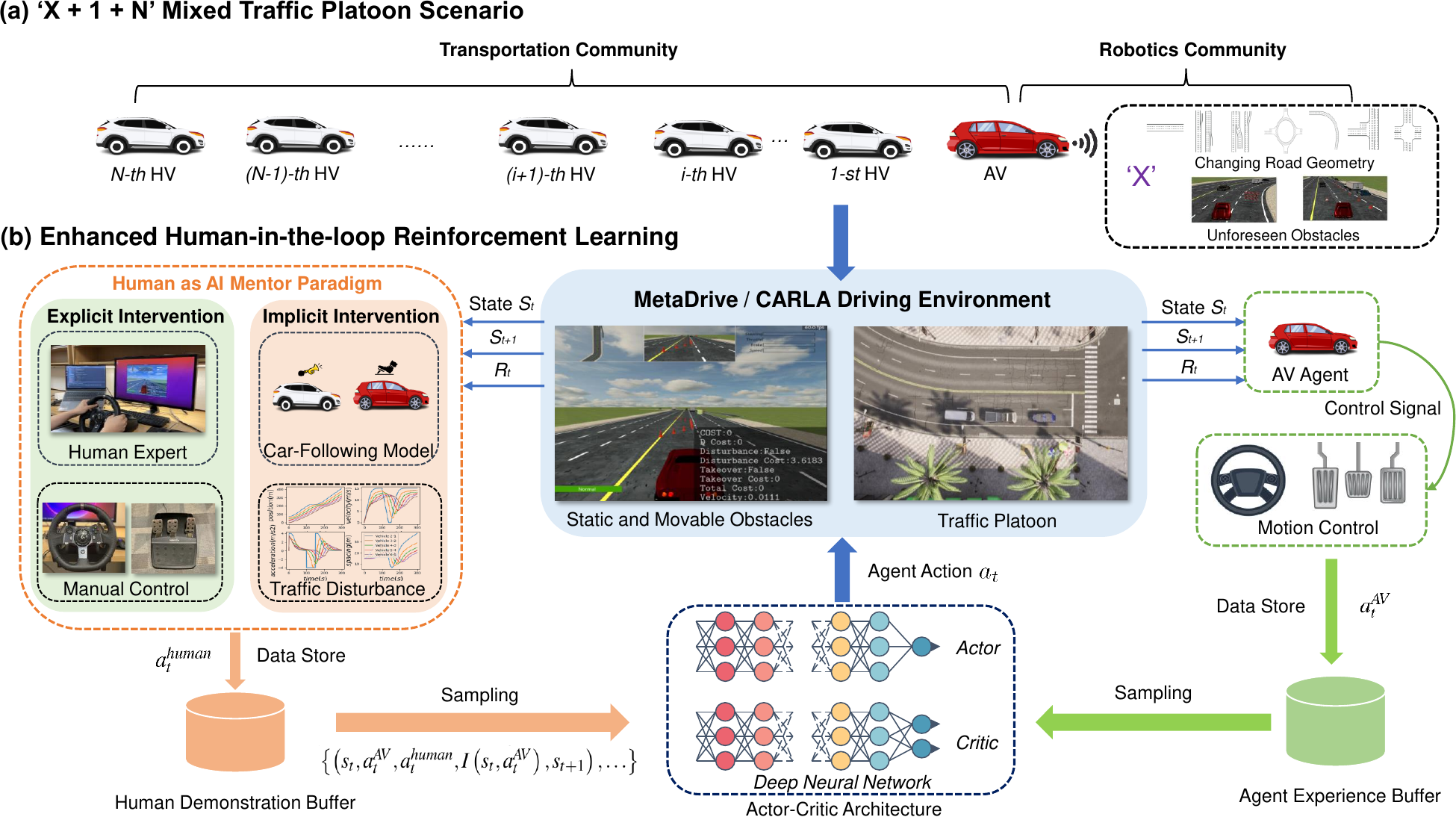}
  \caption{Overview of our proposed `$X+1+N$’ scenario and enhanced human-in-the-loop RL method. (a) The `$X+1+N$’ mixed traffic platoon is a novel concept uniting the transportation and robotics domains. This scenario is characterized by a combination of HVs and an AV, navigating through uncertain traffic environment. (b) HAIM is an innovative learning paradigm that integrates human intelligence into AI, thereby enhancing the learning capabilities of AI agents. Additionally, the HAIM-DRL can be seamlessly embedded into the MetaDrive/CARLA driving environment for testing.}
  \label{fig1}
\end{figure*}

More importantly, as shown in Fig.~\ref{fig1} (a), many studies in the robotics community focus on the safety of individual AVs, such as collision avoidance under environmental uncertainties~\citep{wu2022safe,wu2023toward,peng2022safe,li2022efficient}. Nonetheless, they often neglect the potential impact of these safety maneuvers employed for collision avoidance on subsequent traffic flow. This disregards a crucial factor in traffic management: traffic flow efficiency. On the other hand, many studies in the transportation community employ physics-based (rule-based) models to understand the impact of AVs on system-level performance, such as traffic flow disturbance~\citep{chen2020traffic,eleonora2023potential,ma2023anisotropy}, traffic congestion~\citep{zhou2022congestion,du2023dynamic,ha2023leveraging,zhuo2023evaluation,yue2022effects}, and traffic emission~\citep{stern2019quantifying,ansariyar2022investigating}. Meanwhile, some studies have begun integrating data-driven methodologies into rule-based models~\citep{shi2023deep,jiang2022reinforcement,HAN2023100104,zhu2020safe,chen2021graph,dong2021space,huang2018capturing,wu2019dcl,han2022physics,ding2022enhanced,liu2023longitudinal,dong2023did}. Although primarily targeting specific traffic scenarios, such as intersection optimization~\citep{wu2019dcl,wu2022intersection}, ramp merging~\citep{han2022physics,zhu2022merging}, and platoon control~\citep{shi2023deep, jiang2022reinforcement}, these models from the transportation community provide valuable insights into traffic management.

\subsection{Contributions}
To address these limitations and foster collaboration between the transportation and robotics communities, the primary objective of this paper is to develop an \textbf{enhanced human-in-the-loop RL (HRL)} method that effectively combines the strengths of both domains within the context of driving policy learning, as illustrated in Fig.~\ref{fig1}.

Firstly, we propose a novel concept, the `$X+1+N$’ mixed traffic platoon, which integrate the dynamics of mixed traffic flow and environmental uncertainties, illustrated in Fig.~\ref{fig1} (a). The `$1+N$' scenario, i.e., one leading AV, indexed as vehicle 0, followed by $N=\{1, \ldots, i, \ldots, N\}$ HVs, has already received extensive attention in the field of transportation~\citep{shi2023deep, jiang2022reinforcement,chen2021mixed,mohammadian2023continuum}. The innovation in the paper is its consideration of environmental uncertainties, denoted `$X$’. This includes unforeseen elements in real-world traffic, such as alterations in road geometry (e.g., straight, ramp, roundabout, intersection), sudden appearance of other vehicles, and the presence of obstacles (e.g., traffic cones, warning triangles). These elements require the leading AV to react and maneuver instantly, such as accelerating or decelerating, to ensure its own safety. The actions taken by the AV in response to unforeseen elements ripple through to the following HVs, affecting the stability and efficiency of the entire platoon. Different from other studies in robotics community, we meticulously consider the impact of  the AV's actions on the following HVs, aiming to reduce traffic flow disturbance across the platoon.

Secondly, we reframe the role of humans within the learning process, transforming them from passive evaluators who merely provide preference assessments to active participants, as illustrated in Fig.~\ref{fig1} (b). In contrast to some studies that require human intervention by terminating episodes and resetting the environment, which may be impractical in the real world, we opt for a more intimate human-AI interaction approach, allowing humans and agents to share autonomy~\citep{li2022efficient,wu2023toward}. Specifically, we integrate learning from intervention (LfI) and learning from demonstration (LfD) into a unified architecture. In this setup, human experts can intervene and override the actions of the original agents whenever they deem it necessary, and also provide real-time demonstrations through their actions. Moreover, the agent could be guided to avoid actions that cause significant traffic flow disturbance. Our key insight is that humans can serve as mentors for AI, and through the proposed HRL method, agents can rapidly learn to drive safely and efficiently in unseen scenarios. 

Compared to traditional methods (e.g., IL, safe RL, and conventional human-in-the-loop RL), comparative experiments indicate that our method achieves superior driving safety, reduced traffic flow disturbance, heightened sampling efficiency, and significant generalizability across various traffic environments \footnote{Experimental demo videos are available at: \href{https://www.youtube.com/playlist?list=PL-EmC8vF-RSH2j09uxZeyiVV_ePuA3Eud}{https://www.youtube.com/playlist?list=PL-EmC8vF-RSH2j09uxZeyiVV_ePuA3Eud}}. Our contributions are four-fold:
\begin{itemize}
    \item Inspired by the human learning process, we design an innovative learning paradigm that effectively injects human intelligence into AI, termed \textbf{human as AI mentor (HAIM)}. A defining feature of HAIM is that the human expert serves as a mentor to AI, supervising, intervening, and demonstrating in its learning process. HAIM includes explicit and implicit intervention mechanisms that can improve convergence speed of AI training.
    
    \item We propose a novel HAIM-based deep reinforcement learning framework, named \textbf{HAIM-DRL} \footnote{For those interested in exploring further, code can be accessed at: \href{https://github.com/zilin-huang/HAIM-DRL}{https://github.com/zilin-huang/HAIM-DRL}}, which utilizes data collected from both free exploration and partial human demonstrations as two training sources. By using HAIM-DRL, we can ensure not only the safety of the AV itself but also reduce the traffic flow disturbance, thus optimizing traffic flow efficiency. 

    \item We introduce a reward-free policy learning method that directly learns from human takeover signals, thus avoiding the complex task of manually crafting reward functions. Our key insight is that a latent value function, learned from active human involvement, can guide the learning policy to align with human intentions as represented through this value function.

    \item We design the explicit and implicit intervention value function to train agents for unseen traffic scenarios with minimal human takeover cost and traffic flow disturbance. Specifically, the human mentor's cognitive load can be reduced by minimizing the takeover cost; meanwhile, it can promote courteous driving and avoid behaviors such as sudden braking by minimizing the disturbance cost.
\end{itemize}

The rest of this paper is structured as follows: Section 2 provides the preliminaries. Section 3 illustrates the inspiration behind HAIM and its two key mechanisms. Section 4 describes the HAIM-DRL framework in detail, including the reward-free actor-critic architecture and the value and policy networks within it. Section 5 presents the experimental setup and results, including comparative and ablation experiments. Section 6 serves as the conclusion of the paper and discusses potential future research directions in this field.

\section{Preliminaries}
\subsection{Markov Decision Process}
In this study, our goal is to find a generalized solution for various scenarios to the problem of learning safe and efficient driving policies for AVs in mixed traffic platoon. This problem can be modeled by an infinite-horizon Markov decision process (MDP) with the form $\mathcal{M} = (\mathcal{S}, \mathcal{A}, \mathcal{P}, \mathcal{R}, \gamma, d_{0})$. Here, $\mathcal{S}$ and $\mathcal{A}$ denote the finite state space and action space, respectively. $\mathcal{P}: \mathcal{S}\times\mathcal{A}\times\mathcal{S}\to[0,1]$ is the transition probability function that describes the dynamics of the system. $\mathcal{R}: \mathcal{S}\times\mathcal{A}\to[\mathcal{R}_{\min}, \mathcal{R}_{\max}]$ is the reward function. $d_0:\mathcal{S}\to[0,1]$ is the initial state distribution. $\gamma\in(0,1)$ is the discount factor. 

 The state-action value and state value functions of $\pi$ are defined as $Q^\pi(s,a)=\mathbb{E}_{s_{0}=s,a_0=a, a_{t} \sim \pi\left(\cdot \mid s_{t}\right), s_{t+1} \sim p\left(\cdot \mid s_{t}, a_{t}\right)}$ $\left[\sum_{t=0}^{\infty} \gamma^{t} R\left(s_{t}, a_{t}\right)\right]$ and $V^\pi(s)=\mathbb{E}_{a\sim\pi(\cdot|s)}Q^\pi(s,a)$. Here, $\pi$ denotes a stochastic policy $\pi:\mathcal{S}\times\mathcal{A}\to[0,1]$. The goal of conventional RL is to learn a \textit{agent policy} $\pi_{agent}^*(a \mid s): \mathcal{S} \times \mathcal{A} \rightarrow[0,1]$ that maximizes the expected cumulative return: $J_R(\pi_{agent})=\mathbb{E}_{\tau \sim \pi_{agent}}\left[\sum_{t=0}^{\infty} \gamma^t R\left(s_t, a_t\right)\right]$, where $\tau=\left\{\left(s_t, a_t\right)\right\}_{t \geq 0}$ is the sample trajectory and $\tau \sim \pi_{agent}$ accounts for the distribution over trajectories depending on $s_0 \sim d_0, a_t \sim \pi_{agent}\left(\cdot \mid s_t\right), s_{t+1} \sim P\left(\cdot \mid s_t, a_t\right)$.

\subsection{Longitudinal Dynamical Modeling of Mixed Platoon}

To model the dynamics of the `$X+1+N$’ mixed traffic platoon, we denote the position and velocity of HV $i$ at time $t$ as $loc_{i,t}^{HV}$, $v_{i,t}^{HV}$, respectively. The headway distance between vehicle $i$ and $i-1$, which is their relative (bumper-to-bumper) distance, is defined as $d_{i,t}^{HV}=loc_{i-1,t}^{HV}-loc_{i,t}^{HV}$. Additionally, $acc_{i,t}^{HV}$ represents the acceleration of vehicle $i$, and $\dot{d}_{i,t}^{HV}=v_{i-1,t}^{HV}-v_{i,t}^{HV}$ represents the relative velocity between HV $i$ and $i-1$. For simplicity, we disregard the vehicle length. Extensive research has meticulously dissected the car-following dynamics of HVs, resulting in the development of fundamental models such as the Optimal Velocity Model (OVM)~\citep{bando1995dynamical} and the Intelligent Driver Model (IDM)~\citep{treiber2000congested}. 

As demonstrated in \cite{chen2021mixed}, most of these models can be expressed in the following general form: 
\begin{equation}
acc_{i,t}^{HV}=F\left(d_{i,t}^{HV}, \dot{d}_{i,t}^{HV}, v_{i,t}^{HV}\right)
\label{Eq1}
\end{equation}
which implies that the acceleration of HV $i$ is determined by its headway distance, relative velocity, and its own velocity.

In the state of traffic equilibrium, every vehicle travels at the same equilibrium velocity ${v^{HV}}^*$, i.e., $v_{i,t}^{HV}={v^{HV}}^*$, $\dot{d}_{i,t}^{HV}=0$. Simultaneously, every vehicle maintains a corresponding equilibrium headway distance ${d_{i}^{HV}}^*$. Based on Eq.~\ref{Eq1}, where it holds that 
\begin{equation}
F\left({d_{i}^{HV}}^*, 0, {v^{HV}}^*\right)=0
\label{Eq2}
\end{equation}

Assuming that each vehicle encounters a minor deviation from the equilibrium state $\left({d_{i}^{HV}}^*, {v^{HV}}^*\right)$, the discrepancy between the actual and equilibrium states for HV $i$ can be defined as follows:
\begin{equation}
\tilde{d}_{i,t}^{HV}=d_{i,t}^{HV}-{d_{i}^{HV}}^*, 
\quad 
\tilde{v}_{i,t}^{HV}=v_{i,t}^{HV}-{v^{HV}}^*
\label{Eq3}
\end{equation}

To strike a balance between model fidelity and computational tractability, we adopt the IDM~\citep{treiber2000congested} as the representative car-following model for the HVs in Fig.~\ref{fig1} (a). For the leading AV, it navigates using its throttle and steering signals, serving as the sole external inputs for the entire mixed platoon system.

\section{Human as AI Mentor Paradigm}
\subsection{Inspiration Behind the HAIM Paradigm}
\begin{figure}[!ht]
  \centering
  \includegraphics[width=0.85\textwidth]{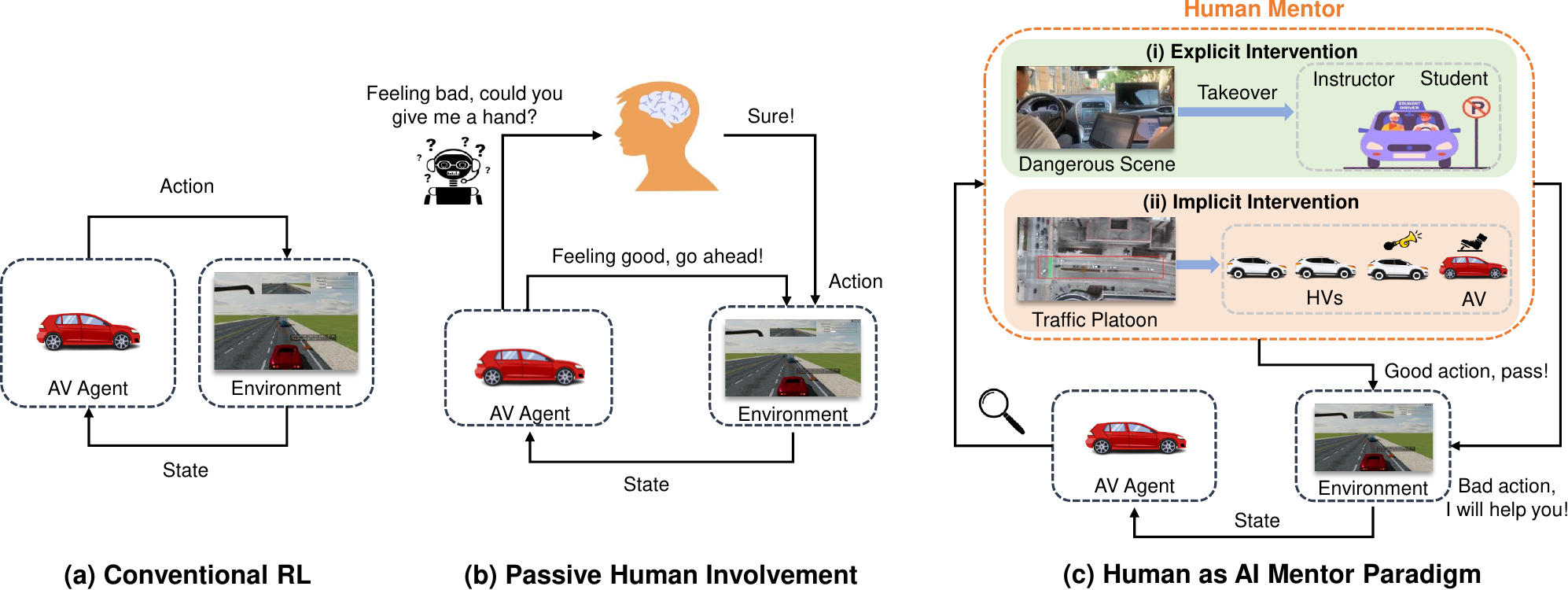}
  \caption{Illustration of the HAIM in the driving school scenario. (a) Traditional RL paradigm learns from trial and errors. (b) Passive
human involvement paradigm merely provides suggestions about which actions are good or evaluates collected trajectories. (c) The proposed HAIM paradigm enables the human expert to assume control in hazardous situations and demonstrate the correct actions to prevent potential accidents. Additionally, the agent is instructed to reduce the disturbance to traffic flow by other road users.}\label{fig2}
\end{figure}

In traditional RL paradigm, as shown in Fig.~\ref{fig2} (a), it heavily relies on exploration through trials and errors, which requires significant training time and inevitably puts the agent in dangerous situations. On the other hand, passive human involvement, as depicted in Fig.~\ref{fig2} (b), merely provides suggestions about which actions are appropriate or evaluates collected trajectories based on human preference. While it helps to shorten the training time, it still exposes the agent to potential hazards. When observing the human learning process, especially in acquiring practical skills, one notices that it does not rely solely on trial and error. Instead, humans often employ explicit and implicit intervention to enhance learning efficiency and ensure the safety of the learning process. As an illustration, consider the driving school scenario depicted in Fig.~\ref{fig2} (c).  In this scenario, while learning to drive, a student with a learner's permit is permitted to operate a vehicle from the driver's seat under the close supervision of an experienced instructor.

Generally, students master driving skills through three approaches: (a) Learning from the instructor's demonstration: Initially, a proficient instructor offers a detailed demonstration, emphasizing the skills and strategies needed for driving in various scenarios. By observing the instructor's actions, the student grasps these demonstrated behaviors and quickly learns through imitation. (b) Taking control under the instructor's supervision: After gaining a basic understanding of driving, students begin to take the reins, operating the vehicle in a more exploratory manner. Here, the instructor acts as a guardian, intervening to prevent potential accidents when students make dangerous moves or encounter complex traffic situations. (c) Learning to become a courteous driver through feedback from other drivers: During the road experience phase, the student's main source of learning shifts from the instructor to feedback from other road users. For example, if the student suddenly brakes, the driver behind may honk the horn. This serves as indirect feedback, indicating to the learner that their action has disrupted the flow of traffic.

\subsection{Details of the HAIM Paradigm}
Drawing inspiration from the human learning process, we propose an innovative learning paradigm, i.e., HAIM, addressing limitation (a) we previously mentioned. In particular, the human mentor not only provides high-quality demonstrations and ensures the safety of the agent's exploration but also instructs the agent to consider system level performance. To achieve this goal, we have carefully designed explicit and implicit intervention mechanisms.  

\subsubsection{Explicit Intervention Mechanism} \label{Explicit Intervention Mechanism}
The role of the explicit intervention mechanism is similar to that of an instructor in the driving school scenario. Specifically, human experts provide guidance, supervision, and direct intervention during the agent’s learning process. For this purpose, we adopt a technique termed the \textit{`switching function'}~\citep{peng2022safe}, which allows the agent to dynamically alternate between exploration and explicit intervention. 

\begin{figure}[t]
  \centering
  \includegraphics[width=0.95\textwidth]{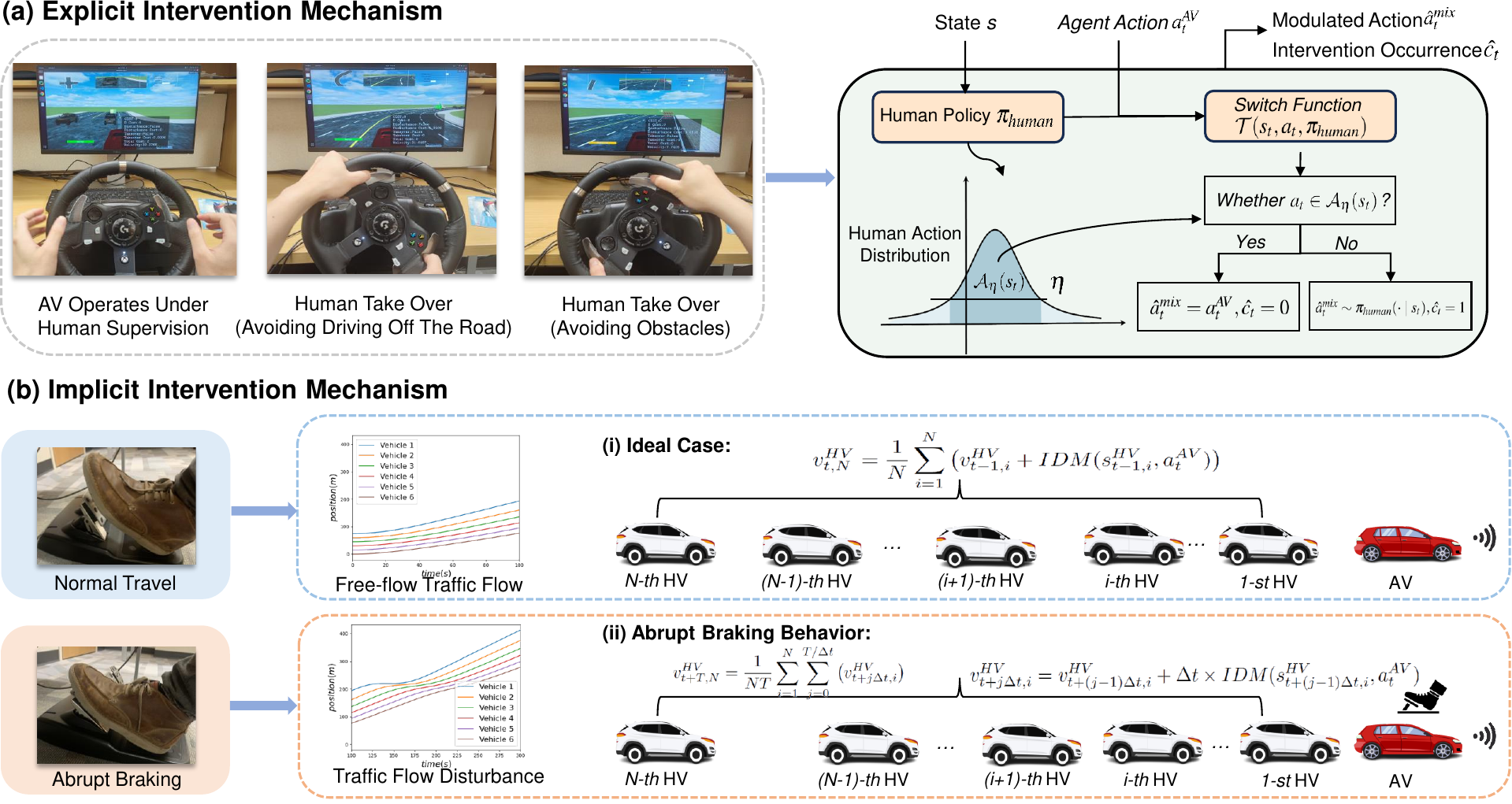}
  \caption{Explicit and implicit intervention mechanism in HAIM. (a) The former involves humans actively taking direct control of the AV agent, guiding it through correct behaviors in hazardous scenarios. (b) The latter involves penalizing the agent for actions that disrupt traffic, indirectly indicating that it should avoid such actions in the future.}\label{fig3}
\end{figure}

Let us define a well-performing human policy as $\pi_{human}: a_t^{human} \sim \pi_{human}(\cdot \mid s_t)$, which can consistently produce safe and reliable actions, denoted as $a_t^{human}$. Our goal is to train an AV agent equipped with a policy $\pi_{AV}(a_t^{AV} \mid s_t)$ so that it can make wise decisions, represented as $a_t^{AV}$, given the state $s_t$. As illustrated in Fig.~\ref{fig3} (a), the switch function $\mathcal{T}$ determines the state and timing for human expert takeover, allowing them to demonstrate correct actions to guide the learning agent. Additionally, this setup provides the probability $\pi_{human}(a_t^{AV} \mid s_t) \in[0,1]$, which reflects the degree of agreement between the human and the agent's action and can be used as an indicator for the switch function.

\textbf{Assumption 1 (High-quality human action).} \textit{The effectiveness of ensuring training safety through explicit intervention depends on the quality of the human policy. In this context, we assume that $a_t^{human}$ is generated by an expert-level human policy and is safe and reliable most of the time. In detail, the step-wise probability of that the human expert produces an unsafe action after takeover is bounded by a small value $\epsilon < 1$. To be more specific, in the implementation of our experiments, we set a requirement that the number of failures by the human expert should not exceed once.}

\textbf{Assumption 2 (Expert-level human takeover).} \textit{While some studies attempt to establish an intervention threshold for identifying 'dangerous situations' \citep{peng2022safe}, this task is challenging due to the involvement of numerous non-quantifiable factors, as discussed by \cite{wu2023toward}. Drawing inspiration from the methods used by instructors to assess risky situations in driving school scenarios, we rely on human expert evaluations of safety during the training process rather than on external objective criteria. We assume that this expert-level human can accurately identify dangerous situations most of the time. In detail, the step-wise probability of that the human expert does not takeover when agents generates an unsafe action is bounded by a small value $\kappa<1$.}

\textit{For Assumptions 1 and 2, we derive the upper bound for the failure discount probability of HAIM's explicit intervention mechanism in \ref{Appendix:Proof of Explicit Intervention Mechanism}, demonstrating our ability to limit training risks associated with the human expert. }

\textbf{Definition 1 (Probability-based Switch Function).} The switch function $\mathcal{T}$ considers both the agent’s and human actions, thereby producing the modulated action $a_t^{mix}$, which is subsequently applied to the environment. Here. the intervention occurrence $\hat{c_t}$ indicates whether the human is in control. In this context, the intervention occurrence $\hat{c}$ indicates whether the human is in control. The function $\mathcal{T}(s_t, a_t, \pi_{human})$ determines which policy is responsible for the action, as discussed in \cite{peng2022safe}:
\begin{equation}
\begin{aligned}
\mathcal{T}(s_t, a_t, \pi_{human})=(a_t^{mix}, \hat{c_t}) = \begin{cases}
    (a_t^{AV}, 0), & \text { if } a_t \in \mathcal{A}_\eta(s_t) \\ 
    \left(a_t^{human} \sim \pi_{human}(\cdot \mid s_t), 1\right), & \text { otherwise }
\end{cases}
\label{Eq4}
\end{aligned}
\end{equation}
where, $\eta$ represents the confidence level in the human action probability, and $\mathcal{A}_\eta(s_t)$ is defined as the set of confident actions within the action space: $\mathcal{A}_\eta(s_t)=\{a_t \in \mathcal{A}: \pi_{human}(a_t^{AV} \mid s_t) \geq \eta\}$. If intervention is deemed necessary, the human takes action $a_t^{human}$ to override the agent's action $a_t^{AV}$.

\textit{Remark 1:} While several methods can convey human experts' intentions with the current action, such as audio or tactile stimulation, we opt for steering wheel takeover. This choice enables us not only to express dissatisfaction with the agent's actions but also to demonstrate the correct maneuvers. It is analogous to the driving school scenario, where instructors rectify students' hazardous maneuvers by directly assuming control of the steering wheel.

\textbf{Definition 2 (Action-based Switch Function).} For the driving policy learning problem, a straightforward approach to designing the switch function is to trigger intervention when the agent's action deviates from the human's action. Let $\mathcal{T}(a_t) = 1$ indicate that the human assumes control, and $\mathcal{T}(a_t) = 0$ otherwise. We represent this process using the action-based switch function as follows: 
\begin{equation} 
\begin{aligned}
\mathcal{T}(s_t, a_t, \pi_{human}) = \begin{cases}
    (a_t^{AV}, 1), & \text {if takeover } \\ 
    \left(a_t^{human} \sim \pi_{human}(\cdot \mid s_t), 0\right), & \text { otherwise }
\end{cases}
\label{Eq5}
\end{aligned}
\end{equation}

We employ a Boolean indicator \(I\left(s_t\right)\) to denote human takeover, and the action applied to the environment, defined as \(a_t^{mix} = I\left(s_t\right) a_t^{human}+\left(1-I\left(s_t\right)\right) a_t^{AV}\). Subsequently, the human policy and agent policy collaborate to form a mixed behavior policy $\pi_{mix}$. As a result, the actual trajectory during the training process is determined by the mixed behavior policy:
\begin{equation} 
\pi_{mix}(a \mid s)=\pi_{AV}(a \mid s)(1-I(s, a))+\pi_{human}(a \mid s) F(s)
\label{Eq6}
\end{equation}
where, $F(s)=\int_{a^{\prime} \notin \mathcal{A}_\eta(s)} \pi_{AV}\left(a^{\prime} \mid s\right) d a^{\prime}$  represents the probability of the agent selecting an action that would be rejected by the human. This setup eliminates unnecessary states and mitigates the safety concerns associated with traditional IL and RL methods.

\textit{Remark 2:} The introduction of explicit intervention primarily serves to ensure safety during training. We only require a single human expert to achieve our objectives. In practice, different human experts may have varying preferences, including different takeover timings and actions after takeover. However, as long as these preferences align with the criteria outlined in Assumption 1 and 2, they can ensure training safety despite their individual differences. For instance, in the context of driving school, students may learn from different instructors with diverse teaching styles. Despite these differences, all instructors ensure that students learn to drive safely. Furthermore, in the ablation studies, we explore two distinct takeover strategies and find that dense human takeover signals are more effective.

\subsubsection{Implicit Intervention Mechanism}
The implicit intervention mechanism is inspired by how learners gain feedback not only from explicit instructor guidance but also from the behavior and reactions of other road users. For a given state $s_t$ and action $a_t$, we typically calculate $Q (s_t, a_t)$ based solely on the immediate impact of $a_t$. Nevertheless, this setup does not effectively evaluate the influence of the current action on future traffic flow. To address this limitation, we propose a model-based prediction method to more accurately assess the impact of current actions on future traffic flow.

\textbf{Assumption 3 (Model-based Traffic Flow Dynamics).} \textit{We assume that the AV's action $a_t^{AV}$ will persist for a short time interval $T$, and the following vehicles will adhere to the IDM~\citep{treiber2000congested}. This assumption is simplified and omits various factors present in real-world scenarios, such as multiple lanes, diverse driver behaviors, and varying vehicle parameters. Importantly, the traffic state we are predicting here is solely used for computing the disturbance cost and interpreting the impact of different agent actions on traffic flow. }

\textbf{Definition 3 (Current Traffic Flow State).} The average velocity of all following vehicles at time $t$, $v_{t,N}^{HV}$, can be calculated as follows:
\begin{equation}
v_{t,N}^{HV} = \frac{1}{N} \sum_{i=1}^N \left( v_{t-1,i}^{HV} + IDM(s_{t-1,i}^{HV}, a_t^{AV}) \right)
\label{Eq7}
\end{equation}
where $s_{t-1,i}^{HV}$ represents the state of the $i^{th}$ HV at time $t-1$. Note that the IDM takes the state of the $i^{th}$ HV, $s_{t-1,i}^{HV}$, and the action of the AV, $a_t^{AV}$, into consideration to determine the velocity of the $i^{th}$ HV at time $t$. 

\textbf{Definition 4 (Predicted Traffic Flow State).} After a duration period $T$, the average velocity of all HVs $v_{t+T, N}^{HV}$ is calculated by
\begin{equation}
v_{t+T,N}^{HV} = \frac{1}{NT} \sum_{i=1}^N \sum_{j=0}^{T/\Delta t} \left( v_{t+j\Delta t,i}^{HV} \right)
\label{Eq8}
\end{equation}
\begin{equation}
v_{t+j\Delta t,i}^{HV} = v_{t+(j-1)\Delta t,i}^{HV} + \Delta t \times IDM(s_{t+(j-1)\Delta t,i}^{HV}, a_t^{AV})
\label{Eq9}
\end{equation}

\textit{Remark 3:} In practice, the current traffic flow state can be obtained through onboard sensors such as LIDAR and camera~\citep{sheng2023epg}. Additionally, it is possible to acquire velocity information for the following HVs of the leading AV at the current time via roadside units (RSUs)~\citep{huang2023cv2x,linrssi,olovsson2022future}. Furthermore, aside from model-based methods, we can also leverage deep learning techniques to enhance the prediction accuracy of future traffic flow states, as demonstrated in our previous works~\citep{wu2020combined,li2022st,xu2023agnp}.

By using Eqs. \ref{Eq7} and \ref{Eq8}, we can calculate the average velocity of the following HVs at the current time $t$, which reflects the current traffic flow state. Additionally, we are able to predict the possible traffic flow state that could arise following the execution of the current action $a_t^{AV}$. In the following sections, we will integrate these two indicators as components of the disturbance cost into the RL training process. This will guide AVs in making decisions that balance their own performance with the system-level performance.

\section{HAIM-based DRL Framework for Driving Policy Learning}
\subsection{HAIM-DRL Overview}

\begin{figure*}[!ht]
  \centering
  \includegraphics[width=0.98\textwidth]{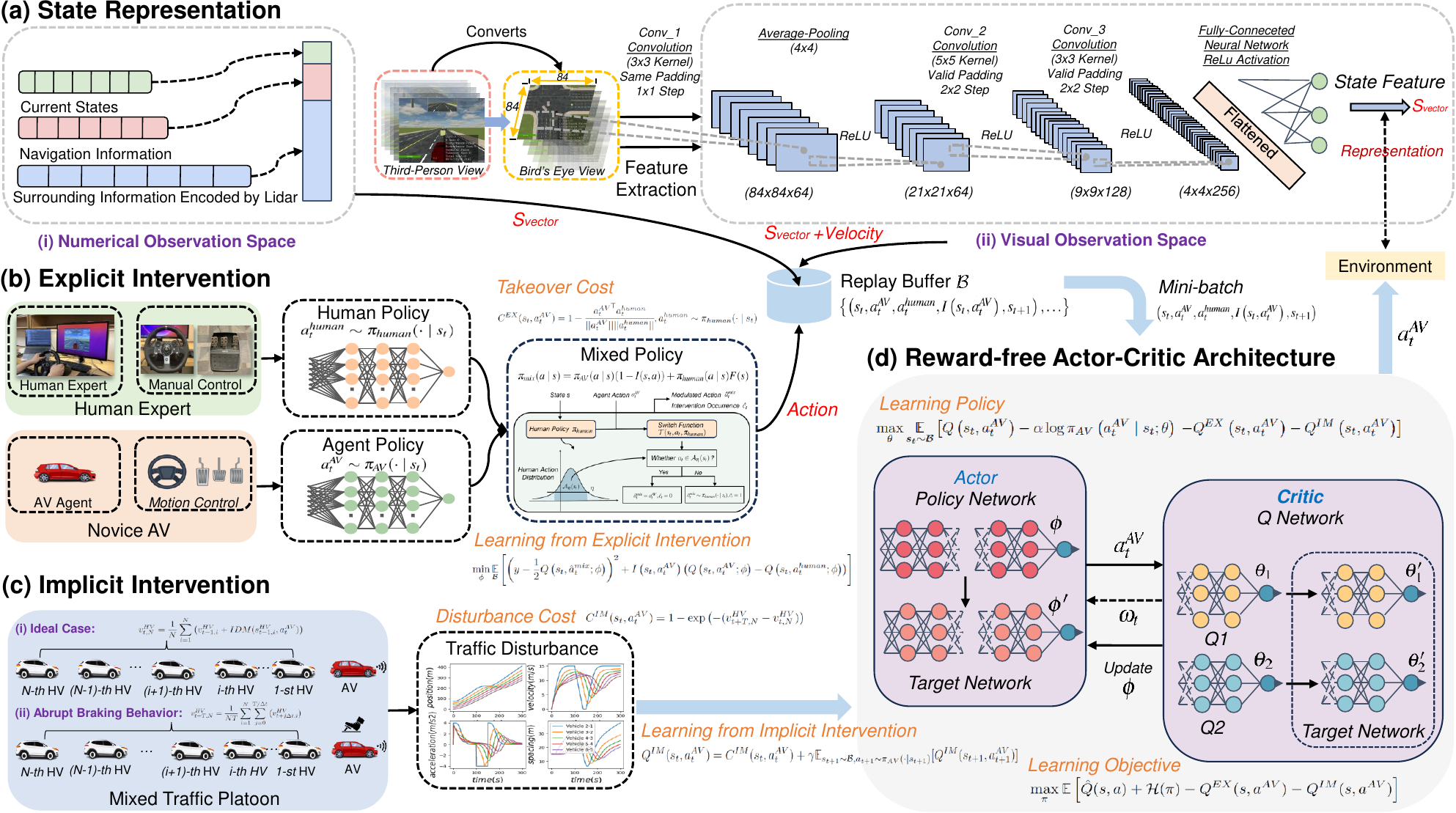}
  \caption{Schematic of the HAIM-DRL framework for driving policy learning.}
  \label{fig4}
\end{figure*}

As shown in Fig.~\ref{fig4}, the HAIM-DRL framework is meticulously designed to ensure the safety of individual AVs amidst environmental uncertainties, while minimizing disturbance to following HVs to optimize overall traffic flow efficiency. We first delineate the observation space of our HAIM-DRL framework, which is designed to accommodate two distinct types of sensory inputs: (i) The first type pertains to numerical observation spaces that encompass data such as current vehicular states, navigation information, and surrounding information encoded by LiDAR. (ii) The second type addresses visual observation spaces, which include image inputs from mounted cameras. Specifically, we employ a convolutional neural network (CNN) to extract feature vectors, which are then combined with velocity. 

Then, we introduce a reward-free policy learning method that explicitly conveys human intentions to the learning policy through human takeover. To encourage the agent to explore within the permissible state-action space, we maximize the entropy of the agent's action distribution, as long as the agent is not undergoing takeover. Simultaneously, we reduce traffic flow disturbance by minimizing the disturbance cost estimated from the current vehicle state and action. Lastly, we also take measures to minimize the cognitive load on the human mentor. Specifically, we reduce takeover costs during training as much as possible, thereby decreasing reliance on expert demonstrations and enhancing the autonomy of the learning agent.

\subsection{Reward-Free Off-Policy Actor-Critic Architecture}
\subsubsection{Vanilla Off-Policy Actor-Critic}
Off-policy actor-critic is a well-known algorithm architecture in the RL field, which consists of two core components: actor and critic~\citep{lillicrap2015continuous}. The actor's task is to generate a policy; in other words, it determines the action to be taken in a given state. Conversely, the `critic' is responsible for estimating the potential value of executing such an action in that state. This architecture is referred to as an `off-policy', as its learning process can be independent of the current policy, learning instead from actions executed under different policies. 

Under a policy $\pi$, the critic $Q$ is defined as follows~\citep{lillicrap2015continuous}:
\begin{equation} 
Q^\pi(s_t, a_t) = \mathbb{E}_{(s_{t+1},a_{t+1})\sim p(.,.|s_t,a_t)}[r_t + \gamma Q^\pi(s_{t+1}, a_{t+1})]
\label{Eq10}
\end{equation}
where, $\mathbb{E}$ denotes the expectation, $\gamma$ is the discount factor, and $p$ represents the state transition dynamics.

The policy function is subsequently attained by maximizing $Q$. This is mathematically expressed as
\begin{equation}
\pi = \arg\max_{\pi} \mathbb{E}_{(s,a)\sim \rho_\pi} Q^\pi(s, a)
\label{Eq11}
\end{equation}
where, $\rho_\pi$ is the state-action distribution under policy $\pi$. The actor and critic are usually parameterized by neural networks with parameters $\phi$ and $\theta$, respectively. 

The objective functions of actor and critic, denoted as $L_{\pi}$ and $L_{Q}$, are defined as
\begin{equation}
L_Q(\theta) = \mathbb{E}_{s_{t+1}\sim p(.|s_t,a_t)}[r_t + \gamma Q(s_{t+1}, \pi(.|s_{t+1}; \phi); \theta) - Q(s_t, a_t; \theta)]^2
\label{Eq12}
\end{equation}
\begin{equation}
L_\pi(\phi) = -\mathbb{E}_{s_t\sim\rho_\pi}[Q(s_t, \pi(.|s_t; \phi); \theta)]
\label{Eq13}
\end{equation}

The Eq.~\ref{Eq13} assumes that the reward can reflect users' intentions and effectively incentivize the desired behaviors of the agents. Reward functions are typically designed through trial-and-error methods. For instance, ~\cite{shi2021effect,jiang2022reinforcement,wu2023toward} manually designed the reward function for autonomous driving, focusing on driving safety and comfort. On the other hand, ~\cite{huang2023conditional} employed IRL to infer the reward function from human demonstrations. However, as mentioned earlier, it is nearly impractical to encapsulate all driving behaviors facing various unforeseeable traffic scenarios in reward functions.

\subsubsection{Removing the Reward Function} \label{Motivation of Removing the Reward Function}
In a recent survey~\citep{booth2023perils}, it was found that 92\% of surveyed expert-level RL practitioners reported using manual trial-and-error reward designs, with 89\% indicating that their designed reward was suboptimal. \cite{knox2023reward} developed eight sanity checks to identify flaws in reward functions and surveyed papers on RL-based AV published in top-tier venues, finding near-universal flaws in reward design for AVs. Thus, designing a reward function that is universally applicable to all scenarios for AVs is a complex and challenging task. Recent work has demonstrated that agents can learn complex goals by interacting with humans without a fixed reward function~\citep{li2022efficient,ibarz2018reward,peng2023learning}.

The distinction of this work from previous efforts is the removal of the reward function. Revisiting the primary goal in the driving policy learning task, we find that reward functions are not necessary, as our true aim is to implement human preference in learning behavior. We observe that when a human expert intervenes by taking over the agent, a clear signal is generated, indicating dissatisfaction with the agent's current performance, whether due to unsafe actions or poorly performing behaviors. Conversely, if the human experts do not intervene, then the current states and the agent's actions are deemed to be human-compatible~\citep{peng2023learning}. Based on this observation, we can use a proxy value function to denote human preference instead of the traditional reward function; this does not include reward but only calculates a proxy $Q$ values~\citep{peng2023learning}. 

\textbf{Definition 5 (Proxy Value Function without Reward).} A key insight is that we can manipulate the proxy $Q$ value to induce the desired behavior, since value-based RL has the property of seeking value-maximizing policy, as shown in Eq. \ref{Eq11}. Specifically, if immediate reward are discarded, the $Q$ value update changes from $Q^\pi\left(s_t, a_t\right) \leftarrow R(s, a)+\gamma \max _{a_{t+1}} Q^\pi\left(s_{t+1}, a_{t+1}\right)$ to the proxy $Q$ value $\hat{Q}^\pi\left(s_t, a_t\right) \leftarrow \gamma \max _{a_{t+1}} \hat{Q}^\pi\left(s_{t+1}, a_{t+1}\right)$. This practice of removing reward would mitigate to limitation (c) mentioned above.

Based on the learned value function, the policy $\pi_\theta$ can be learned by optimizing
\begin{equation}
\theta=\arg \max _\theta \mathbb{E}_{(s, a) \sim \rho_\pi} \hat{Q}^\pi(s, a)
\label{Eq14}
\end{equation}

\textit{Remark 4:} Although Eq. \ref{Eq14} uses proxy $Q$ values, this does not affect the formulation of MDP, we just don't need to keep track of the reward metrics of MDP. The temporal difference (TD)-based method first relabels the proxy $Q$ value of partial human demonstrations, and propagates them to other states. Subsequently, the policy is optimized to align with the human intentions represented through this value function. Since the partial demonstration data collected from human takeover are highly off-policy for the learning agent, we employ offline RL techniques to maintain the proxy value function, as detailed in Section~\ref{Learning from Explicit Intervention}. 

By using Eq. \ref{Eq14}, we can transform a standard RL task into a reward-free setting where the agent can learn from active human involvement. The advantage of this is that we can avoid manually designing the reward function, which is very tricky in practice. This is because an ideal driving policy should be capable of adapting skills in diverse traffic situations, including overtaking, yielding, emergency stopping, and negotiating with other vehicles.

\subsection{Learning Objectives for Value Network}
We propose a comprehensive set of objectives that can effectively utilize existing human data to optimize the driving policy of AV agent in mixed traffic platoon: (a) The agent should aim to maximize the proxy value function, denoted as $\hat{Q}(s, a)$, which reflects human intentions. (b) The agent should actively explore the state-action space within the range allowed by humans. This is achieved by maximizing the entropy of the action distribution, denoted as $\mathcal{H}(\pi(\cdot \mid s))$. (c) The agent should strive to reduce the cognitive load of the human mentor by minimizing the explicit intervention value function, denoted as $Q^{EX}(s, a^{AV})$. (d) The agent should consider the impact of its actions on subsequent vehicles and aim to minimize traffic flow disturbance by minimizing the implicit intervention value function, denoted as $Q^{IM}(s, a^{AV})$. Overall, the learning objective of HAIM-DRL can be summarized as follows:

\begin{equation}
\max _\pi \mathbb{E}\left[\hat{Q}(s, a)+\mathcal{H}(\pi)-Q^{EX}(s, a^{AV})-Q^{IM}(s, a^{AV})\right]
\label{Eq15}
\end{equation}

Next, we will delve into the practical implementation details of the previously mentioned design objectives.

\subsubsection{Learning from Explicit Intervention} \label{Learning from Explicit Intervention}

According to the observation in Section~\ref{Motivation of Removing the Reward Function}, we should strive to make agent's behavior close to the behavior demonstrated by human experts, as well as avoid performing actions that human experts have intervened in. A closer examination of Eqs. \ref{Eq13} and \ref{Eq14} reveals that the optimal deterministic strategy will consistently choose the action with the highest $Q$ value. Thus, if human experts intervene in certain states, the human actions $a^{human} \sim \pi_{human}$ should always have higher values than other actions in these states. Meanwhile, the agent actions $a^{AV} \sim \pi_{AV}$ should always have lower values than other actions, as they are rejected by human experts.

Given that we use a safe and reliable mixed behavior policy, i.e., $\pi_{\operatorname{mix}}(a \mid s)$, for environmental exploration, part of the collected transition sequences observed during the takeover are considered as partial demonstrations, denoted as $\left\{\left(s_t, a_t^{AV},a_t^{human}, I\left(s_t,  a_t^{AV}\right), s_{t+1}\right), \ldots\right\}$. The partial demonstrations and transitions resulting from free exploration will be captured and stored in the replay buffer, denoted as $\mathcal{B}$. These will then be integrated into the training pipeline. It's important to note that there is no need to record environmental reward and cost in this buffer. 

As we can only learn from partial demonstrations in $\mathcal{B}$, this leads to a distribution shift problem. To address this issue, we adopt the recent CQL \citep{kumar2020conservative} to train using off-policy data generated by humans. In detail, we sample data $\left(s_t, a_t^{AV},a_t^{human}, I\left(s_t,  a_t^{AV}\right)\right)$ from the replay buffer $\mathcal{B}$, and label the proxy $Q$ value of human action $a_t^{AV}$ with $\hat{Q}\left(s_t, a^{human}\right)$ and the agent action $a_t^{AV}$ with $\hat{Q}\left(s_t, a^{AV}\right)$.

The optimization problem of the proxy value function is formulated as follows:

\begin{equation}
\begin{aligned}
\min _\phi \underset {\left(s_t, a^{AV}, a^{human}, I\left(s_t, a^{AV}\right)\right) \sim \mathcal{B}} {\mathbb{E}} \left[I\left(s_t, a^{AV}\right)\left(\hat{Q}\left(s_t, a^{AV} ; \phi\right) -\hat{Q}\left(s_t, a^{human} ; \phi\right)\right)\right]
\end{aligned}
\label{Eq16}
\end{equation}

The optimization objective described above can be understood as adopting an optimistic view towards human actions $a^{human}$, and a pessimistic stance regarding the agent's actions $a^{AV}$. By minimizing the difference between the proxy $Q$ values of human experts and the proxy $Q$ values of the agent, as formulated in Eq. \ref{Eq16}, we can guide the agent's actions towards the high-value state-action subspace preferred by human experts.

\subsubsection{Learning from Exploration with Entropy Regularization}

When the learning agent fails to adequately explore the human-preferred subspace during free exploratory sampling, states that trigger high proxy values are infrequently encountered. This poses a challenge in propagating the proxy value back to previous states, impeding the learning process. To encourage exploration, we utilize the entropy regularization technique \citep{haarnoja2018soft}, creating an auxiliary signal for updating the proxy value function:
\begin{equation}
\min _\phi \underset{\left(s_t, \hat{a}_t^{mix}, s_{t+1}\right) \sim \mathcal{B}}{\mathbb{E}}\left[y-\frac{1}{2}Q\left(s_t, \hat{a}_t^{mix} ; \phi\right)\right]^2
\label{Eq17}
\end{equation}
where
\begin{equation}
y = \gamma \underset{a_{t+1} \sim \pi_{AV}\left(\cdot \mid s_{t+1}\right)}{\mathbb{E}}\left[Q\left(s_{t+1}, a_{t+1} ; \phi^{\prime}\right)-\alpha \log \pi_{AV}\left(a_{t+1} \mid s_{t+1}\right)\right]
\label{Eq18}
\end{equation}
where $\hat{a}_t^{mix}$ is the action performed under state $s_t$, $\phi^{\prime}$ represents the delayed update parameters of the target network, and $\gamma$ is the discount factor. 

Since HAIM-DRL operates without the need for a reward, we exclude the reward term from the update target $y$. By combining Eqs.~\ref{Eq15} and \ref{Eq16}, the formal optimization objective for the proxy value function is established as follows:
\begin{equation}
\min _\phi \underset{\mathcal{B}}{\mathbb{E}}\left[\left(y-\frac{1}{2}Q\left(s_t, \hat{a}_t^{mix} ; \phi\right)\right)^2 + I\left(s_t, a_t^{AV}\right)\left(\hat{Q}\left(s_t, a_t^{AV} ; \phi\right)-\hat{Q}\left(s_t, a_t^{human} ; \phi\right)\right)\right]
\label{Eq19}
\end{equation}

Different from previous offline RL works~\citep{nair2018overcoming,kumar2020conservative}, which train from a fixed dataset without access to closed-loop feedback, HAIM-DRL utilizes online exploration and partial human demonstration data. Furthermore,  HAIM-DRL ensures continuity in state visitation between the human expert and the agent, effectively mitigating the potential OOD issues mentioned earlier in limitation (b).

\subsubsection{Learning from Implicit Intervention}

Eq.~\ref{Eq2} illustrates that even minor changes in the state of the leading AV can lead to disturbances in the following HVs, disrupting the overall traffic flow balance. In the transportation community, there are many indicators used to assess the performance of the traffic system, such as acceleration, headway, average velocity, and time to collision (TTC)~\citep{shi2023deep, jiang2022reinforcement, zhu2020safe, mahdinia2021integration}. In particular, \cite{sharma2021assessing} proposed using a wavelet energy-based method to calculate traffic oscillation metrics (i.e., oscillation duration and amplitude) to analyze traffic flow disturbance. The main idea of their work is to identify significant changes in velocity. Inspired by this, and building upon Eq.~\ref{Eq3}, we propose a measurement method for disturbance cost, which can quantify the potential downstream impact of AVs on the following vehicles.

\textbf{Definition 6 (Disturbance Cost).} In the equilibrium traffic state, AVs do not cause disturbance. We can use Eq.~\ref{Eq7} to calculate the average velocity of all following HVs $v_{t,N}^{HV}$ at time $t$. Then, we can simulate the behavior of the following HVs using the IDM and calculate their desired velocity $v_{t+T,N}^{HV}$ after the AV has performed action $a_t^{AV}$ by using Eq.~\ref{Eq8}. The disturbance cost $C^{IM}$ can then be defined as
\begin{equation}
C^{IM}(s_t, a_t^{AV}) = 1 - \exp\left( -(v_{t+T,N}^{HV} - v_{t,N}^{HV}) \right)
\label{Eq20}
\end{equation}

This exponential formulation of the disturbance cost ensures that it increases with the deviation of the actual average velocity of the following vehicles from their desired velocity, thus encouraging the AV to behave in a way that minimizes this deviation. The desired velocity calculation of the following vehicles is closely tied to the current velocity of the leading AV. Thus, disturbance cost is closely related to the current velocity of the AV. For instance, if the throttle is depressed to the same degree, a lower velocity of the AV would result in a relatively smaller cost.

\textit{Remark 5:} Considering that in real-world traffic scenarios, the average velocity of the actual fleet will fluctuate up and down due to changing road situations, it's crucial to differentiate normal traffic flow variations from those induced by the AV's actions. To avoid unnecessary over-penalization from minor braking or slight steering adjustments triggering the disturbance cost, we've carefully incorporated an additional condition based on the AV's acceleration, $acc_t^{AV}$. The condition is introduced mainly to more accurately take into account the actual operation of the vehicle when deciding whether to calculate the disturbance cost. 

Specifically, the disturbance cost is calculated only when the AV's acceleration exceeds a predetermined threshold $\lambda$, effectively separating normal traffic variations from those significantly influenced by the AV. Mathematically, this additional condition can be represented as

\begin{equation}
C^{IM}(s_t, a_t^{AV}) =
\begin{cases}
C^{IM}, \quad & \text{if } acc_t^{AV} \geq \lambda \text{ and }acc_{t-1}^{AV} < \lambda \\
0, & \text{otherwise}
\end{cases}
\label{Eq21}
\end{equation}

To mitigate the traffic flow disturbance, we devise an implicit intervention value function $Q^{IM}(s, a^{AV})$ to capture the expected cumulative disturbance cost. This approach is similar to the method of approximating state-action values using the Bellman equation in Q-learning.
\begin{equation}
Q^{IM}(s_t, a_t^{AV}) = C^{IM}(s_t,a_t^{AV}) + \gamma \mathbb{E}_{s_{t+1}\sim \mathcal{B}, a_{t+1}\sim \pi_{AV}(\cdot | s_{t+1})}[Q^{IM}(s_{t+1},a_{t+1}^{AV})]
\label{Eq22}
\end{equation}

This value function can be directly used to optimize the policy. The process of computing the disturbance cost is outlined in pseudocode in Algorithm~\ref{algorithm1}.

\begin{algorithm}[!ht]
\caption{Computing Disturbance Cost based on Implicit Intervention Mechanism}
\label{algorithm1}
\DontPrintSemicolon
\KwData{State $s_t$, Action $a_t^{AV}$, IDM function, Duration $T$, Number of following HVs $N$, Threshold $\lambda$}
\KwResult{Disturbance cost $C^{IM}(s_t, a_t^{AV})$}
\While{Training is not finished}{
    
    \For{$i=1$ \KwTo $N$}{
        $v_{t,i}^{HV}  \gets$ Compute the current HVs' velocity following based on Eq~\ref{Eq7};
    }
    $v_{t,N}^{HV} \gets \frac{1}{N} \sum_{i=1}^N v_{t,i}^{HV}$\;

    \For{$i=1$ \KwTo $N$}{
        $v_{t+T,i}^{HV} \gets$ Compute the desire HVs' velocity based on Eq~\ref{Eq9};
    }
    $v_{t+T,N}^{HV} \gets \frac{1}{NT} \sum_{i=1}^N v_{t+T,i}^{HV}$\;

    Compute the velocity deviation: $\Delta v \gets v_{t+T,N}^{HV} - v_{t,N}^{HV}$\;
    
    \eIf{$acc_t^{AV} \geq \lambda \text{ and } acc_{t-1}^{AV} < \lambda$}{
        $C^{IM}(s_t, a_t^{AV}) \gets 1 - \exp\left( -\Delta v \right)$\;
    }{
        $C^{IM}(s_t, a_t^{AV}) \gets 0$\ Set disturbance cost to zero;
    }
}
\Return $C^{IM}(s_t, a_t^{AV})$\;
\end{algorithm}

\subsubsection{Reducing the Human Mentor's Cognitive Load}
The learning agent faces the risk of excessive dependence on human if the frequency of human takeover is not restricted, leading to failure when evaluating the agent without human supervision~\citep{peng2022safe,li2022efficient,wu2023toward}. This risk arises because $\hat{Q}(s_t,a_t)$ represents the proxy $Q$ value of the mixed behavior policy $\pi_{mix}$, rather than the AV's learning policy $\pi_{AV}$, owing to the presence of human takeover. In this case, there's a likelihood that the agent may select actions that go against human wishes, such as veering off the road when approaching the boundary. This behavior compels human to intervene and offer corrective demonstrations. In consequence, the AV's degree of automation remains low, leading to the human mentor experiencing fatigue due to the continual need to provide demonstrations.

\textbf{Definition 7 (Takeover Cost).} To mitigate human mentor's cognitive load and enhance the AV's level of automation, we apply a mild penalty to agent behaviors that trigger human explicit intervention. To achieve this, we employ the cosine similarity between the actions of the agent and those of a human as the takeover cost, which is formulated as follows~\citep{li2022efficient}:
\begin{equation}
C^{EX}(s_t, a_t^{AV}) = 1-\cfrac{{a_t^{AV}}^{\mathsf{T}}a_t^{human}}{||a_t^{AV}|| ||a_t^{human}||}, a_t^{human} \sim \pi_{human}(\cdot \mid s_t)
\label{Eq23}
\end{equation}

The agent incurs a significant penalty only when there's a substantial difference between the cosine similarity of its actions and those of humans.~\cite{li2022efficient} illustrated that employing this cosine similarity yields superior results compared to a fixed cost, such as using a `+1' penalty. Additionally, we attribute the takeover cost to the agent only during the initial step of human takeover. This approach is grounded in the observation that human takeover, triggered by a specific action $a^{AV}_t$, signify a deviation from human intentions at that particular moment. 

By reducing the occurrence of such actions, the level of automation of the agent can be increased, thus reducing the human mentor's cognitive load. To mitigate the abuse of human takeover, we introduce an explicit intervention value function, denoted as $Q^{EX}(s, a_t^{AV})$, which represents the expected cumulative cost of human takeover. This method parallels the technique of estimating state-action values via the Bellman equation in Q-learning.
\begin{equation}
Q^{EX}(s_t, a_t^{AV}) = C^{EX}(s_t,a_t^{AV}) + \gamma \mathbb{E}_{s_{t+1}\sim \mathcal B, a_{t+1}\sim \pi_{AV}(\cdot | s_{t+1})}[Q^{EX}(s_{t+1},a_{t+1}^{AV})]
\label{Eq24}
\end{equation}

The value function is employed to optimize the policy directly.

\begin{algorithm}[t]
\caption{HAIM-based Deep Reinforcement Learning (HAIM-DRL)}
\label{algorithm2}
\DontPrintSemicolon
\KwData{Policy network parameters $\theta$, Q-function parameters $\phi$, and replay buffer $\mathcal{B}$}
\While{Training is not finished}{
    \While{Episode is not terminated}{
        $a_t^{AV} \sim \pi_{AV}(\cdot | s_t)$  // Retrieve agent's action\;
        Execute $a_t^{AV}$ and observe new state $s_{t+1}$\;
        $I(s_t, a_t^{AV}) \gets$ Human expert determines whether to intervene by observing current state $s_t$\;
        \uIf{$I(s_t,a_t^{AV})$ is True}{
            $a_t^{human} \gets \pi_{human}(\cdot | s_t)$ // Retrieve human's action\;
            Execute $a_t^{human}$ within the environment\;
        }
        \Else{
            Execute $a_t^{AV}$ within the environment\;
        }
        \uIf{$I(s_t, a_t^{AV})$ is True and $I(s_{t-1}, a_{t-1}^{AV})$ is False}{
            $C^{EX}(s_t, a_t^{AV}) \gets$ Compute takeover cost following Eq. \ref{Eq23}\;
        }
        \Else{
            $C^{EX}(s_t, a_t^{AV}) \gets 0$ // Set takeover cost to zero\;
        }
        \uIf{$acc_t^{AV} \geq \lambda$ is True and $ acc_{t-1}^{AV} \geq \lambda$ is False}{
            $C^{IM}(s_t, a_t^{AV}) \gets$ Compute disturbance cost following Eq. \ref{Eq21}\;
        }
        \Else{
            $C^{IM}(s_t, a_t^{AV}) \gets 0$ // Set disturbance cost to zero\;
        }
        Record $s_t, a_t^{AV}$, $a_t^{human}$, $I(s_t, a_t^{AV})$ and $s_{t+1}$ to the buffer $\mathcal B$\;
    }
    Update proxy $Q$ value, explicit intervention value $Q^{EX}$, implicit intervention value $Q^{IM}$, and policy $\pi$ according to Eq.\ref{Eq19}, Eq. \ref{Eq22}, Eq. \ref{Eq24}, Eq. \ref{Eq25}, respectively.\;
}
\end{algorithm}

\subsection{Learning Policy for Policy Network}
The policy network is responsible for determining control actions and strives to optimize the value network. In the `$X+1+N$’ scenario discussed in this study, we use the policy network to enhance the comprehensive performance of autonomous driving, focusing on safety and efficiency. Considering Eqs.~\ref{Eq14} and \ref{Eq15}, the batch gradient of the policy network can be expressed as follows:
\begin{equation}
\begin{aligned}
\max _\theta \underset{s_t \sim \mathcal{B}}{\mathbb{E}} \left[\psi \hat Q\left(s_t, a_t^{AV}\right)-\alpha \log \pi_{AV}\left(a_t^{AV} \mid s_t ; \theta\right) \right. \left. - \beta Q^{EX}\left(s_t, a_t^{AV}\right)- \varphi Q^{IM}\left(s_t, a_t^{AV}\right)\right]
\end{aligned}
\label{Eq25}
\end{equation}

The entropy regularization coefficient $\alpha$ enhances the policy by encouraging a balance between exploitation and exploration. Meanwhile, the coefficients $\psi$ and $\varphi$ serve as weighting factors, allowing for a controlled trade-off between ensuring the agent's safe driving and optimizing traffic flow efficiency. When safety is a paramount concern, the weight of $\psi$ can be increased, reflecting a greater emphasis on aligning the agent's actions with human intentions. In scenarios where traffic flow efficiency is prioritized, increasing the weight of $\varphi$ can be advantageous. A higher $\varphi$ makes the agent more sensitive to its impact on traffic flow. On the other hand,  A higher $\beta$ encourages the agent to develop policies that are less reliant on human takeover. During the learning process, these factors can be adjusted to prioritize the AV's operational objectives within mixed traffic conditions.  

The overall workflow of the HAIM-DRL framework is shown in Algorithm \ref{algorithm2} as pseudocode.

\section{Experimental Evaluation}
\subsection{Experimental Setting}
\subsubsection{Experimental Scenarios} 

Recalling our research goals, we aim to develop a robust driving policy within the context of the `$X+1+N$’ mixed traffic platoon. This policy seeks not only to ensure the safety of AVs amidst environmental uncertainties but also to reduce traffic flow disturbance. Specifically, we are most interested in four aspects of performance: (a) \textbf{Safety.} This encompasses measures to avoid obstacles during both the training and testing phases, as well as ensuring the AV reaches its destination; (b) \textbf{Generalization ability.} This involves evaluating the agents on new maps not encountered during training. The focus is to verify their ability to adapt and navigate safely in unfamiliar environments; (c) \textbf{Sampling efficiency.} This refers to the training time required for the method to exhibit satisfactory performance; (d) \textbf{Minimizing traffic flow disturbance.} This refers to ensure that the AV's actions do not cause unnecessary delays on other vehicles in the traffic system.

\begin{figure*}[!ht]
  \centering
  \includegraphics[width=0.95\textwidth]{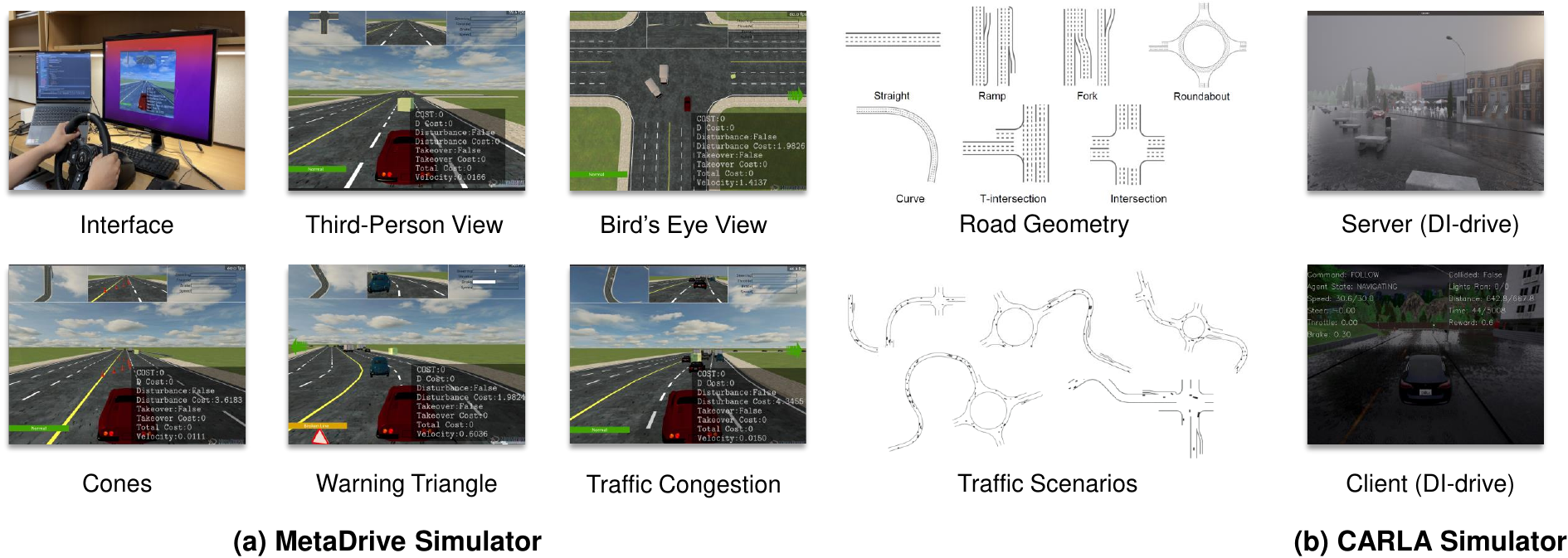}
  \caption{Experimental scenes in MetaDrive/CARLA simulator.}\label{fig5}
\end{figure*}

Due to the potential risks associated with involving human experts in physical experiments, we conducted benchmarking using a driving simulator. Specifically, we employed the latest lightweight driving simulator, MetaDrive~\citep{li2022metadrive}, which enables the evaluation of safety and generalizability in unknown environments. MetaDrive is implemented using Panda3D and the Bullet Engine, providing efficient and accurate physics-based 3D dynamics. Some generated driving scenarios are depicted in Fig.~\ref{fig5} (a). While the MetaDrive setup is primarily described in this section, experiments were also conducted using the CARLA simulator \citep{dosovitskiy2017carla} to verify the transferability of the proposed method. For interaction with the CARLA core, we utilize the CARLA client wrapper implemented in DI-drive~\citep{didrive}, as depicted in Fig.~\ref{fig5} (b). 

To sufficiently encompass dangerous and long-tail scenarios, we used MetaDrive to generate a diverse array of driving scenarios, each training session comprising 50 different scenarios. As depicted in Fig. \ref{fig5} (a), this encompasses road scenarios made up of various typical block types such as straight, ramp, fork, roundabout, curve, T-intersection, and intersection. Surrounding vehicles are controlled using two distinct models: the IDM for longitudinal dynamics, and the minimizing overall braking induced by lane change model (MOBIL) \citep{kesting2007general} for lateral movement. Furthermore, we introduced traffic flows at varying densities to interact with the AV. To enhance the complexity of the driving task, each scene included randomly placed obstacles, such as moving traffic vehicles, stationary traffic cones, and triangular warning signs.

\subsubsection{User Interface Reconfiguration}

In the simulator, the agent's task is to manipulate the leading AV using acceleration, braking, and steering to reach a predetermined destination and obtain a `\textit{Success}' flag. The `\textit{Success Rate}' is defined as the proportion of episodes in which the agent successfully reaches the destination. In addition, as shown in Fig.~\ref{fig5}, we also defined the `\textit{Takeover}' and `\textit{Disturbance}' flags in the simulator. The `\textit{Takeover}' flag is triggered when human experts takeover. The `\textit{Disturbance}' flag indicates that the acceleration of the current action exceeds the predetermined threshold $\lambda$,  causing a disturbance in the following traffic flow. 

When a takeover action or abrupt braking is detected, we begin to calculate the corresponding `\textit{Cost}' and `\textit{Disturbance Cost}'. The total takeover cost and total disturbance cost represent the sum of all individual takeover cost and disturbance cost within the event, labeled as `\textit{Takeover Cost}' and `\textit{D Cost}'. As described in Eq.~\ref{Eq20}, the calculation of disturbance cost is closely related to the current velocity of the agent. Therefore, we display the real-time velocity information on the user interface, labeled as `\textit{Velocity}'.

\subsubsection{Observation, Reward, Cost, and Generalization Ability}

In this subsection, we define the observation space, the environmental reward, the environmental cost, and the generalization ability indicator.

\textbf{Observation Space.} we use the numerical observation space of Fig.~\ref{fig4} (a), which comprises: (a) the current state, including elements such as steering, heading, velocity, and the relative distance to the boundary; (b) navigation data that guides the vehicle towards its intended destination; (c) information about the surroundings, represented by a vector of 240 LiDAR-like distance measurements from nearby vehicles. 

\textbf{Environmental Reward.} While HAIM-DRL does not rely on environmental reward during its training phase, we provide a reward function for training baseline methods and for evaluating HAIM-DRL during testing. This reward function encompasses a dense driving reward, a velocity reward, and a sparse terminal reward. The driving reward evaluates the longitudinal movement towards the destination. Additionally, the agent is rewarded based on its velocity, with a sparse reward of `+20' issued upon reaching the destination.

\textbf{Environmental Cost.} Each collision with traffic vehicles or obstacles results in an environmental cost of `+1'. Note that HAIM-DRL does not have access to this cost during training. This cost is utilized for training safe RL methods and for testing the safety of trained policies. Detailed definitions of reward and costs can be found in \cite{li2022metadrive}. The episodic cost, referred to as `\textit{Safety Violation}', serves as an indicator of a policy’s safety level.

\textbf{Generalization Ability.} Success rate is a more appropriate criterion for assessing generalization compared to episodic reward. This is because the evaluated scenes exhibit diverse attributes, such as road length and traffic density, which result in significant reward fluctuations across different scenes \citep{li2022metadrive}.

\subsubsection{Implementation Details}

As emphasized in Assumption 1, the HAIM-DRL framework requires only a single experienced human expert for implementation. In this experiment, as shown in Figs.~\ref{fig5} (a), a human participant with both United States and Chinese driving licenses and 7 years of driving experience was invited to supervise the learning agent's real-time exploration with hands on the steering wheel. Upon perceiving an impending dangerous situation, the human takes over by pressing the paddle beside the wheel, thereby gaining control of the vehicle through steering and pedal manipulation. Additionally, we assume that there are $N$ HVs following the AV agent, driving according to the IDM. Too large an `$N$’ can complicate traffic management, whereas too small an `$N$’ might oversimplify real-world traffic dynamics. Similarly, a very high `$X$’ can overwhelm the AV with unpredictable elements, while a low `$X$’ may not adequately test its adaptability. In our study, the `$N$’ value aligns with \cite{shi2023deep}, and `$X$’ with \cite{li2022efficient}. The \ref{Appendix:Hyper-parameters} describes the detailed hyperparameter settings used in this experiment.

\begin{table*}[t]
\centering
\begin{small}
\caption{The performance of different baselines in the MetaDrive simulator.}
\label{tab1}
\resizebox{\textwidth}{!}{%
\begin{tabular}{@{}ccccccccc@{}}
\toprule
Category & Method & \shortstack{Total Training \\ Safety Violation} & \shortstack{Training\\Data Usage} & \shortstack {Test\\Return} &  \shortstack{{Test Safety}\\ {Violation}} & \shortstack{Disturbance \\Rate} & \shortstack{Test\\Success Rate}  \\
\toprule
\multirow{1}*{\shortstack{\textit{Expert}}} &
\textit{Human} & -&-	& \textit{388.16 {\tiny $\pm$45.00}} & \textit{{0.03\tiny $\pm$0.00}} & 0 & \textit{1} \\
\midrule
\multirow{2}*{\shortstack{RL}} 
& SAC-RS \citep{haarnoja2018soft} &  2.78K {\tiny $\pm$ 0.97K }  &	1M & \textbf{384.56} 	{\tiny $\pm$37.5} &	0.87 	{\tiny $\pm$1.47} &	0.015 {\tiny $\pm$ 0.005} & 0.83 	{\tiny $\pm$0.32} \\ 
& PPO-RS \citep{schulman2017proximal} & 27.51K {\tiny $\pm$3.86K} & 1M& 305.41 {\tiny $\pm$14.23}  &	4.12 {\tiny $\pm$1.24}  &	0.021 {\tiny $\pm$ 0.014} & 0.67{\tiny $\pm$0.12}   \\ \midrule
\multirow{3}*{\shortstack{Safe RL}}
& SAC-Lag \citep{ha2021learning}& 1.98K {\tiny $\pm$ 0.75K} &	1M & 352.46	{\tiny $\pm$108.78} &	0.78 	{\tiny $\pm$0.58} &	0.019 {\tiny $\pm$ 0.007} &0.71 	{\tiny $\pm$0.79} \\
& PPO-Lag \citep{stooke2020responsive} &15.46K {\tiny $\pm$ 5.13K} &	1M & 298.98 	{\tiny $\pm$50.99} &	3.28 {\tiny $\pm$0.38} &	0.025 {\tiny $\pm$ 0.016} & 0.52	{\tiny $\pm$0.27} \\
& CPO \citep{achiam2017constrained} &
4.36K  {\tiny $\pm$2.22K}
& 1M & 194.06 {\tiny $\pm$108.86} & 1.71 {\tiny $\pm$1.02} & - & 0.21 {\tiny $\pm$0.29} \\
\midrule
\multirow{1}*{\shortstack{Offline RL}}
& CQL \citep{kumar2020conservative}& -  &	49K&  116.45	{\tiny $\pm$34.94} &3.68	{\tiny $\pm$7.61} & 0.007 {\tiny $\pm$ 0.0006} &	0.13 	{\tiny $\pm$0.09} \\\midrule
\multirow{2}*{\shortstack{IL}}
& BC \citep{bain1995framework} & - & 49K	& 36.13	{\tiny $\pm$10.66} &		1.05 	{\tiny $\pm$0.54} &	0.012 {\tiny $\pm$ 0.0170} & 0.01 	{\tiny $\pm$0.02} \\
& GAIL \citep{ho2016generative} &
3.68K {\tiny $\pm$3.17K}&	49K &108.36 	{\tiny $\pm$16.08} &	4.18 	{\tiny $\pm$1.25} &	0.001 {\tiny $\pm$ 0.0009} & 0.03 	{\tiny $\pm$0.01} \\
\midrule

\multirow{2}*{\shortstack{Conventional Human-in-the-loop}}
& HG-DAgger \citep{kelly2019hg} & 35.58 & 50K	& 106.21 &		2.63  &	0.108 & 0.04\\
& IWR \citep{mandlekar2020human} & 69.74 & 50K	& 298.87 & 3.61  & 0.122 &0.61\\
\midrule

\multirow{1}*{\shortstack{Human-AI Copilot}}
& HACO \citep{li2022efficient} &
30.05 {\tiny $\pm$10.89}&	30K & 350.01	{\tiny $\pm$9.72} &	0.78 	{\tiny $\pm$0.85} &	0.038 {\tiny $\pm$ 0.0083} & 0.83 	{\tiny $\pm$0.07} \\
\midrule

\multirow{1}*{\shortstack{{Human as AI Mentor}}} & \textbf{HAIM-DRL (Ours)}	& 
\textbf{29.84} {\tiny $\pm$ 10.25}
& \textbf{30K}\textsuperscript{*} & { 354.34 {\tiny $\pm$ 11.08 }} & \textbf{0.76}	{\tiny $\pm$ 0.28 } &	\textbf{0.0023}	{\tiny $\pm$ 0.00072 } & { \textbf{0.85} 	{\tiny $\pm$ 0.03}} \\
\bottomrule
\end{tabular}%
}
\end{small}
\begin{flushleft}
\textsuperscript{*} 
Throughout the HAIM-DRL training process, the human expert intervenes and overrides the agent's actions in approximately 7759 {\tiny $\pm$ 462.35} steps out of the total 30K steps. The entire training process spans roughly 65 minutes. In contrast, RL and safe RL methods, such as SAC-RS, PPO-RS, SAC-Lag, and PPO-Lag, require approximately 42 hours for training. Offline RL and IL methods, including CQL, BC, and GAIL, take around 80 hours for training.
\end{flushleft}
\vspace{-1em}
\end{table*}

\subsubsection{Training Environment}
We divided the generated driving scenarios into two sets: a training set and a test set. The AV agent was exclusively trained on the training set and subsequently evaluated on the reserved test set. At the outset of each episode, we randomly selected a scenario from either the training set or the test set. Following each training iteration, we deployed the learned agent into the test environment without any human supervision. Experiments are carried out using the MetaDrive/CARLA simulator, and methods are implemented with RLLib~\citep{liang2018rllib}, an efficient distributed learning system. 

The training of baseline methods is performed through eight concurrent trials on two Nvidia GeForce RTX 4090 GPUs, each equipped with 24GB of memory. Each trial utilizes 2 CPUs with 8 parallel rollout workers. Except for those involving human-in-the-loop, all baseline experiments are repeated five times using different random seeds. The primary experiments for HAIM-DRL and HACO are conducted on a local computer equipped with an Nvidia GeForce RTX 3060 with 6GB of memory, and these are repeated three times. Owing to restricted human resources, both the ablation studies and the human-in-the-loop baseline experiments are conducted only once.

\subsection{Comparative Results in the MetaDrive Simulator}
In the comparative analysis, each table and each figure report the standard deviation of multiple replicated experiments, each using a different randomized seed. The data presented in Table \ref{tab1} represent the average of the maximum values observed at the last checkpoint of each evaluation, following the protocol of repeating the experiments five times to ensure the robustness and reliability of the results.

\begin{figure*}[!ht]
  \centering
  \subfloat[]{\includegraphics[width=0.33\textwidth, height=4.65cm]{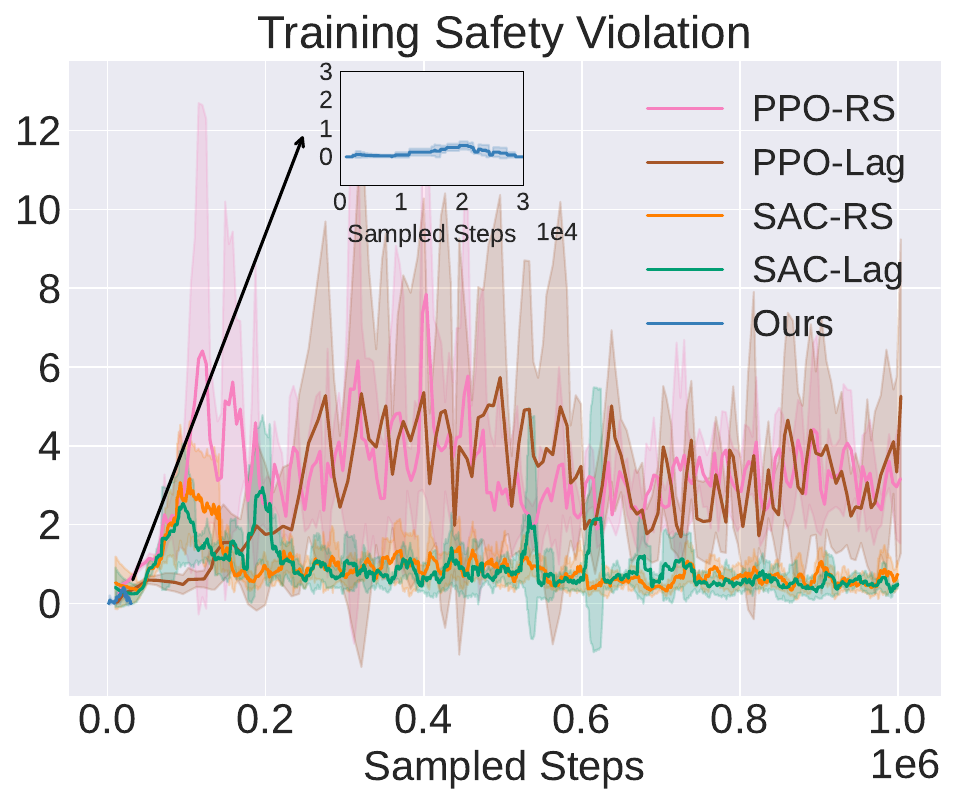}}
  \hfill
  \subfloat[]{\includegraphics[width=0.33\textwidth, height=4.65cm]{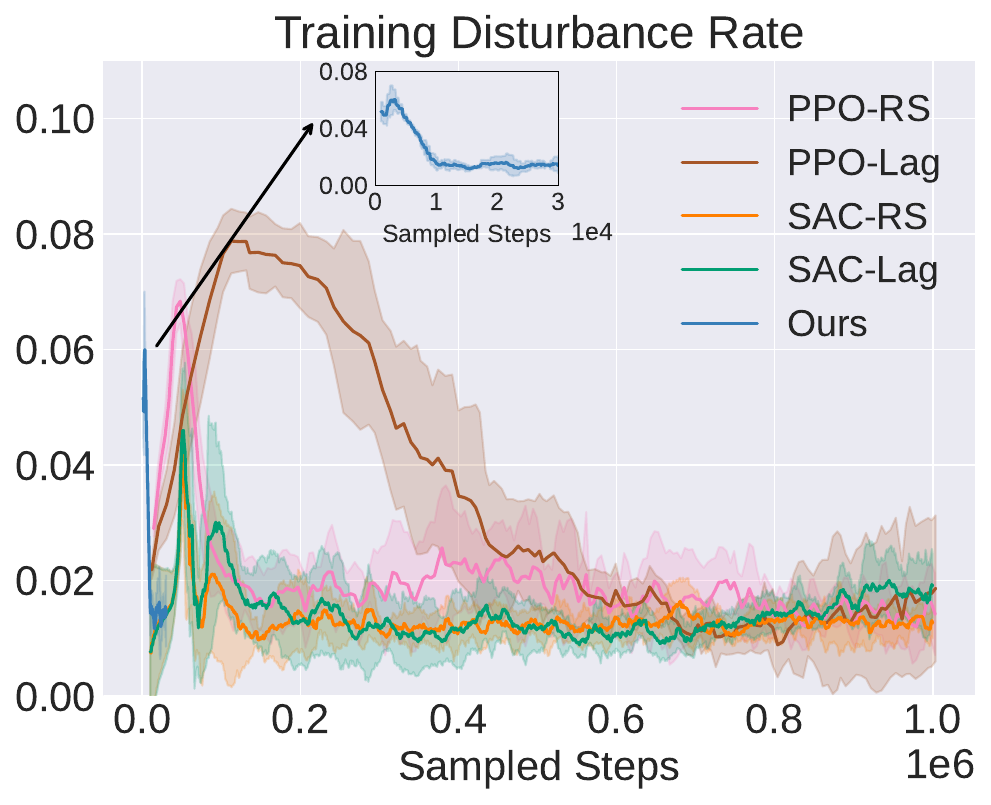}}
  \hfill
  \subfloat[]{\includegraphics[width=0.33\textwidth, height=4.65cm]{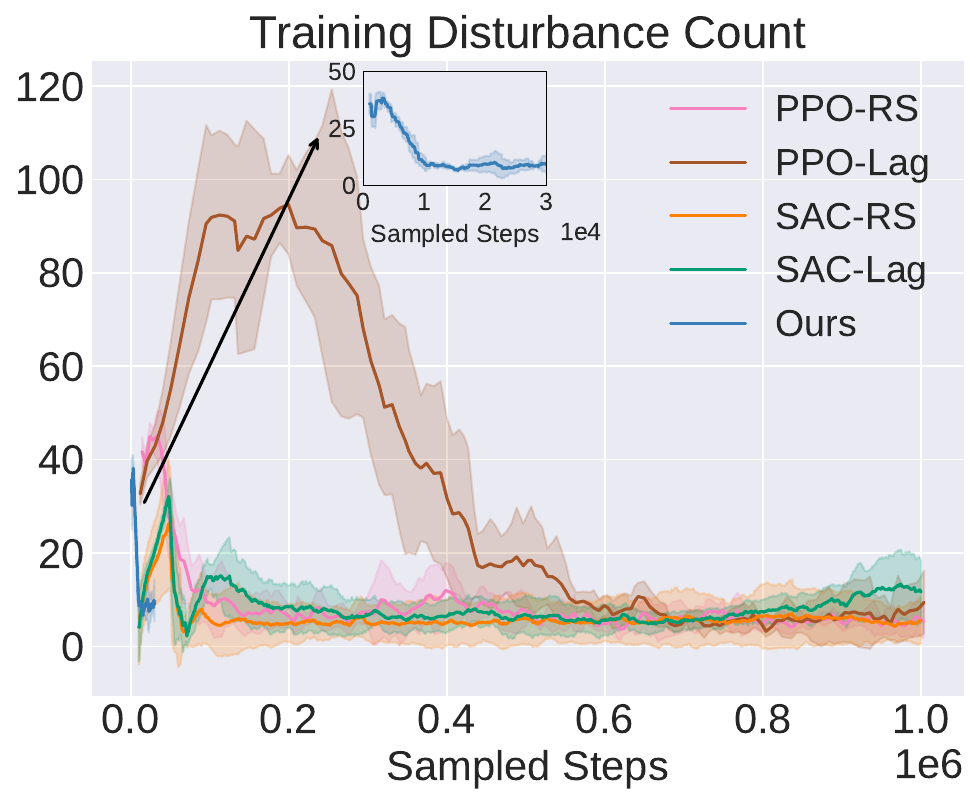}}
  \newline
  \subfloat[]{\includegraphics[width=0.33\textwidth, height=4.65cm]{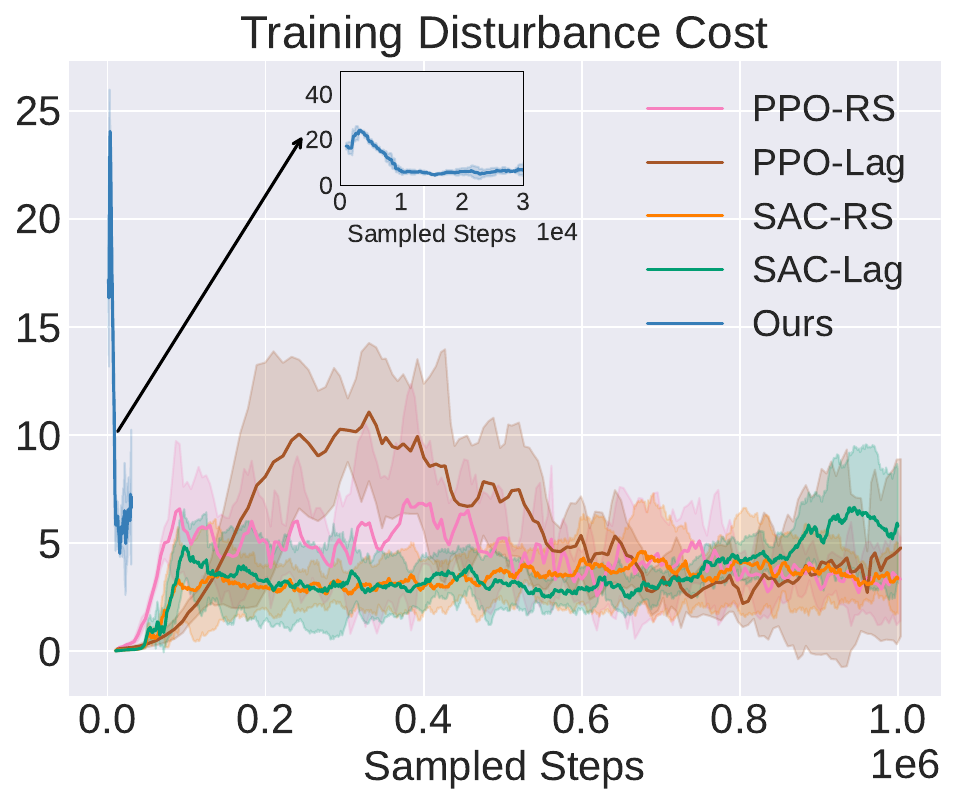}}
  \hfill
  \subfloat[]{\includegraphics[width=0.33\textwidth, height=4.65cm]{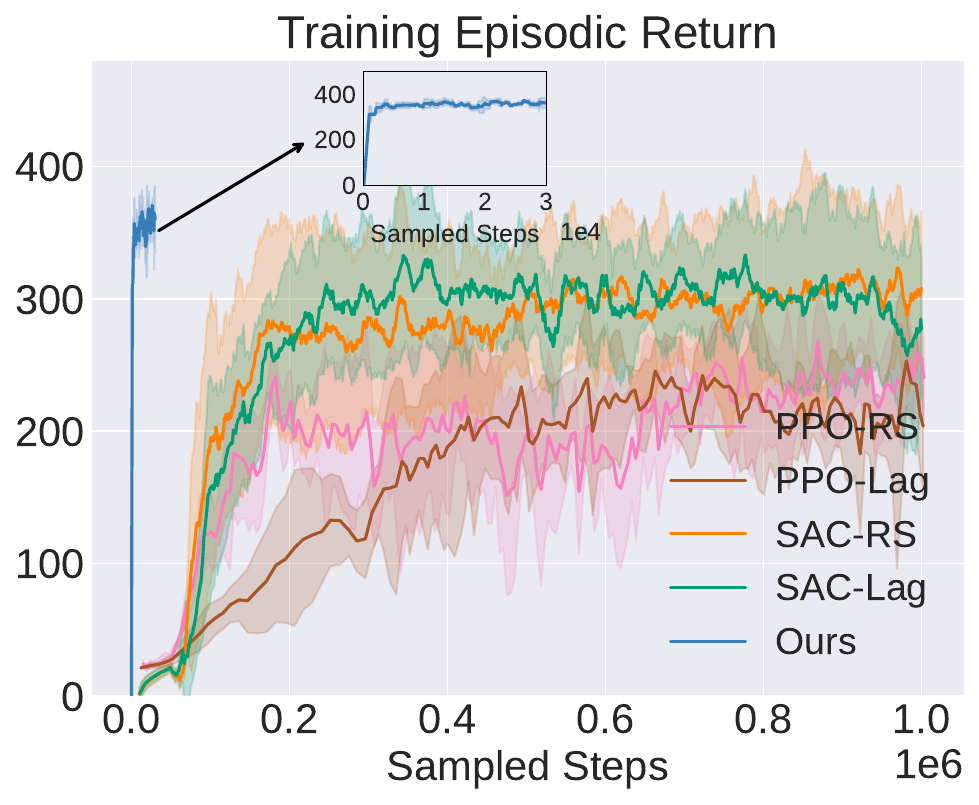}}
  \hfill
  \subfloat[]{\includegraphics[width=0.33\textwidth, height=4.65cm]{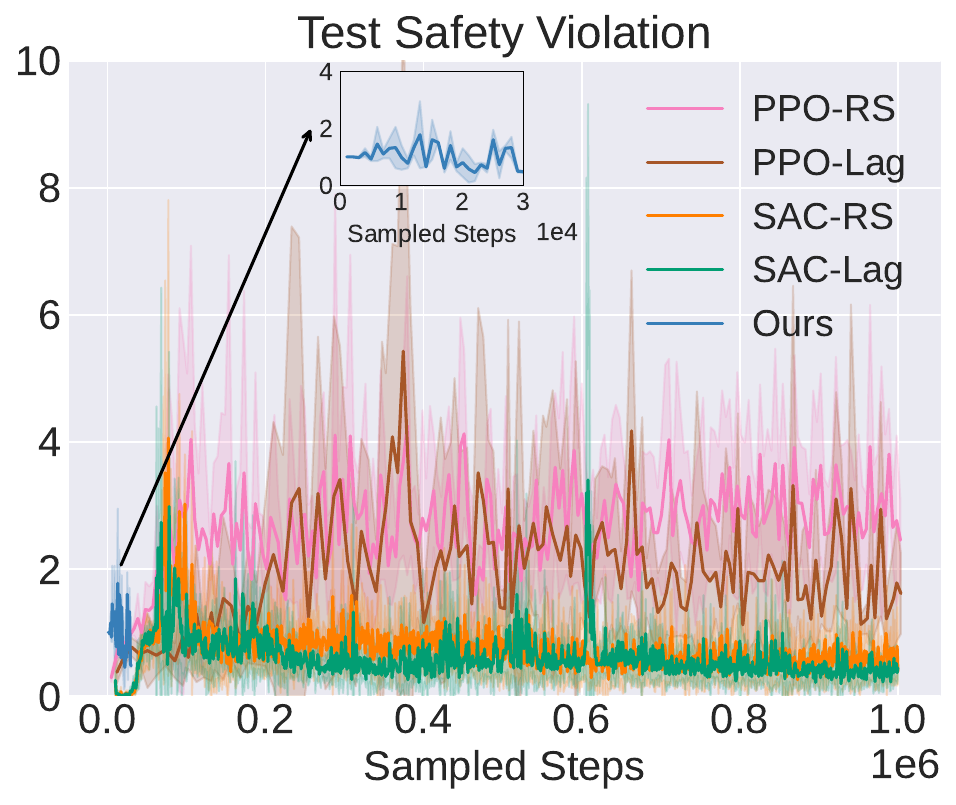}}
  \newline
  \subfloat[]{\includegraphics[width=0.33\textwidth, height=4.65cm]{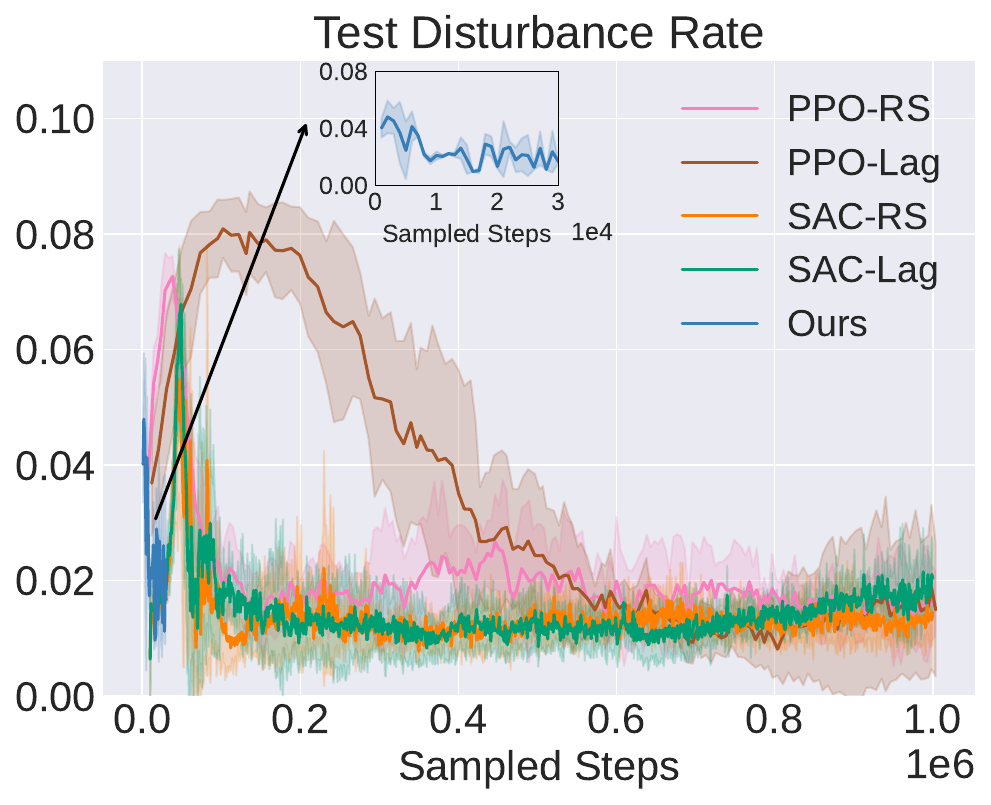}}
  \hfill
  \subfloat[]{\includegraphics[width=0.33\textwidth, height=4.65cm]{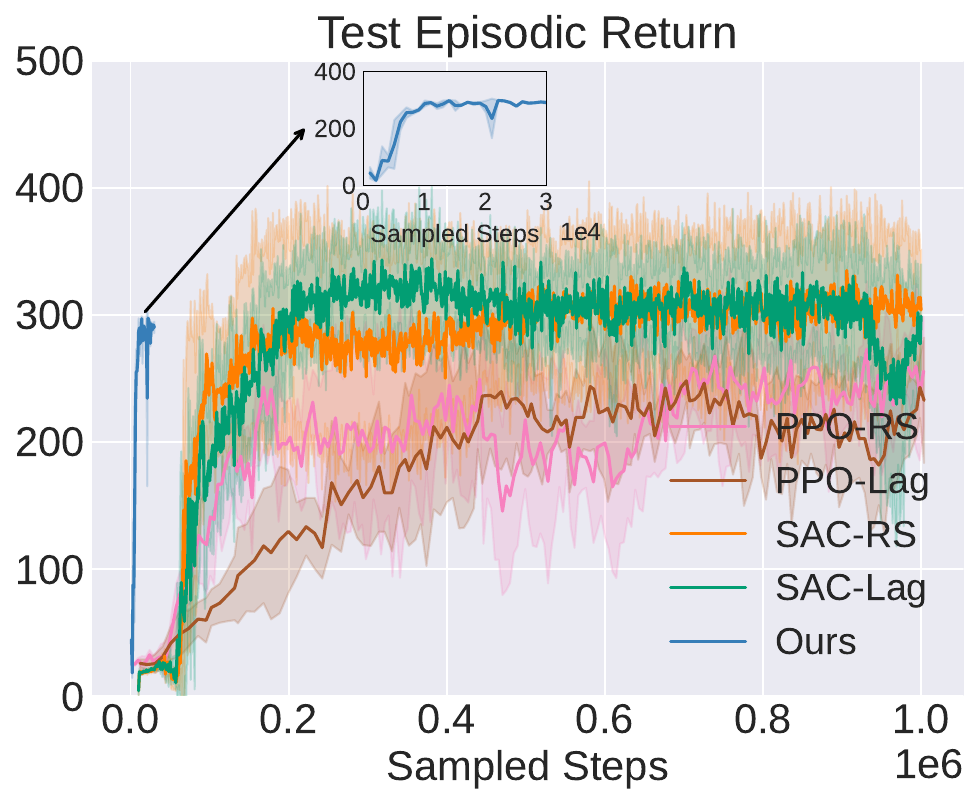}}
  \hfill
  \subfloat[]{\includegraphics[width=0.33\textwidth, height=4.65cm]{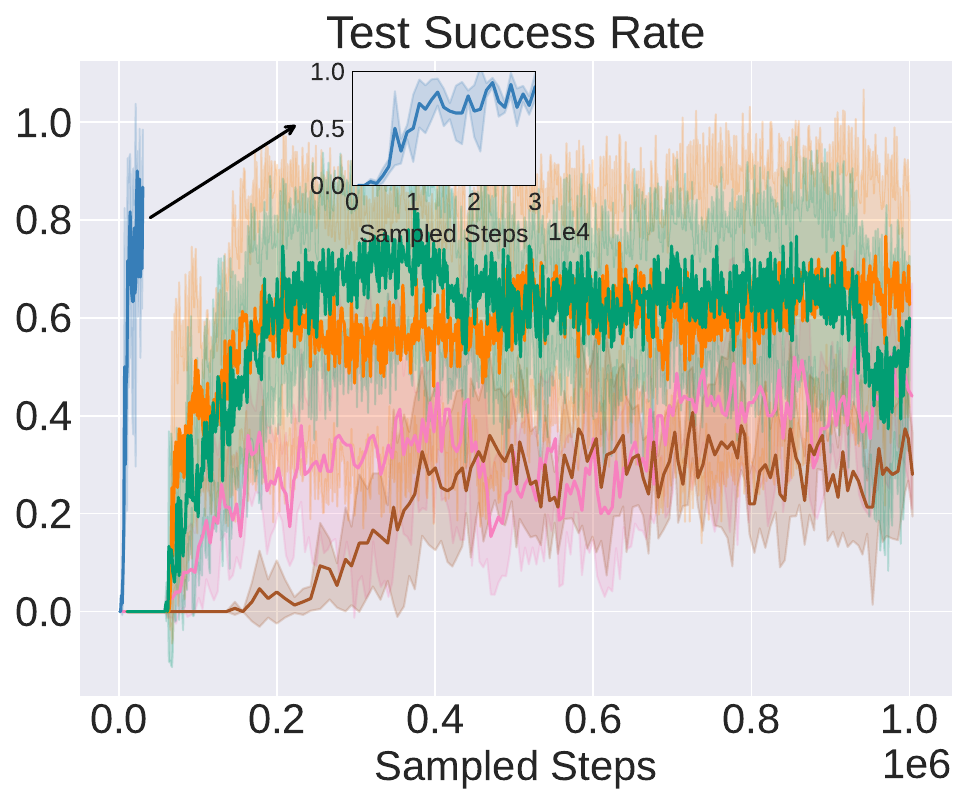}}
  \caption{Performance comparison of HAIM-DRL with traditional and safe RL methods.}
  \label{fig6}
\end{figure*}

\subsubsection{Compared to RL and Safe RL Methods}
The HAIM-DRL was meticulously evaluated against both traditional RL and Safe RL methods. Traditional RL methods such as SAC~\citep{haarnoja2018soft} and PPO~\citep{schulman2017proximal} were evaluated, with cost applied as an additional negative reward in a process known as reward shaping. In the realm of Safe RL, we scrutinized established benchmarks, including CPO~\citep{achiam2017constrained}, SAC-Lag~\citep{ha2021learning}, and PPO-Lag~\citep{stooke2020responsive}. The comparative data, as illustrated in Fig.~\ref{fig6} and Table~\ref{tab1}, demonstrate that our HAIM-DRL method surpasses these traditional and safety-conscious methods in crucial performance metrics. Notably, due to its underwhelming safety performance, CPO's results are omitted from Fig.~\ref{fig6} to maintain the focus on more successful methods. 

Safety is our primary concern, and HAIM-DRL significantly reduces safety violations. It records a mere 29.84 safety violation during the entire training phase. This figure is significantly lower, by two orders of magnitude, compared to other RL methods, despite not incorporating environmental cost. We find that the RS techniques in SAC and PPO variants are less effective than the Lagrangian methods. Furthermore, PPO exhibits more violation than SAC, which might be attributable to its relatively lower sampling efficiency and slower convergence rate. Regarding generalization ability, HAIM-DRL achieves a notable test success rate of 0.85 after with a safety violation rate of just 0.76. This achievement not only showcases its potential for navigating to destinations in new and uncertain driving environments but also ensures the AV's safety amid these environmental uncertainties. This result is pivotal as it has the potential to increase public trust in AV technologies.

In terms of sampling efficiency, HAIM-DRL demonstrate rapid convergence, achieving a test return exceeding 300 in just 10K iterations—a notable contrast to the 200K iterations required by other methods. Impressively, HAIM-DRL attains a test success rate of 0.85 with only 30K environmental interactions, of which a mere 7,759 are safe operation steps provided by human demonstrators. During approximately 65 minutes of human-AI cooperative driving, steps contributed by humans constitute only 25.86\% of the interactions. This high efficiency shows that HAIM-DRL not only benefits from effective human guidance but also achieves greater sampling efficiency with a limited number of interaction steps. Furthermore, the disturbance rate, indicative of the smoothness of traffic flow, is minimized by HAIM-DRL to less than 3\%. This efficiency is particularly notable given the limited number of human demonstration steps required, suggesting that HAIM-DRL can foster smoother traffic with minimal human takeover. This aligns with our goal to minimize traffic flow disturbance and enhance overall traffic flow efficiency.

\subsubsection{Compared to IL and Offline RL Methods}

\begin{figure*}[t]
  \centering
  \subfloat[]{\includegraphics[width=0.245\textwidth, height=3.85cm]{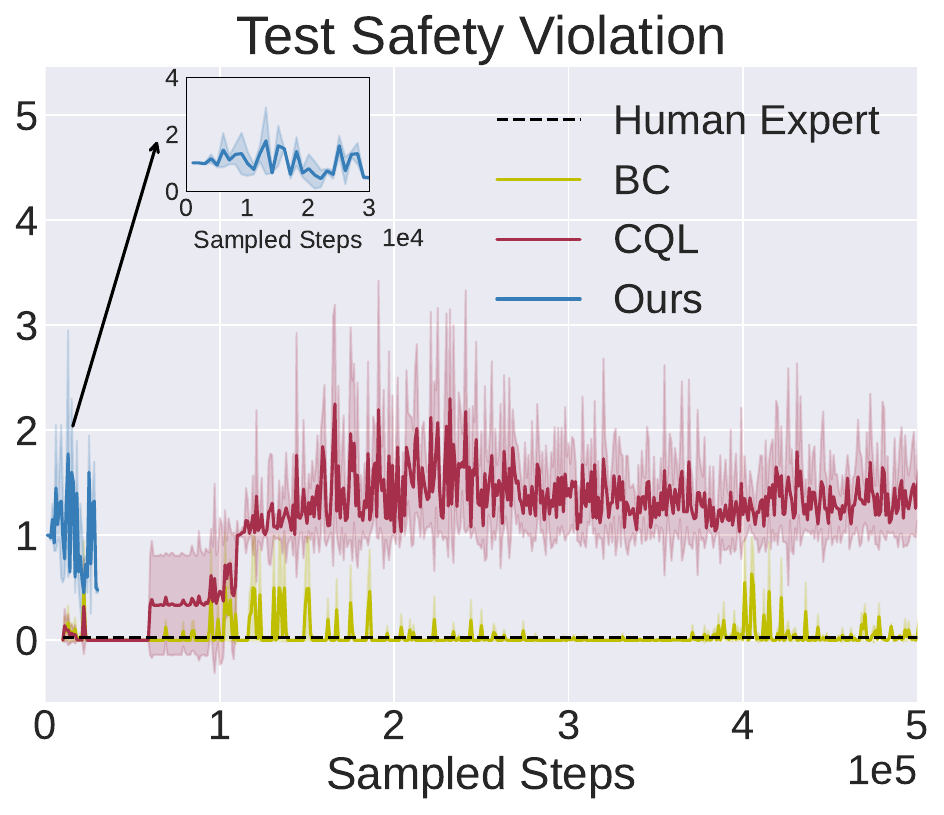}}
  \hfill
  \subfloat[]{\includegraphics[width=0.245\textwidth, height=3.85cm]{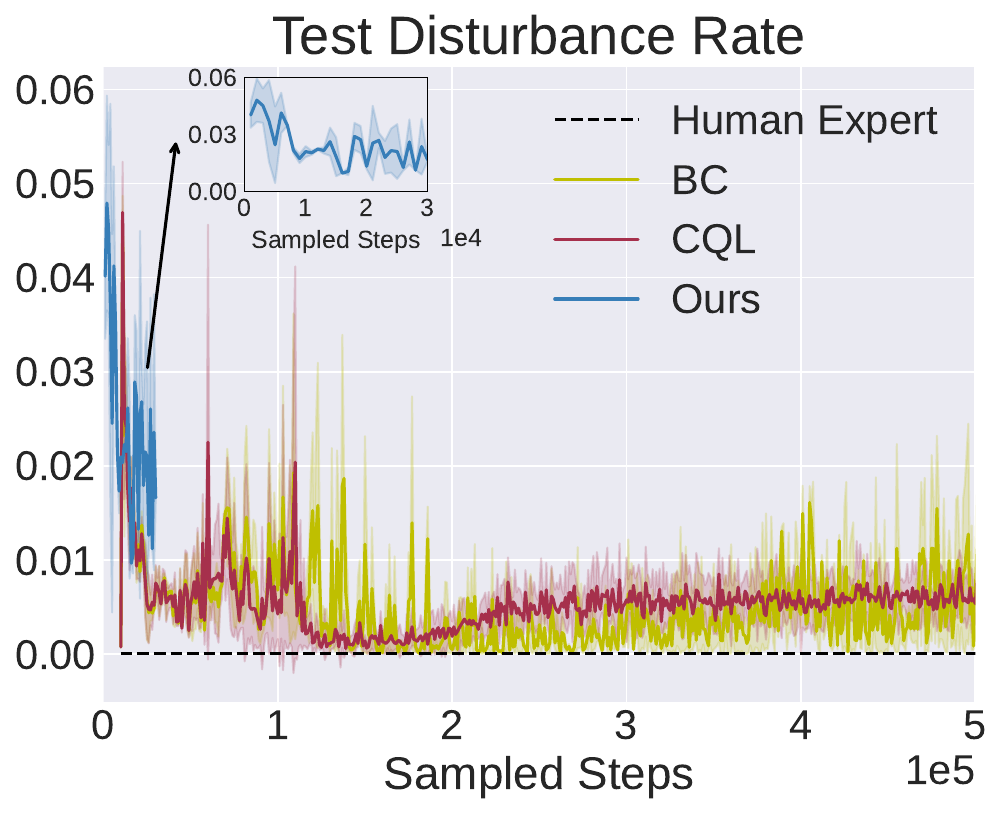}}
  \hfill
  \subfloat[]{\includegraphics[width=0.245\textwidth, height=3.85cm]{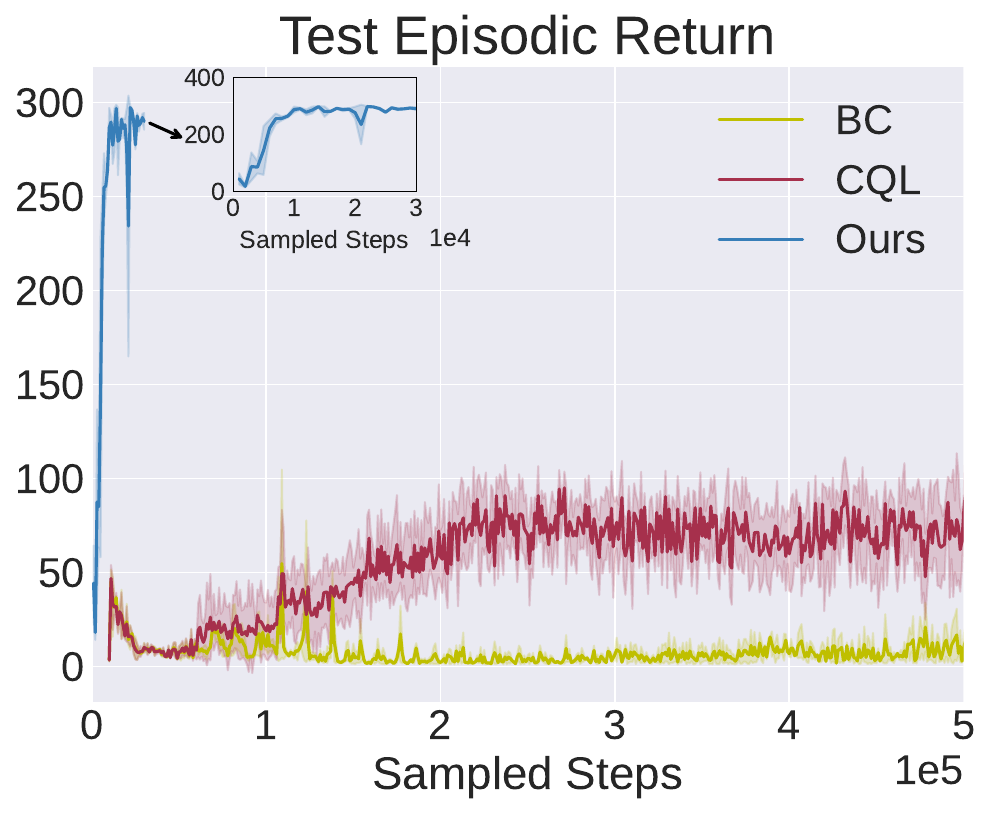}}
  \hfill
  \subfloat[]{\includegraphics[width=0.245\textwidth, height=3.85cm]{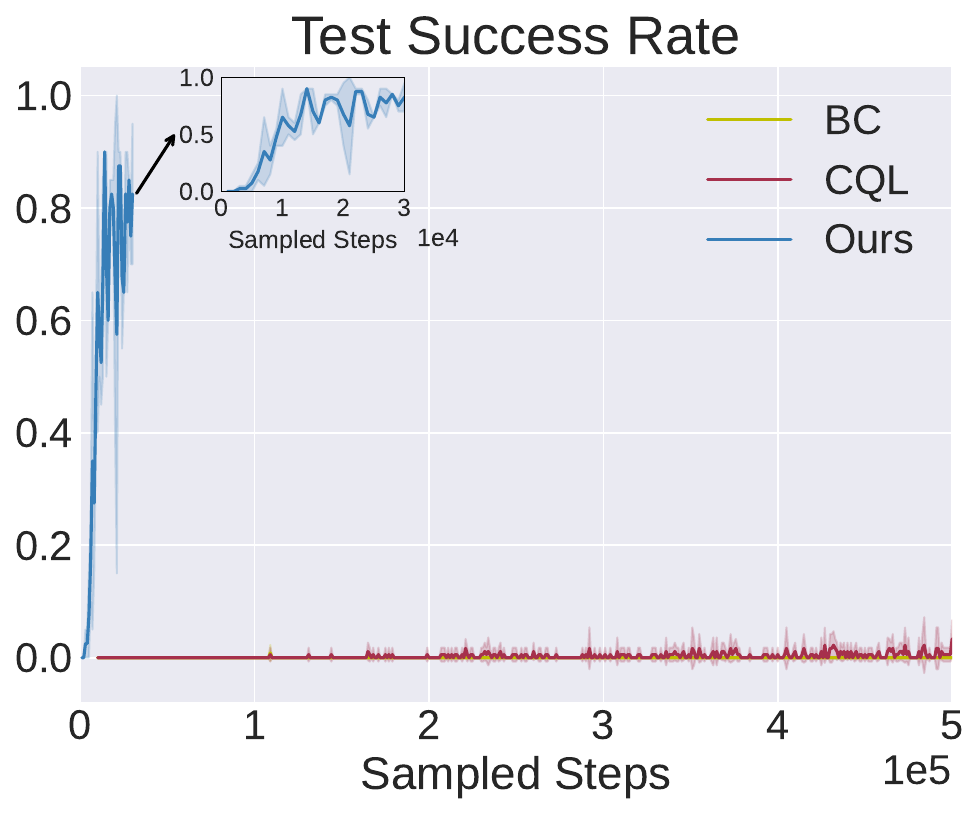}}
  \newline
  \subfloat[]{\includegraphics[width=0.245\textwidth, height=3.85cm]{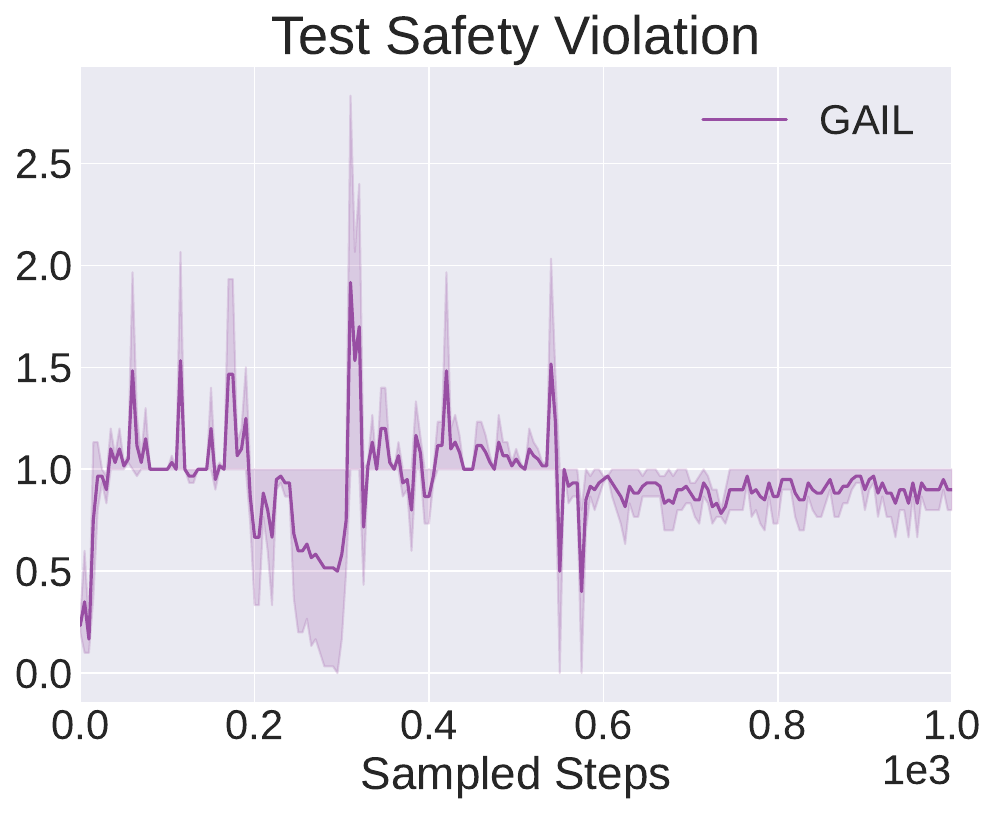}}
  \hfill
  \subfloat[]{\includegraphics[width=0.245\textwidth, height=3.85cm]{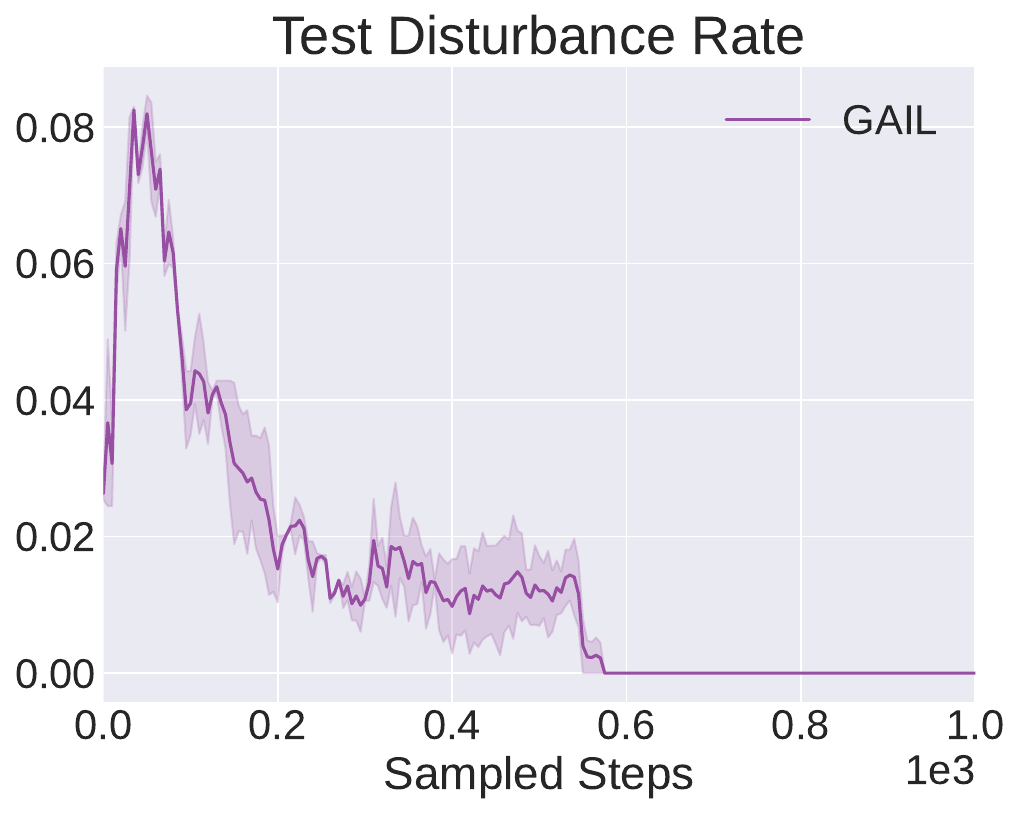}}
  \hfill
  \subfloat[]{\includegraphics[width=0.245\textwidth, height=3.85cm]{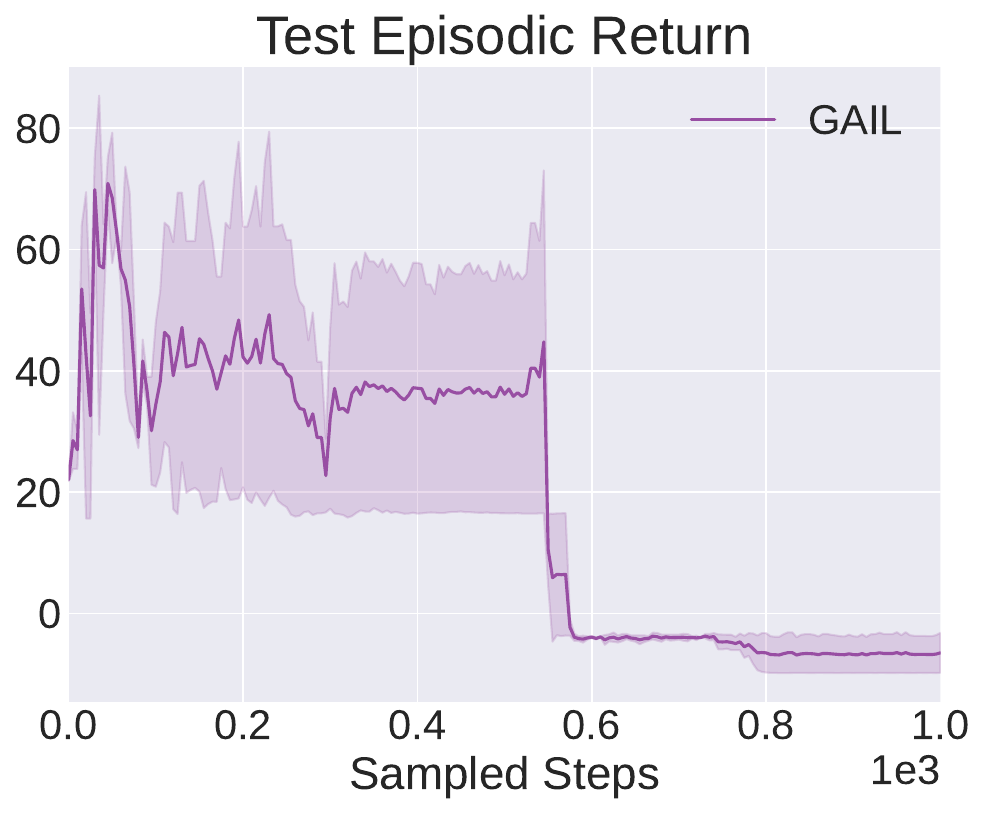}}
  \hfill
  \subfloat[]{\includegraphics[width=0.245\textwidth, height=3.85cm]{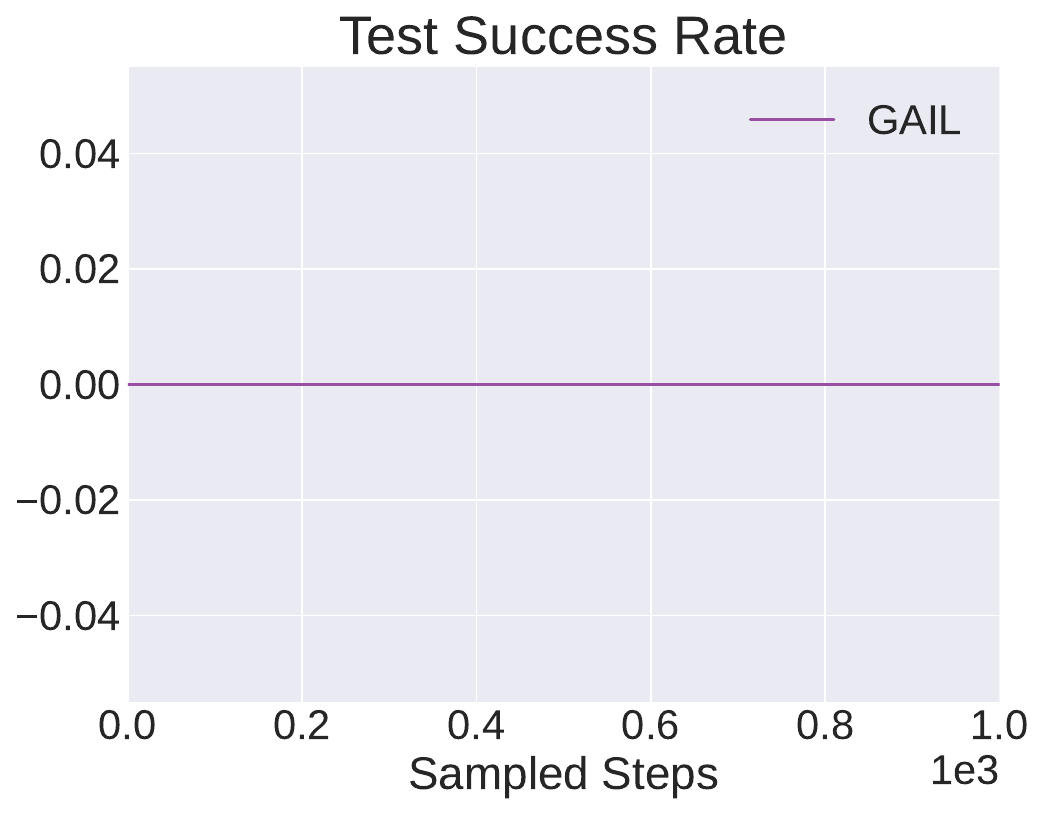}}
  \caption{Performance comparison of HAIM-DRL with IL/offline RL methods.}
  \label{fig7}
\end{figure*}

We have collected a human demonstration dataset consisting of an hour’s worth of expert demonstrations. This collection translates to about 49K transitions within the training environments \footnote{The high-quality demonstration dataset collected by the human expert in this study is available at: \href{https://github.com/zilin-huang/HAIM-DRL/releases/tag/v0.0.0}{https://github.com/zilin-huang/HAIM-DRL/releases/tag/v0.0.0}}. This high-quality demonstration dataset boasts a perfect success rate of 100\%, with only three instances of disturbance and a notably low safety violation rate of 0.03, thereby establishing a benchmark for our evaluations. Leveraging this dataset, we evaluated passive IL methods such as BC \citep{bain1995framework}, active IL methods such as GAIL \citep{ho2016generative}, and offline RL methods such as CQL \citep{kumar2020conservative}.

Fig.~\ref{fig7} illustrates the detailed training processes of BC/CQL up to 100K steps and GAIL within the first 1000 steps, with Table~\ref{tab1} presenting the experimental results at 1M and 3000 steps, respectively. As shown in Fig.~\ref{fig7} and Table~\ref{tab1}, HAIM-DRL exhibits a pronounced advantage in test success rate over the IL and offline RL methods. Different from the success with HAIM-DRL, IL and offline RL methods learning from demonstrations fail on the dataset containing 49K transitions. Notably, the GAIL has a success rate of 0 during the initial 1000 steps. This is because GAIL is designed to match the distribution of the expert data, and if the expert data does not cover all the scenarios encountered during testing, it can lead to failure in these tests. Furthermore, compared to the high test episodic return achieved by HAIM-DRL, the test episodic returns of BC, CQL, and GAIL are extremely low.

In terms of safety, the number of safety violation by the CQL is significantly higher than HAIM-DRL, showing unsatisfactory performance. The seemingly superior safety of BC and GAIL is due to the almost non-existent forward movement of the AV. This can be verified by the test episodic returns of BC in Fig.~\ref{fig7}. Similarly, it is not surprising that BC, CQL, and GAIL have a low disturbance rate for the same reason. The reason for this is that IL methods optimize the agent at each timestep to mimic the accurate actions. In contrast, HAIM-DRL considers trajectory-based learning. We encourage the agent to choose actions that could potentially yield reward in future trajectories, rather than simply mimicking expert behavior at each step. Moreover, HAIM-DRL collects expert data online, alleviating the severe distribution shift problem in offline RL methods, thus achieving better performance.

\subsubsection{Compared to Conventional Human-in-the-loop RL Methods}

\begin{figure*}[t]
  \centering
  \subfloat[]{\includegraphics[width=0.245\textwidth, height=3.85cm]{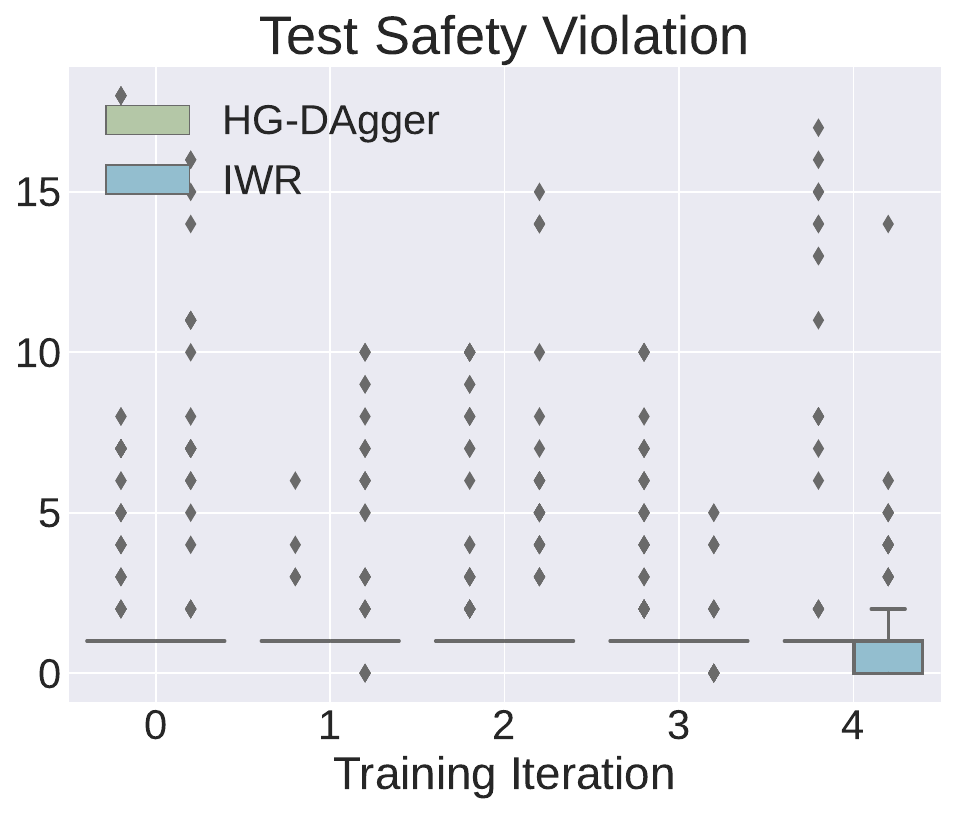}}
  \hfill
  \subfloat[]{\includegraphics[width=0.245\textwidth, height=3.85cm]{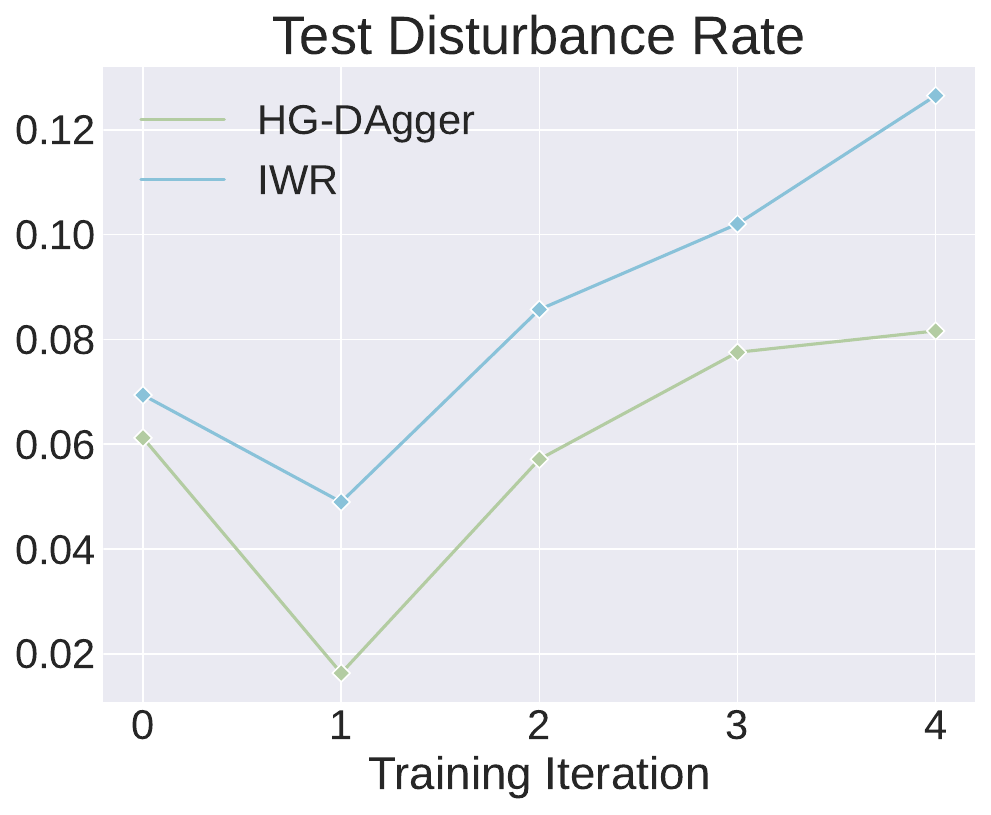}}
  \hfill
  \subfloat[]{\includegraphics[width=0.245\textwidth, height=3.85cm]{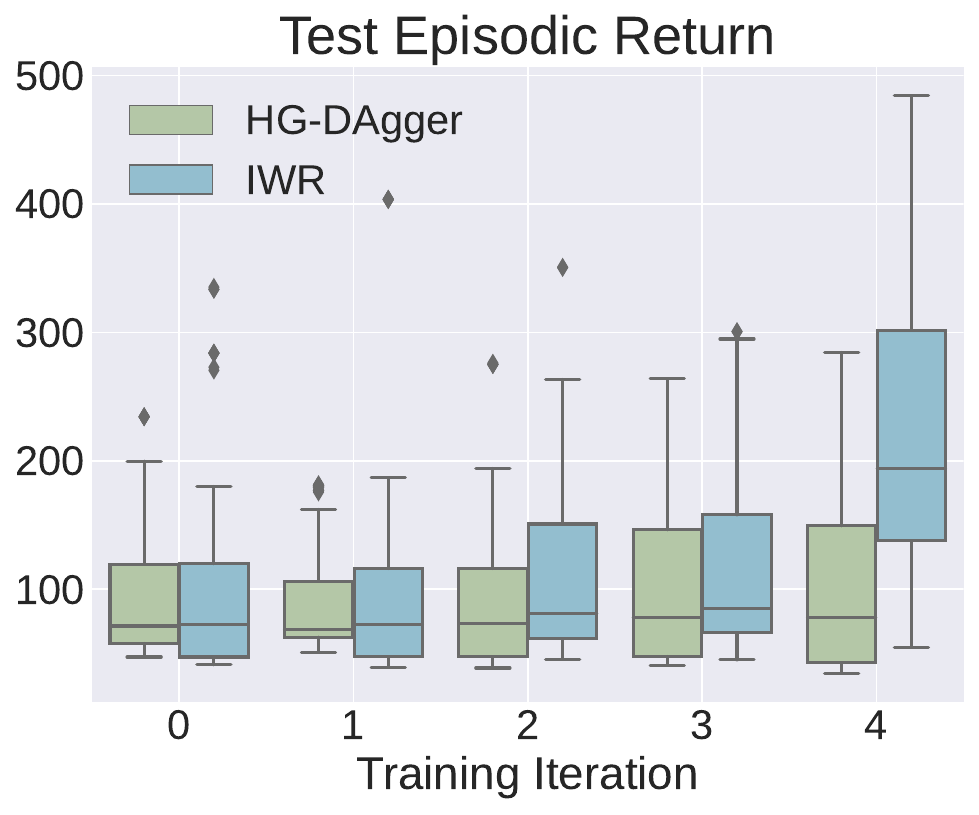}}
  \hfill
  \subfloat[]{\includegraphics[width=0.245\textwidth, height=3.85cm]{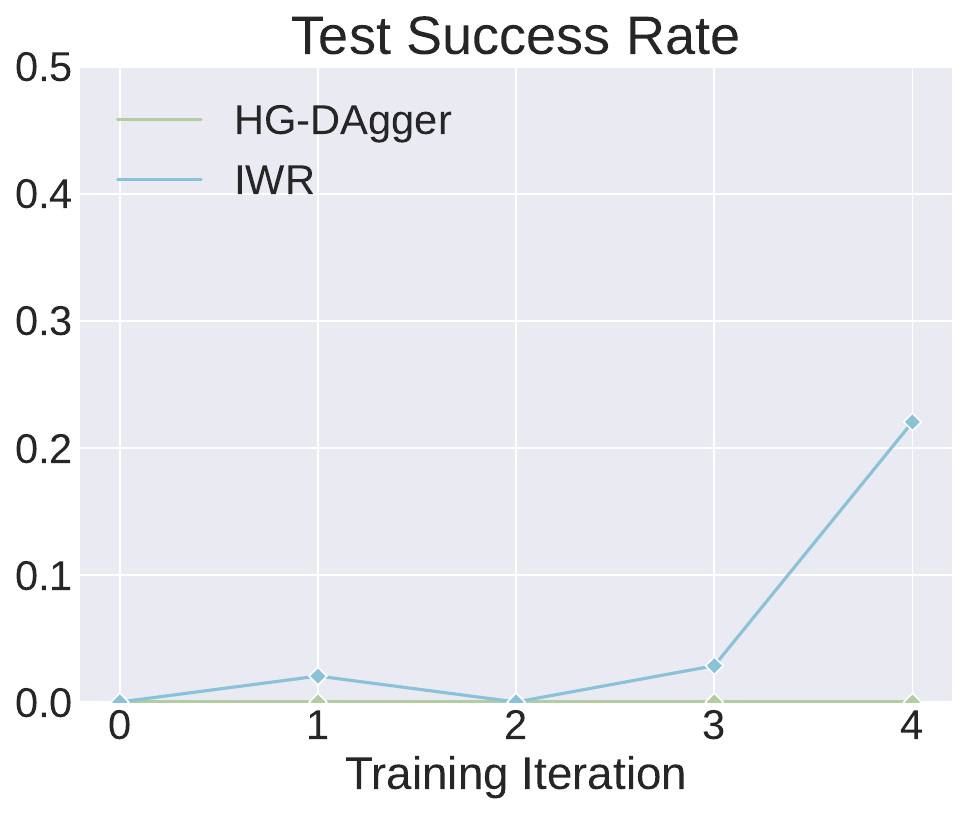}}
  \caption{Performance comparison of HAIM-DRL with conventional human-in-the-loop RL methods.}
  \label{fig8}
\end{figure*}

We benchmarked the performance of two conventional human-in-the-loop methods using the human dataset collected above: HG-DAgger~\citep{kelly2019hg} and IWR~\citep{mandlekar2020human}. Both of these methods necessitate a preliminary warm-up phase, which involves behavior cloning from a pre-collected dataset. \cite{li2022efficient} revealed that lesser quantities, such as 10K or 20K steps of human-collected data, were inadequate to initialize the policy with basic driving skills. Consequently, we employed a more substantial dataset containing 30K transitions for the warm-up phase. Following the initial warm-up phase, both HG-DAgger and IWR integrated human intervention data into their training buffers. Subsequently, they executed behavior cloning to update the policy over 4 epochs. Throughout each epoch, these human-in-the-loop methods necessitated a minimum of 5,000 steps of real-time human supervision.

The learning curves for both methods, following a 30K transition warm-up, are depicted in Fig.~\ref{fig8}. Table~\ref{tab1} also details the performance of these two methods under the 50K transition warm-up condition. From Fig. \ref{fig8} and Table\ref{tab1}, it is evident that only IWR can achieve an acceptable success rate, as it prioritizes human intervention samples and successfully learns crucial maneuvers to avoid hazardous situations caused by compounded errors. Conversely, HG-DAgger struggles to learn from a limited number of critical human demonstrations, as it does not incorporate a mechanism for re-weighting human takeover data. Even with its relative superiority over HG-DAgger, IWR's performance does not reach the benchmark set by HAIM-DRL. Both the success rate and the disturbance rate for IWR fall below the anticipated levels of satisfaction.

\begin{figure*}[t]
  \centering
  \subfloat[]{\includegraphics[width=0.245\textwidth, height=3.85cm]{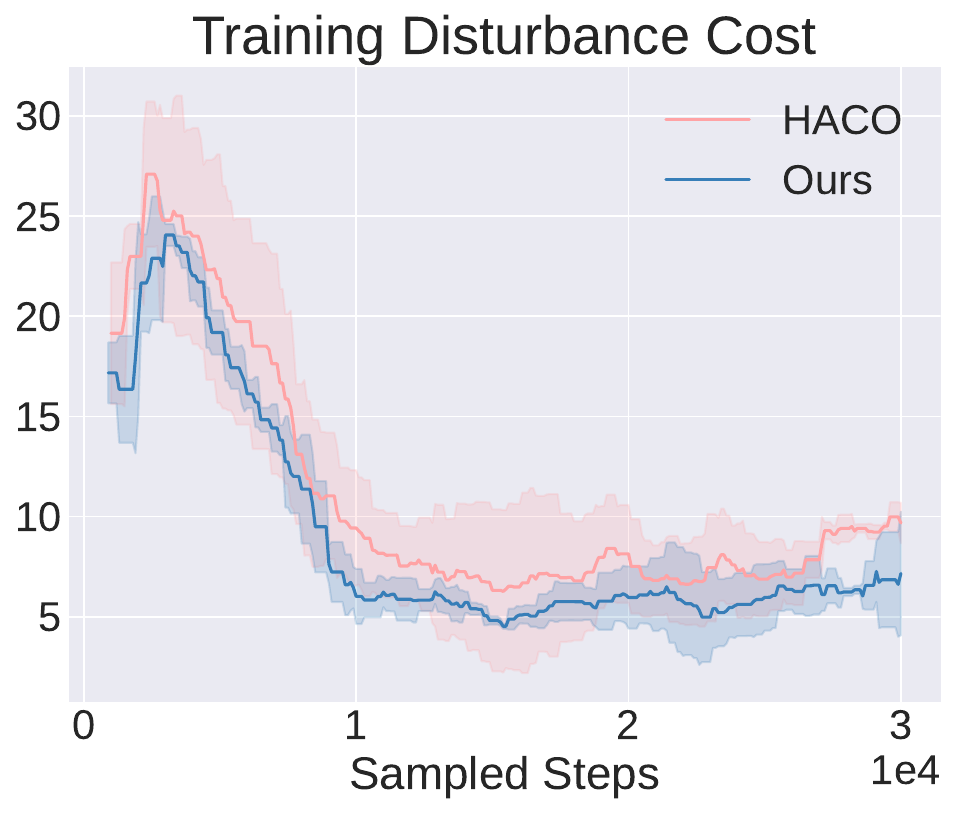}}
  \hfill
  \subfloat[]{\includegraphics[width=0.245\textwidth, height=3.85cm]{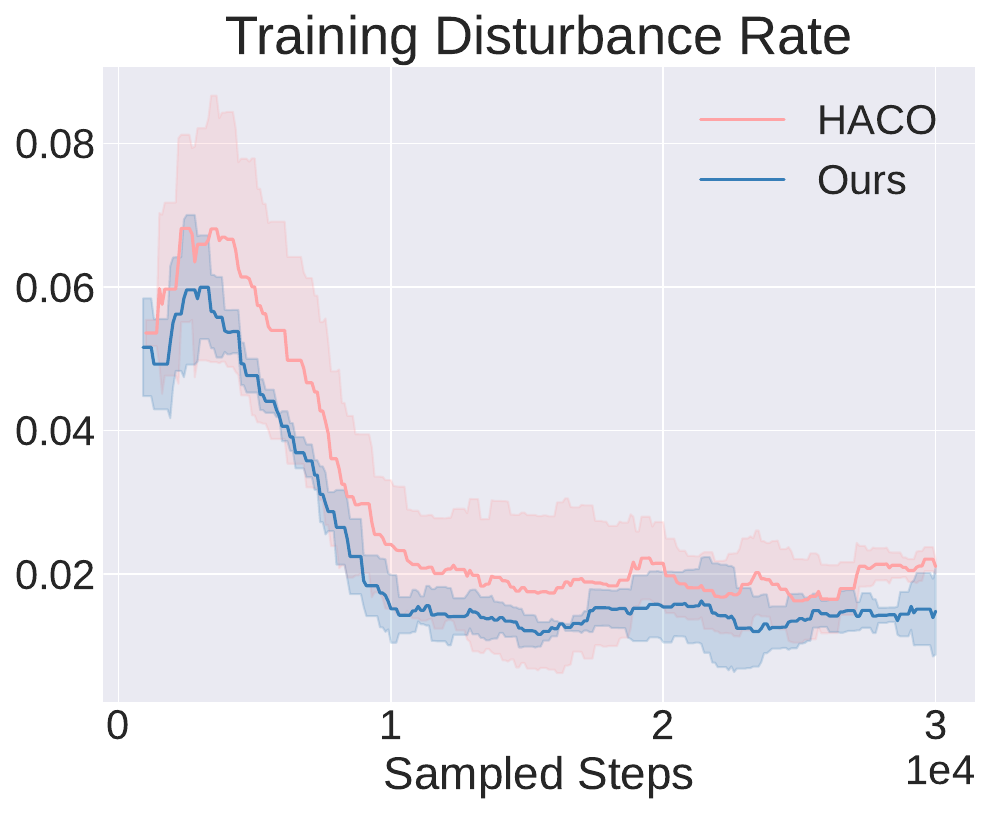}}
  \hfill
  \subfloat[]{\includegraphics[width=0.245\textwidth, height=3.85cm]{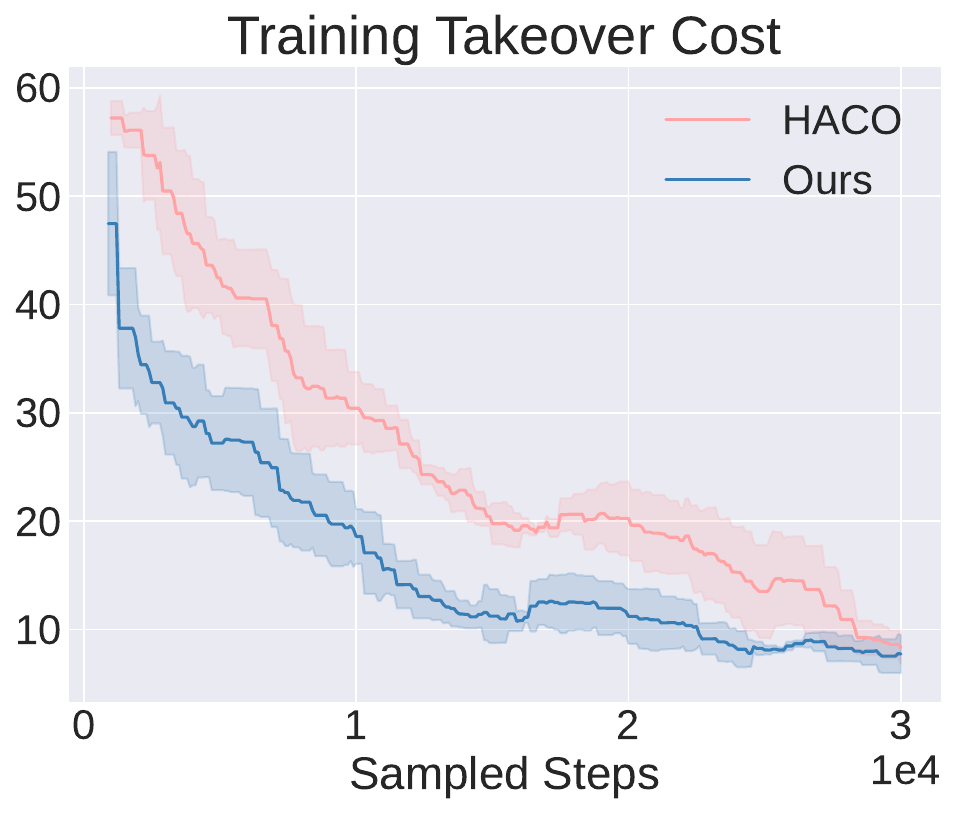}}
  \hfill
  \subfloat[]{\includegraphics[width=0.245\textwidth, height=3.85cm]{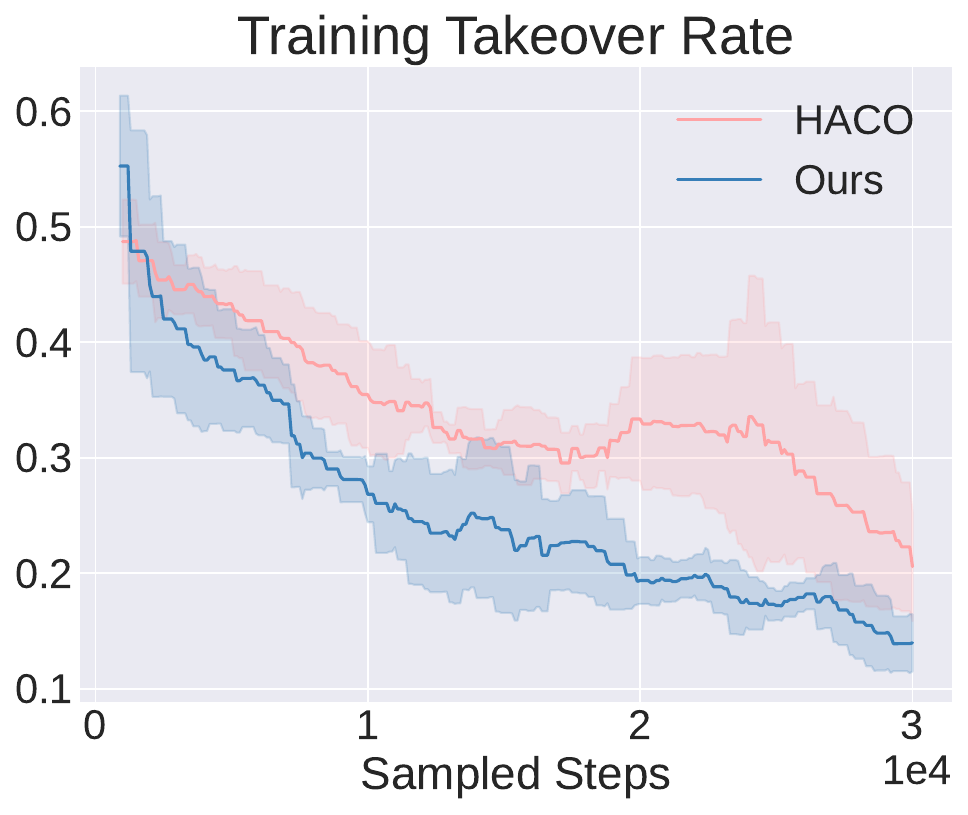}}
  \newline
  \subfloat[]{\includegraphics[width=0.245\textwidth, height=3.85cm]{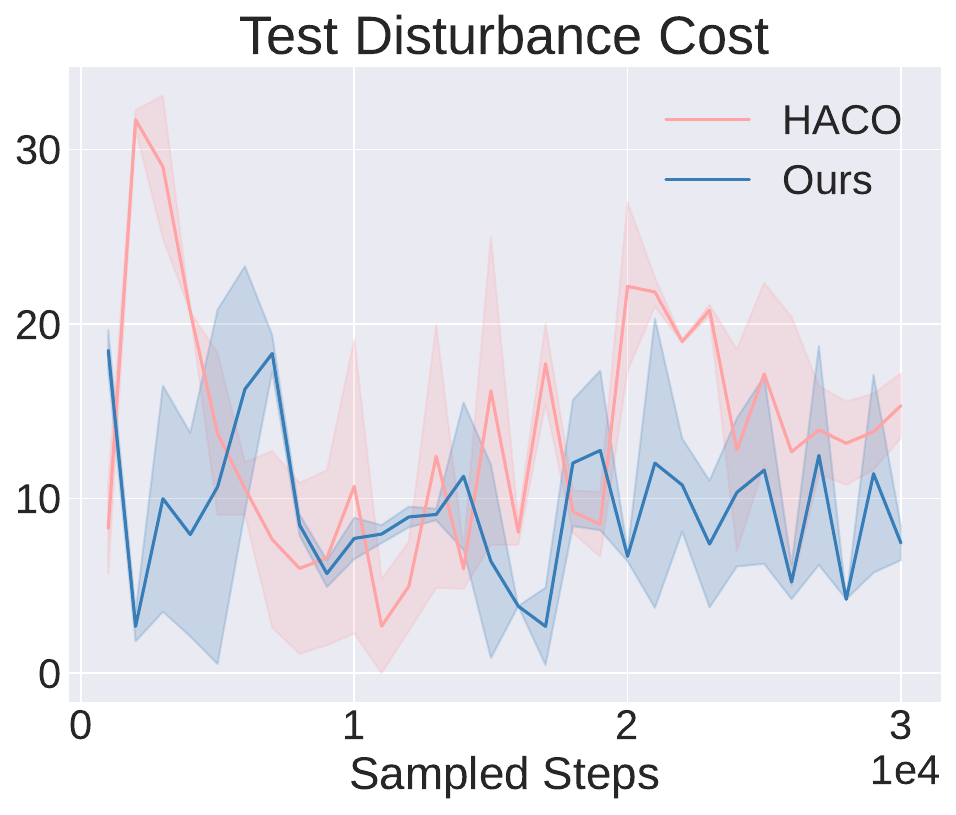}}
  \hfill
  \subfloat[]{\includegraphics[width=0.245\textwidth, height=3.85cm]{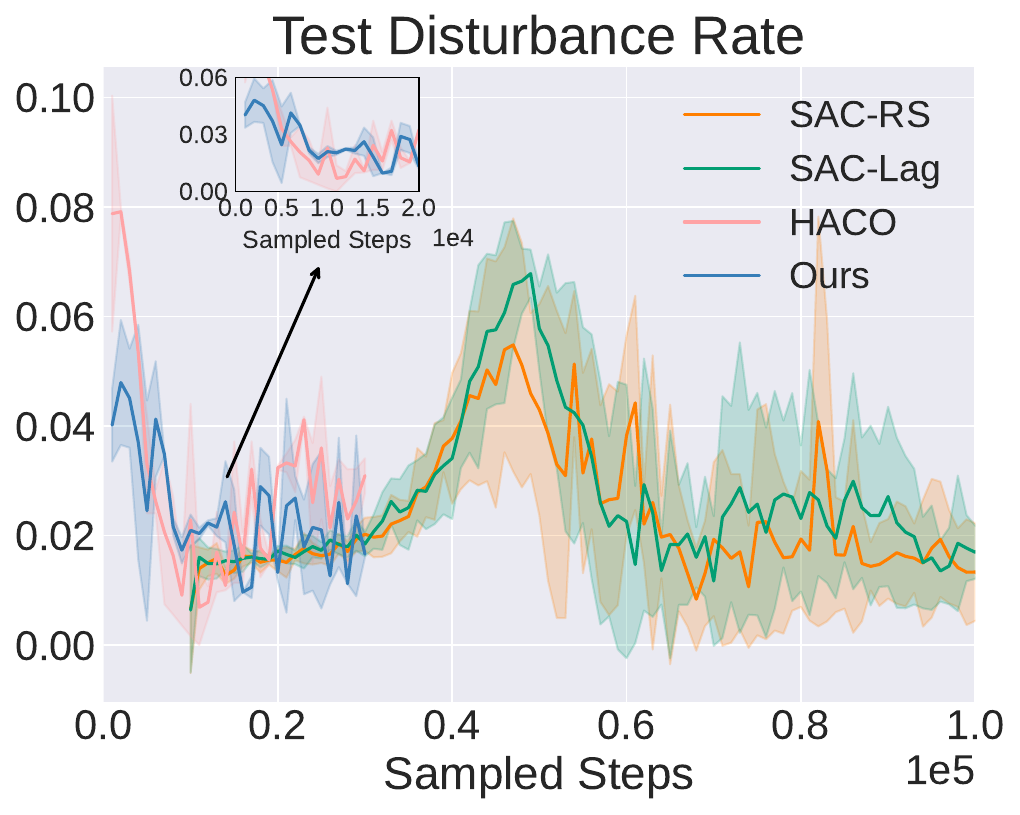}}
  \hfill
  \subfloat[]{\includegraphics[width=0.245\textwidth, height=3.85cm]{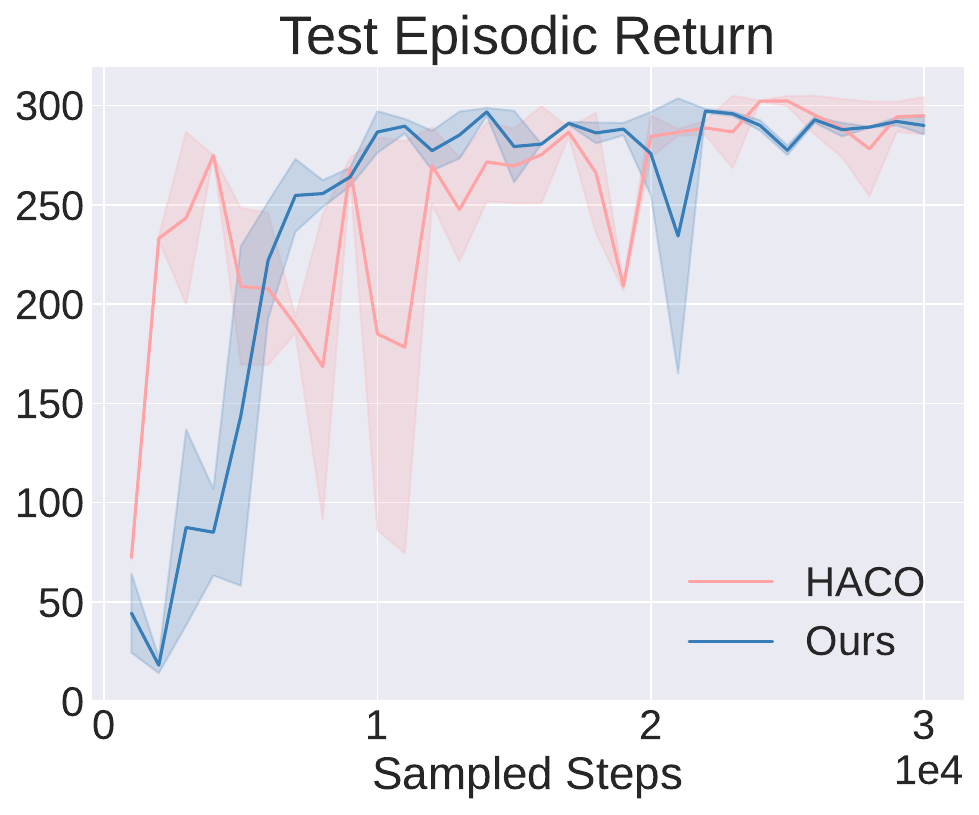}}
  \hfill
  \subfloat[]{\includegraphics[width=0.245\textwidth, height=3.85cm]{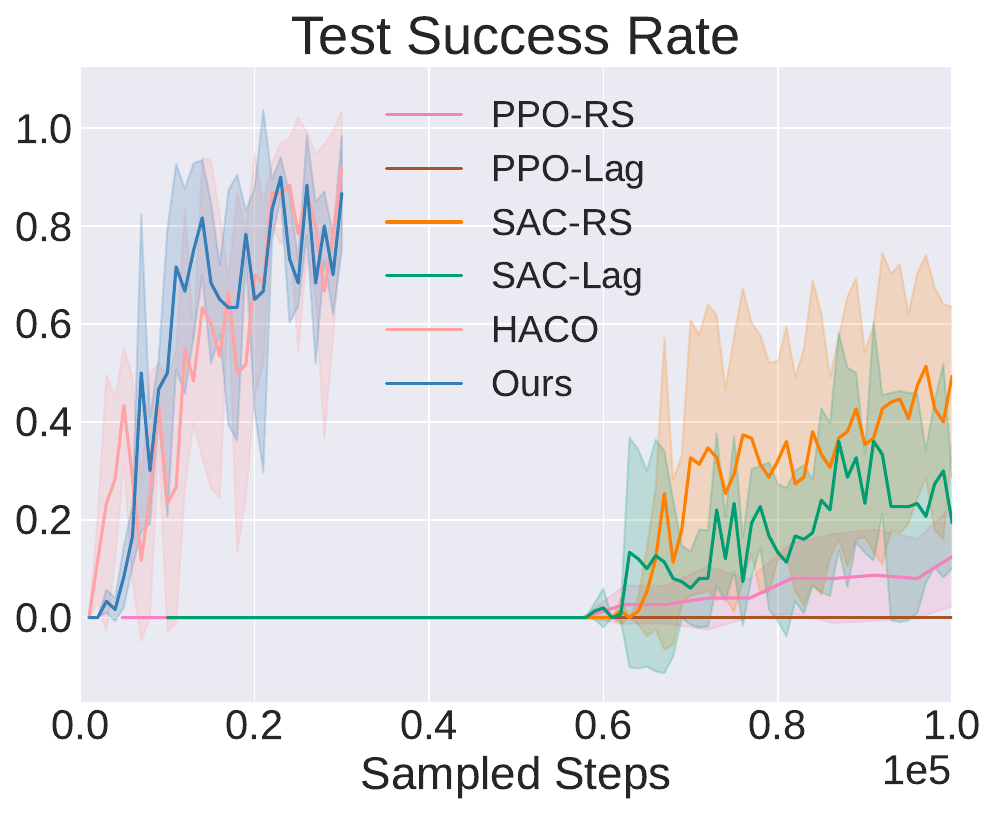}}
  \caption{Performance comparison of HAIM-DRL with human-AI copilot RL method.}
  \label{fig9}
\end{figure*}

\subsubsection{Compared to Human-AI Copilot Method}

We also conducted a comparative analysis with the recently developed human-AI copilot optimization (HACO) method\citep{li2022efficient}. We used the average episodic intervention occurrence to represent the frequency of interventions. As indicated in Fig.~\ref{fig9}, at the start of the training, the human expert is more frequently involved in driving demonstrations to prevent the agent from entering hazardous states. Subsequently, as the agent acquires basic driving skills, it tends to select actions more agreeable to the human expert, thereby reducing takeover rate. The takeover cost and takeover rate curves show that both HACO and HAIM-DRL experience a decline in human takeover frequency, while the consistency between agent and human behavior increases. This decline in HAIM-DRL can be attributed to its intervention minimization mechanism, which reduces both the cost and frequency of human takeover.

\begin{figure*}[!ht]
  \centering
  \subfloat[]{\includegraphics[width=0.47\textwidth, height=4.65cm]{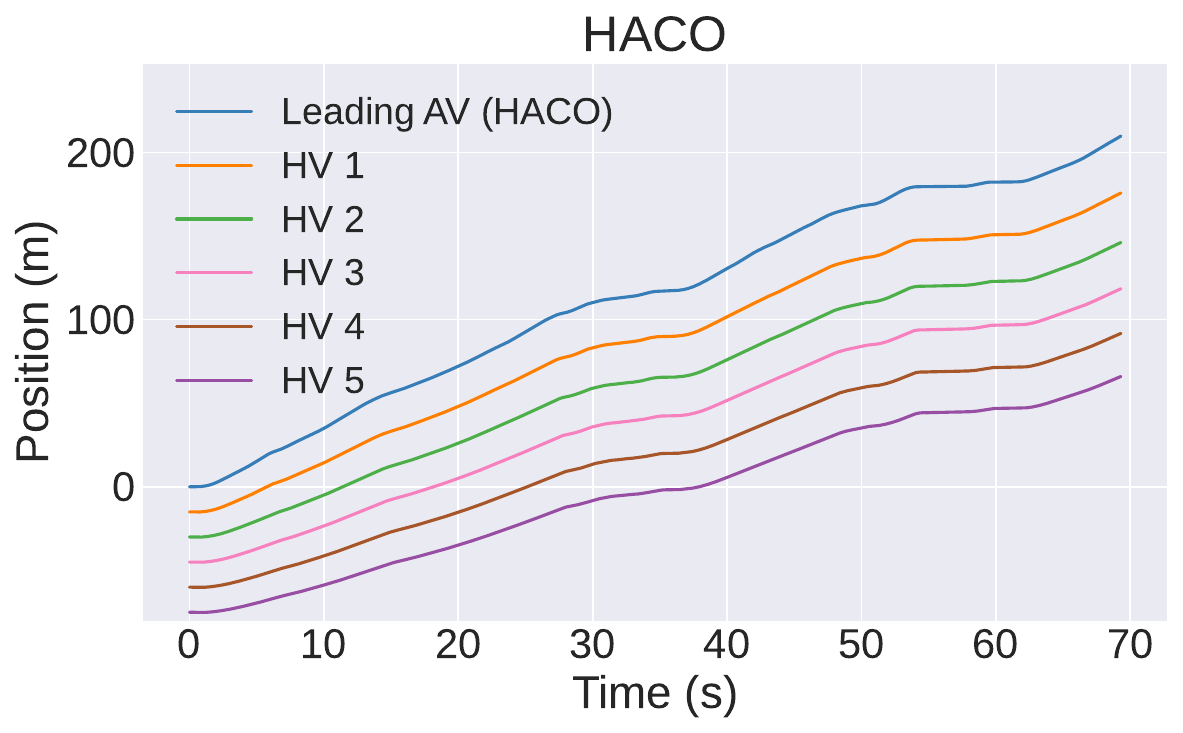}}
  \hfill
  \subfloat[]{\includegraphics[width=0.47\textwidth, height=4.65cm]{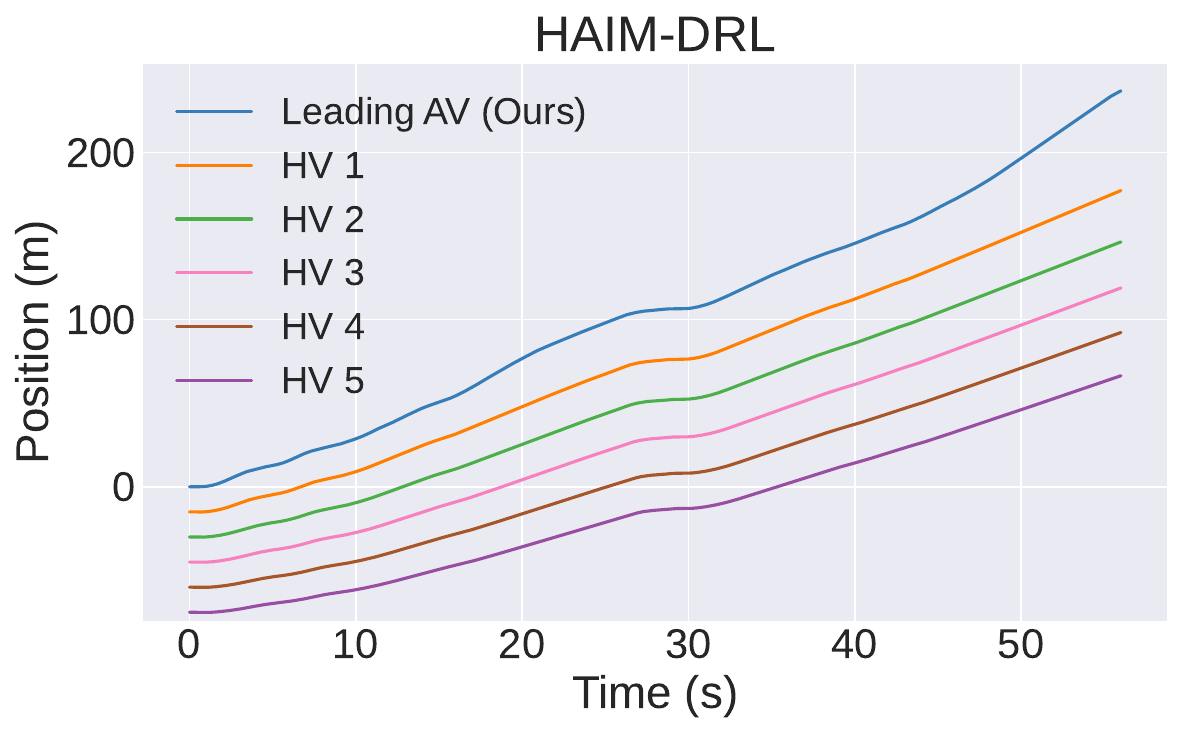}}
  \newline
  \subfloat[]{\includegraphics[width=0.47\textwidth, height=4.65cm]{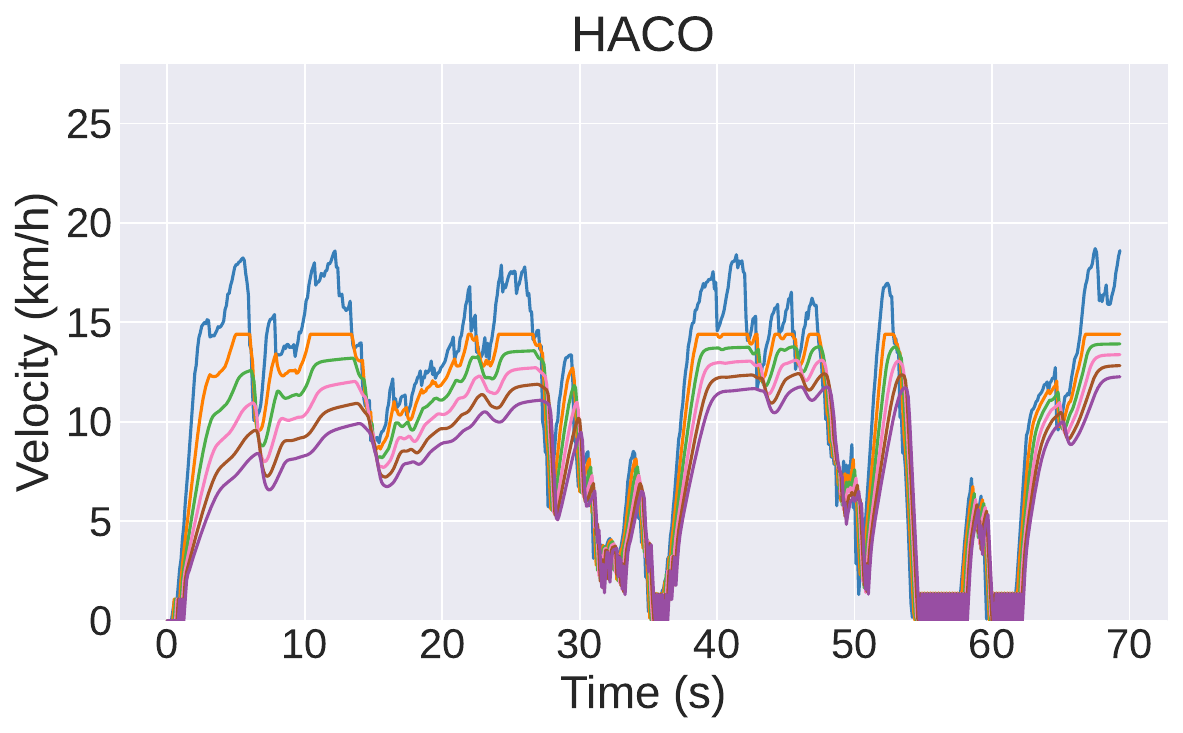}}
  \hfill
  \subfloat[]{\includegraphics[width=0.47\textwidth, height=4.65cm]{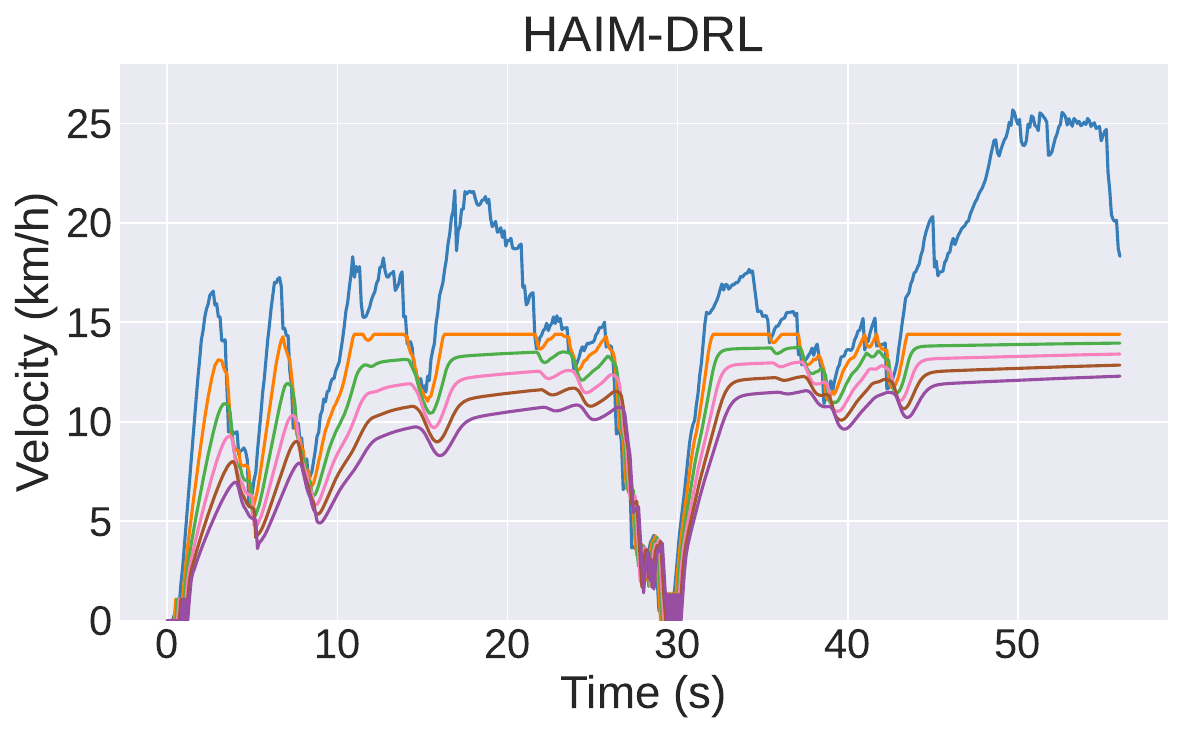}}  
  \newline
  \subfloat[]{\includegraphics[width=0.47\textwidth, height=4.65cm]{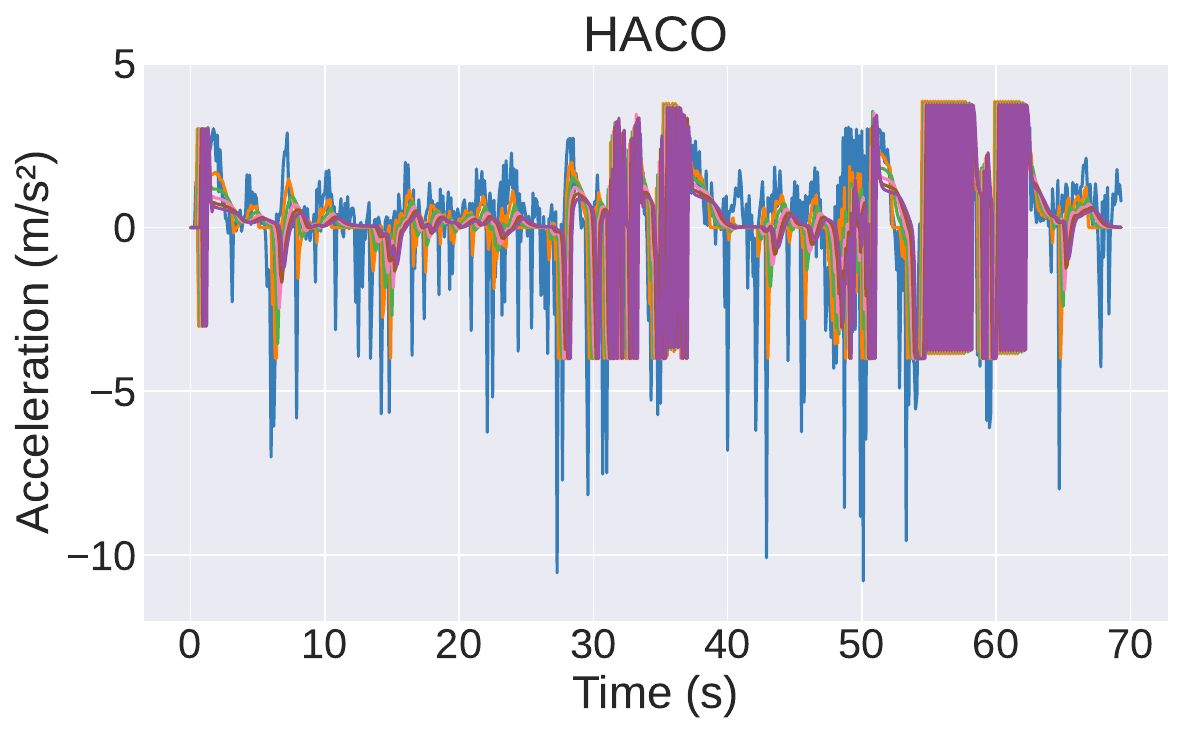}}
  \hfill
  \subfloat[]{\includegraphics[width=0.47\textwidth, height=4.65cm]{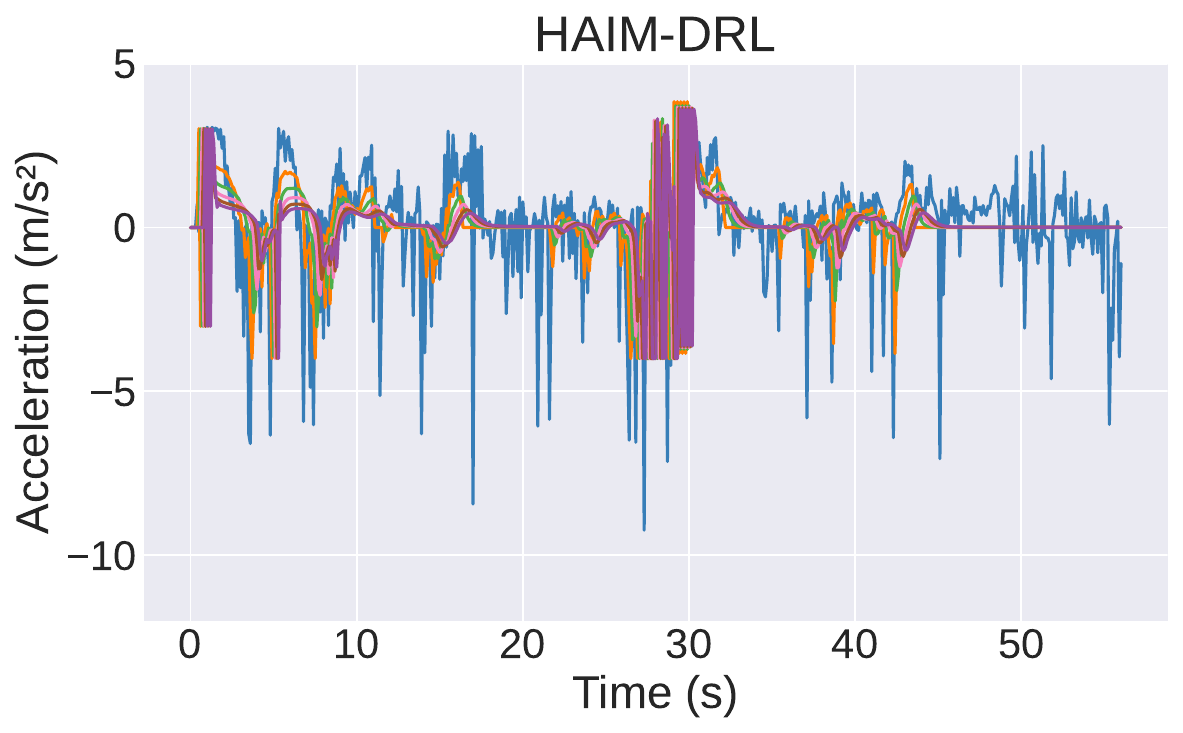}} 
  \caption{Comparative analysis of traffic flow efficiency: HACO vs. HAIM-DRL. (a), (c), and (e) illustrate the time-based dynamics of position, velocity, and acceleration for the HACO method. In contrast, (b), (d), and (f) present these dynamics for the HAIM-DRL method. }
  \label{fig10}
\end{figure*}

Compared to HACO, HAIM-DRL achieves faster convergence speeds, with reward stabilizing at 300 at the 10K step mark and a success rate of 80\% at the 20K step mark. During the training phase, after 20K steps, the takeover cost for HAIM-DRL is as low as 10 with a takeover rate of only 0.2, while HACO still requires a takeover cost of 20 and a takeover rate of 0.3. This disparity is likely due to HACO's lack of consideration for implicit intervention, resulting in poorer convergence of takeover cost. This difference is directly reflected in sampling efficiency: HAIM-DRL requires only 7,759 {\tiny $\pm$ 462.35} human expert interventions, whereas HACO needs 8,369 {\tiny $\pm$ 453.78}. This can be understood as HACO being akin to a student learning from a single teacher through explicit intervention, whereas our HAIM-DRL benefits from the instruction of two teachers—using both explicit and implicit interventions, which enables faster learning.

Importantly, HACO did not take into account the impact on subsequent traffic flow, leading to fluctuations in traffic flow. As shown in Fig.~\ref{fig9}, at the 20k step, HAIM-DRL's disturbance rate already stabilizes at around 0.02 with a disturbance cost of about 10, while HACO's disturbance rate is still fluctuating around 0.04, with a disturbance cost of approximately 15. To further illustrate the advantages of HAIM-DRL in terms of traffic flow efficiency, we selected an episode for a detailed presentation, as shown in Fig.~\ref{fig10}. The blue curve represents the trajectories of both HACO and HAIM-DRL, while the other curves represent IDM-controlled HVs following the AV. The difference in the horizontal axis lengths between the HACO and HAIM-DRL plots is attributed to the higher average velocity of HAIM-DRL, which enables the vehicles to reach their destination more quickly. 

It is evident that the HACO-controlled AV exhibits more frequent stopping behaviors, whereas the trajectory of HAIM-DRL is smoother. In terms of velocity, HACO leads to three abrupt drops to zero, while HAIM-DRL experiences only one. This stop for HAIM-DRL occurs at a roundabout, necessitating a yield, which is reasonable. Notably, the relatively low upper-velocity limit is due to restrictions on the maximum velocity of steering wheel-controlled vehicles to 40 km/h in the MetaDrive simulator. From Fig.~\ref{fig10} (c), HACO sometimes decelerates sharply, exceeding -10, causing traffic flow disturbance (indicated by the dense purple area). In contrast, HAIM-DRL, due to its implicit intervention mechanism, employs gentler braking, thus causing less disturbance to subsequent traffic flow. Overall, HAIM-DRL not only ensures safe driving for AVs but also positions them as courteous and considerate drivers within the traffic system.

\subsection{Ablation Study}

\subsubsection{Analysis of Explicit Intervention Strategy}

We engaged a human participant to test two different takeover strategies. The first strategy involves a small number of takeovers, each producing long trajectories and typically generating sparse takeover cost. The second strategy requires more frequent takeover, leading to fragmented demonstrations. As illustrated in Table~\ref{tab2} (a), HAIM-DRL performs best with dense human takeover signals. AVs trained with longer trajectories, in contrast to those receiving dense takeover signals, demonstrate lower success rates and experience reward in a more erratic manner.

\subsubsection{Cosine Similarity as Takeover Cost Function}

Table~\ref{tab2} (b) displays the outcomes when we altered the takeover cost, as specified in Eq.~\ref{Eq23}, to a constant value of `+1' in response to a human takeover. It was observed that the agent learned to remain stationary at spawn points, and no movements were recorded during the testing phase. This could be due to the incorrect timing of human takeover, which disrupts the agent's ability to drive accurately. This is consistent with the findings of ~\cite{li2022efficient}. However, the negative cosine similarity, when used to measure the divergence between the actions of the agent and the human, helps mitigate this issue. This is because the penalty for human takeover is lessened when the agent's actions align with the human's takeover.

\subsubsection{The Necessity of Minimizing Explicit Intervention}

Table~\ref{tab2} (c) indicates that in the absence of an intervention minimization mechanism, the AV agent is prone to driving straight toward the boundary. This is aligned with the results of ~\cite{li2022efficient,wu2023toward}. This occurs because the agent learns to exploit the human expert's tendency to take over persistently, thereby inflating the proxy $Q$ values and increasing the cognitive load on the human mentor. Agents might even engage in risky behaviors, such as deliberately deviating from the road when approaching boundaries, to trick the intervention mechanism and compel the human expert to take control and provide demonstrations.

\subsubsection{The Importance of Implicit Intervention}

After the implicit intervention mechanism was removed, the disturbance rate noticeably increased, as illustrated in Table~\ref{tab2} (d). Hence, the implicit intervention mechanism not only helps lower the takeover cost, and accelerate the learning speed of the agent, but most importantly, it can minimize disturbance of the traffic flow. Additionally, by reducing the gap between the transportation and robotic communities, it facilitates the formation of a unified objective, thereby accelerating the practical application of AVs at an early stage.

\begin{table*}[!ht]
\centering
\begin{small}
\caption{
Performance outcomes with ablation studies in the HAIM-DRL.
}
\label{tab2}
\begin{tabular}{@{}lcccc@{}}
\toprule 
Experiment & \shortstack {Test\\Return} &  \shortstack{Test\\Cost} & \shortstack{Test\\Success Rate} & \shortstack{Test\\ Disturbance Rate} \\
\midrule
\textbf{(a)} Human takeover less frequently & 219.0 & 0.865 & 0.397 & - \\
\textbf{(b)} W/o cosine similarity takeover cost & 25.0 & 0.44 & 0.02 & - \\
\textbf{(c)} W/o intervention minimization & 82.0 & 1.05 & 0.02 & - \\
\textbf{(d)} W/o implicit intervention  & 347.0 & 0.77 & 0.85 & 0.035	{\tiny $\pm$ 0.0094 } \\
\midrule
\textbf{HAIM-DRL} & 354.34 {\tiny $\pm$ 11.08 } & 0.76 {\tiny $\pm$ 0.28 } & 0.85{\tiny $\pm$ 0.03} & 0.023	{\tiny $\pm$ 0.00072 } \\
\bottomrule
\end{tabular}%
\end{small}
\end{table*}

\subsection{Comparative Results in the CARLA Simulator} \label{Comparative Results in the CARLA Simulator}

\begin{figure*}[!ht]
  \centering
  \subfloat[]{\includegraphics[width=0.245\textwidth, height=3.85cm]{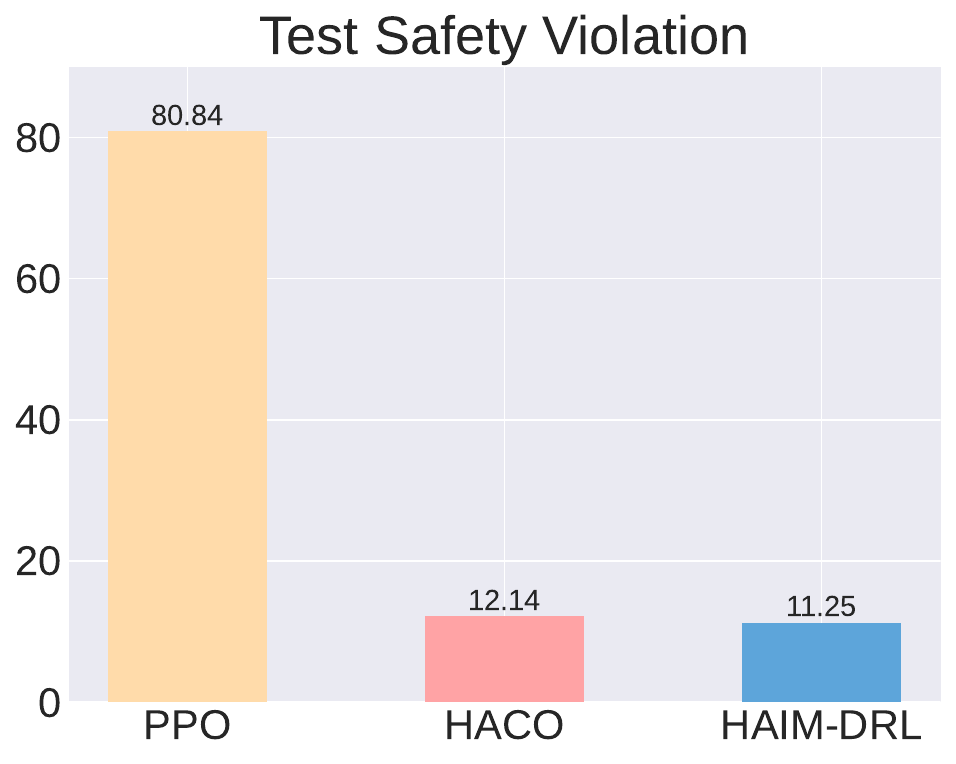}}
  \hfill
  \subfloat[]{\includegraphics[width=0.245\textwidth, height=3.85cm]{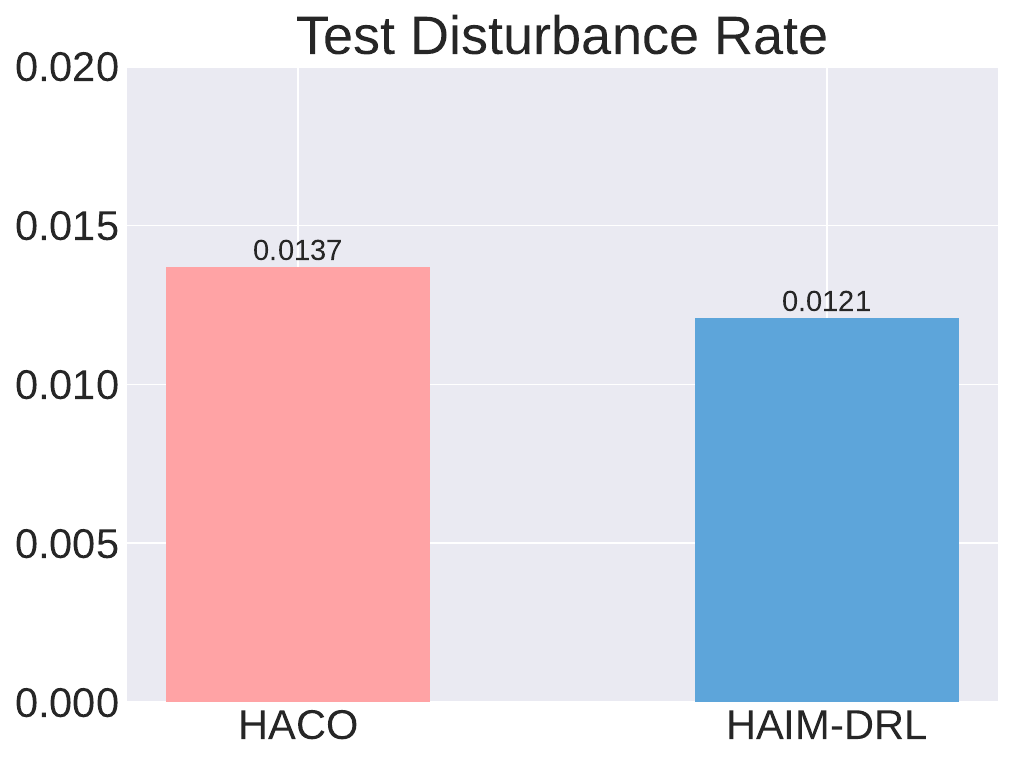}}
  \hfill
  \subfloat[]{\includegraphics[width=0.245\textwidth, height=3.85cm]{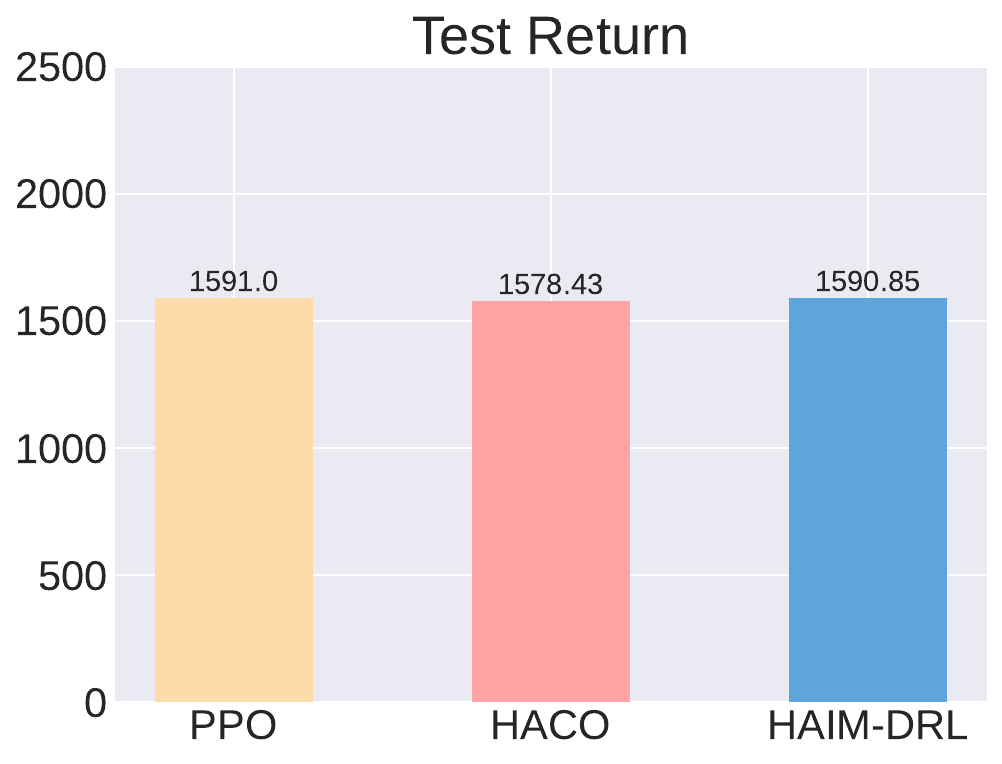}}
  \hfill
  \subfloat[]{\includegraphics[width=0.245\textwidth, height=3.85cm]{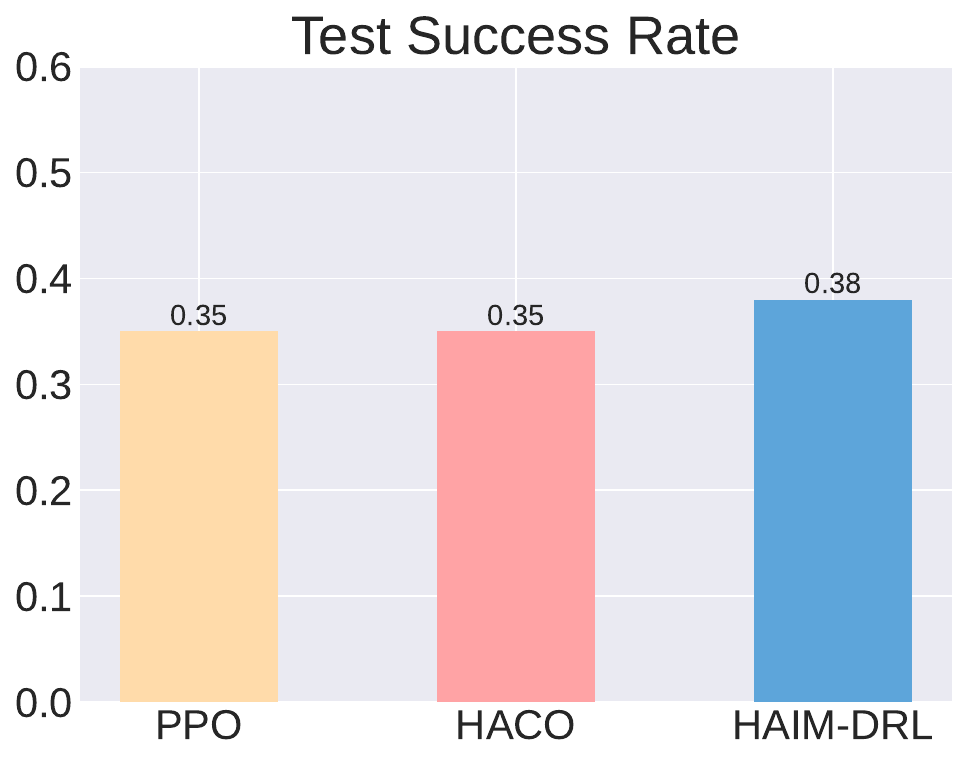}}
  \caption{Performance comparison of HAIM-DRL with PPO/HACO methods in the CARLA simulator.}
  \label{fig11}
\end{figure*}

To examine the transferability of HAIM-DRL, we deployed it within the CARLA simulator. Specifically, we compared the performance of three representative methods, i.e., PPO, HACO, and HAIM-DRL, with the results detailed in Fig.~\ref{fig11} and Table~\ref{tab3}. In the experiments conducted using the CARLA simulator, we used the visual observation space of Fig.~\ref{fig4} (a). Specifically, the experiments employed the top-down semantic view offered by CARLA as the input. A 3-layer CNN was used as the feature extractor for all three methods tested. For PPO, the reward settings were based on those described in CARLA. We trained HACO and HAIM-DRL (with a human expert), as well as PPO, in CARLA Town 1, while performance testing was reported for Town 2.

As evidenced by Fig. \ref{fig11} and Table \ref{tab3}, our HAIM-DRL method is successfully implemented in the CARLA simulator, where it exhibits superior performance, particularly in safety violation, success rate, and disturbance rate. This success is a testament to the robustness and versatility of HAIM-DRL across different simulation environments. In addition, both HACO and HAIM-DRL demonstrate significant sampling efficiency, requiring only 8k samples from the environment to efficiently train the driving agent. 

All three methods show a decline in test success rate in CARLA compared to their performance in MetaDrive. This difference is primarily attributed to the variance in training sample sizes. While HAIM-DRL is trained with 30K samples in MetaDrive, its training in CARLA is limited to just 8K samples. This significant reduction in the number of training samples inevitably impacts the methods' ability to learn and adapt to the driving environment. Besides, in CARLA, the training is confined to a single scenario, limiting the scope and diversity of the learning experience. In contrast, MetaDrive offers a virtually infinite array of non-repetitive scenarios, providing a richer and more varied training environment.

\begin{table*}[!ht]
\centering
\begin{small}
\caption{The performance of PPO/HACO/HAIM-DRL methods in the CARLA simulator.}
\label{tab3}
\begin{tabular}{@{}cccccc@{}}
\toprule 
Method & \shortstack{Test Safety Violation} & \shortstack{Test Return} & \shortstack{Test Disturbance Rate} & \shortstack{Test Success Rate} & \shortstack{Train Samples}  \\
\midrule
PPO &	80.84	& 1591.00 &	- &	0.35	& 500,000 \\
HACO &	12.14	& 1578.43 &	0.0137 &	0.35	& 8,000 \\
HAIM-DRL & 11.25 & 1590.85 & 0.0121 & 0.38 & 8,000 \\
\bottomrule
\end{tabular}%
\end{small}
\vspace{-1em}
\end{table*}

\section{Conclusions and Future Work}

Despite the significant advancements in autonomous driving technology, ensuring the safety of autonomous vehicles (AVs) and traffic flow efficiency in mixed traffic environments remains a complex challenge. Addressing this, our paper proposes an enhanced human-in-the-loop reinforcement learning method for driving policy learning, leveraging the synergies between the transportation and robotics communities. At the core of our approach is an innovative learning paradigm inspired by the human learning process, which is termed \textbf{human as AI mentor (HAIM)}. A defining feature of HAIM is that the human expert serves as a mentor to the AI agent, supervising, intervening, and demonstrating in its learning process. Building upon this concept, we propose a HAIM-based deep reinforcement learning (DRL) framework, named \textbf{HAIM-DRL}. By using HAIM-DRL, not only can we ensure the safety of the AV itself, but also optimize the traffic flow efficiency.

Our experimental findings highlight the distinct advantages of HAIM-DRL over traditional methods, including IL, Safe RL, and conventional human-in-the-loop RL. Specifically, the HAIM-DRL framework not only enhances the safety of individual AVs but also considers their impact on traffic systems, offering several key benefits: (a) Improved training safety: This aspect is crucial for the potential real-world testing and deployment of AVs. (b) Enhanced sampling efficiency: The method accelerate the training process, a significant step towards practical implementation. (c) Smoother traffic flow: HAIM-DRL contributes to reducing traffic flow disturbance, facilitating a more harmonious integration of AVs into existing traffic systems. (d) Superior generalizability: The ability of HAIM-DRL to adapt across different scenarios underpins its robustness.

In essence, our proposed HAIM-DRL not only champions safe and efficient autonomous driving but also sets a new benchmark for human-AI collaboration in the transportation field. Looking to the future, the exploration of real-world applications, scalability, and further enhancement of our method will be pivotal. Specifically, considering a relaxation of the current reliance on perfect human experts and learning from a diversity of human drivers could be a significant direction. Designing mechanisms to better incorporate transportation knowledge into our framework will be a key area of focus. Continued research in these areas will be instrumental in advancing the integration of AVs into mixed traffic systems, ultimately contributing to the evolution of smart and safe transportation ecosystems.

\section{Replication and Data Sharing}
The code necessary for replication of this study is accessible at \href{https://github.com/zilin-huang/HAIM-DRL}{https://github.com/zilin-huang/HAIM-DRL}

\section{Declaration of Interests}
The authors declare that they have no known competing financial interests or personal relationships that could have appeared to influence the work reported in this paper.

\section{Acknowledgement}
No external funding was received for this work.

\bibliography{mybibfile}

\appendix

\section{Proof of Main Theorem on Explicit Intervention Mechanism }
\label{Appendix:Proof of Explicit Intervention Mechanism}
In this section, we derive the upper bound on the discounted probability of failure for HAIM's explicit intervention mechanism, showing that we can bound the training safety with the human experts.

\textbf{Theorem 1} (Upper bound of training risk). The expected cumulative probability of failure, denoted as ${V_{\pi_{mix}}}$, for the mixed behavior policy ${\pi_{mix}}$ within HAIM, is bounded by three factors: the error rate of the human expert's action ($\epsilon$), the error rate of the human expert's takeover decisions ($\kappa$), and the human expert's threshold of tolerance ($K'$):

\begin{equation}
\label{equation: main theorem}
{V_{\pi_{mix}}}
\le
\cfrac{1}{1 - \gamma}[
\epsilon + \kappa + \cfrac{\gamma \epsilon^2}{1 - \gamma} K'
]
\end{equation}
wherein $K'= \max_s K(s) = \max_s \mathop{\int}_{a \in {\mathcal A_{human}}(s)}da \ge 0$ is referred to as the human expert tolerance. $\gamma\in(0,1)$ is the discount factor. The human expert's tolerance $K'$ will increase when human takeover is reduced, enabling the agent to explore the environment more freely.

In the following, we present the proof.

\textit{\textbf{Notations.}} Before we begin, let's recap and explain the notations. In HAIM, a human expert has the authority to intervene and override the original agent's actions when necessary, providing real-time demonstrations through their actions. The human's policy is denoted as $\pi_{human}: a_t^{human} \sim \pi_{human}(\cdot \mid s_t)$, and the agent's policy is $\pi_{AV}\left(a_t^{AV} \mid s_t\right)$. Both policies generate actions within the bounded action space $\mathcal{A} \in \mathbb{R}^{|\mathcal{A}|}$. The action-based switch function used in this paper is:

\begin{equation} 
\begin{aligned}
\mathcal{T}(s_t, a_t, \pi_{human}) = \begin{cases}
    (a_t^{AV}, 0), & \text {if takeover } \\ 
    \left(a_t^{human} \sim \pi_{human}(\cdot \mid s_t), 1\right), & \text { otherwise }
\end{cases}
\end{aligned}
\end{equation}

We use the Boolean indicator \(I\left(s_t\right)\) to represent human takeover, and the resulting action applied to the environment is \(\hat{a}_t^{mix} = I\left(s_t\right) a_t^{human}+\left(1-I\left(s_t\right)\right) a_t^{AV}\) \citep{wu2023toward}. As a result, the actual trajectory during the training process is determined by the mixed behavior policy:

\begin{equation} 
\pi_{mix}(a \mid s)=\pi_{AV}(a \mid s)(1-I(s, a))+\pi_{human}(a \mid s) F(s)
\end{equation}
where, $F(s)=\int_{a^{\prime} \notin \mathcal{A}_\eta(s)} \pi_{AV}\left(a^{\prime} \mid s\right) d a^{\prime}$  represents the probability of the agent selecting an action that would be rejected by the human.

Hence, for a given state, we can partition the action space into two categories: actions where human takeover is likely to occur and actions where it is not expected. We refer to the set of actions that are unlikely to be rejected by a human expert at state $s$ as the \textit{confident action space}, denoted as \citep{peng2022safe}:
\begin{equation}
{\mathcal A_{human}}(s) = \{a: I(a|s) \text{ is False}\}.
\end{equation}
The confident action space comprises actions that are unlikely to be rejected by a human expert at the given state $s$.

Here, we further elucidate the concept of safety within the learning environment by integrating a ground-truth indicator $C^{gt}$. This indicator serves as a binary signal to denote the potential of a given action $a_t$ to transition the system into an unsafe state from the current state $s_t$. It's important to note that this unsafe state $s_{t+1}$ is determined by the environment and is not disclosed to the learning policy:

\begin{equation}
C^{gt}(s_t, a_t) = 
\begin{cases}
  1, & \text{if }s_{t+1} = \mathcal P(s_{t+1}|s_t, a_t) \text{ is an unsafe state,} 
  \\
  0, & \text{otherwise.}
\end{cases}
\end{equation}

Therefore, at a given state $s$, the \textit{step-wise probability of failure} for an arbitrary policy $\pi$ is defined as \citep{li2022efficient}:
\begin{equation}
\underset{a \sim \pi(\cdot \mid s)}{\mathbb{E}} C^{g t}(s, a) \in[0,1]
\end{equation}

Moving forward, we introduce the concept of the \textit{cumulative discounted probability of failure}, denoted as $V_\pi\left(s_t\right)$. This is represented mathematically as \citep{li2022efficient}:
\begin{equation}
V_\pi\left(s_t\right)=\underset{\tau \sim \pi}{\mathbb{E}} \sum_{t^{\prime}=t} \gamma^{t^{\prime}-t} C^{g t}\left(s_{t^{\prime}}, a_{t^{\prime}}\right)
\label{equation:cumulative discounted probability of failure}
\end{equation}
where, $\gamma$ is a discount factor that reduces the weight of potential failure in future states. The summation extends from the current time step $t$ to all subsequent time steps $t^{\prime}$, thus encompassing both the present and all possible future trajectories. Eq. \ref{equation:cumulative discounted probability of failure} accounts for the likelihood of entering dangerous states in the current time step as well as in future trajectories predicted by the policy $\pi$.

Building upon Eq. \ref{equation:cumulative discounted probability of failure}, we use $V_{\pi_{human}}=\mathbb{E}_{\tau \sim \pi_{human}} V_{\pi_{human}}\left(s_0\right)$ to represent the expected cumulative discounted probability of failure of the human expert. Similarly, following the same definition as $V_{{\pi_{human}}}$, we can express the expected cumulative discounted probability of failure of the mixed behavior policy as: $V_{\pi_{mix}}=\mathbb{E}_{\tau \sim \pi_{mix}} V_{\pi_{mix}}\left(s_0\right)=\mathbb{E}_{\pi_{mix}} \sum_{t=0} \gamma^t C^{g t}\left(s_t, a_t\right)$.

\textit{\textbf{Assumption.}} As described in Section \ref{Explicit Intervention Mechanism}, we now introduce two important assumptions of human experts in the HAIM’s explicit intervention mechanism.

\textbf{Assumption 1} (High-quality human action). We assume that $a_t^{human}$ is generated by an expert-level human policy and is safe and reliable most of the time. In detail, the step-wise probability of that the human expert produces an unsafe action after takeover is bounded by a small value $\epsilon < 1$:
\begin{equation}
\underset{a \sim \pi_{human}(\cdot \mid s)}{\mathbb{E}} C^{g t}(s, a) \leq \epsilon
\end{equation}

\textit{Remark 1:} This assumption places a demand on the quality of actions taken by the human expert after takeover. It reflects the likelihood of encountering an unsafe action, is restricted to a low threshold. This constraint ensures a high level of safety and reliability for the decisions made by the human expert.

\textbf{Assumption 2} (Expert-level human takeover). We assume that this expert-level human can accurately identify dangerous situations most of the time. In detail, the step-wise probability of that the human expert does not takeover when agents generates an unsafe action is bounded by a small value $\kappa<1$:
\begin{equation}
\int_{a \in \mathcal{A}}[1-I(s, a)] C^{g t}(s, a) d a=\int_{a \in \mathcal{A}_{human}(s)} C^{g t}(s, a) d a \leq \kappa
\end{equation}

\textit{Remark 2:} This assumption highlights the human expert's ability to effectively identify hazardous situations, setting a probabilistic limit on their likelihood of missing a required takeover. Moreover, these two assumptions do not place any constraints on the structure of the human expert's policy.

\textit{\textbf{Lemmas.}} We propose several useful lemmas and their corresponding proofs, which are employed in the theorem.

\textbf{Lemmas 1 } (The performance difference lemma). This lemma was introduced and proven by \cite{kakade2002approximately} and serves the purpose of demonstrating the safety of the mixed behavior policy:
\begin{equation}
V_{\pi_{mix}}=V_{\pi_{human}}+\frac{1}{1-\gamma} \underset{s \sim P_{\pi_{mix}}}{\mathbb{E}} \underset{a \sim \pi_{mix}}{\mathbb{E}}\left[A_{\pi_{human}}(s, a)\right]
\end{equation}
where, the $P_{{\pi_{human}}}$ refers the states are subject to the marginal state distribution derived from the mixed behavior policy ${\pi_{mix}}$. ${A_{\pi_{human}}}(s_t, a_t)$ represents the advantage of the human expert in the current state-action pair: ${A_{\pi_{human}}}(s_t, a_t) = C^{gt}(s_t, a_t) + \gamma V_{{\pi_{human}}}(s_{t+1}) - V_{{\pi_{human}}}(s_t)$ and $s_{t+1} = \mathcal P(s_t, a_t)$ is the next state. In the original proposition, the $V$ and $A$ represent the expected discounted return and advantage concerning the reward, respectively. However, we substitute the reward with the indicator $C^{gt}$ to make the value functions ${V_{\pi_{mix}}}$ and $V_{{\pi_{human}}}$ reflect the expected cumulative failure probability.

\textbf{Lemmas 2} The cumulative probability of failure of the expert, denoted as $V_{{\pi_{human}}}(s)$, is bounded for all state \citep{li2022efficient}:
\begin{equation}
V_{{\pi_{human}}}(s) \le \cfrac{\epsilon}{1 - \gamma}
\end{equation}

\textit{Proof.} Following Assumption 1:
\begin{equation}
V_{\pi_{human}}\left(s_t\right)=\underset{\pi_{human}}{\mathbb{E}}\left[\sum_{t^{\prime}=t}^{\infty} \gamma^{t^{\prime}-t} C^{g t}\left(s_{t^{\prime}}, a_{t^{\prime}}\right)\right]=\sum_{t^{\prime}=t}^{\infty} \gamma^{t^{\prime}-t} \underset{\pi_{human}}{\mathbb{E}}\left[C^{g t}\left(s_{t^{\prime}}, a_{t^{\prime}}\right)\right] \leq \sum_{t^{\prime}=t}^{\infty} \gamma^{t^{\prime}-t} \epsilon=\frac{\epsilon}{1-\gamma}
\end{equation}

\textit{\textbf{Theorem.}} We introduced the Theorem for HAIM’s explicit intervention mechanism, as represented in Eq. \ref{equation: main theorem}, in this work. It illustrates the relationship between training safety, the action error rate $\epsilon$ (related to Assumption 1), and the intervention error rate $\kappa$ (related to Assumption 2) of the human expert. The proof is provided below.

\textit{Proof.} 
We utilize Lemma 1 and 2 to establish the upper bound. To begin, we decompose the advantage by splitting the mixed behavior policy $\pi_{mix}$ as follows:
\begin{equation}
\begin{aligned}
& \underset{a \sim \pi_{mix}(\cdot \mid s)}{\mathbb{E}} A_{\pi_{human}}(s, a)=\int_{a \in \mathcal{A}} \pi_{mix}(a \mid s) A_{\pi_{human}}(s, a) \\
= & \int_{a \in \mathcal{A}}\left\{\pi_{AV}(a \mid s)(1-I(s, a)) A_{\pi_{human}}(s, a)+\pi_{human}(a \mid s) F(s) A_{\pi_{human}}(s, a)\right\} d a \\
= & \int_{a \in \mathcal{A}_{human}(s)}\left[\pi_{AV}(a \mid s) A_{\pi_{human}}(s, a)\right] d a+F(s) \underset{a \sim \pi_{human}}{\mathbb{E}}\left[A_{\pi_{human}}(s, a)\right] 
\end{aligned}
\end{equation}

The second term equals zero as per the definition of advantage. Consequently, we only need to calculate the first term. Let's expand the advantage into its detailed form:
\begin{equation}
\begin{aligned}
& \underset{a \sim \pi_{mix}(\cdot \mid s)}{\mathbb{E}} A_{\pi_{human}}(s, a)=\int_{a \in \mathcal{A}_{human}(s)}\left[\pi_{AV}(a \mid s) A_{\pi_{human}}(s, a)\right] d a \\
= & \int_{a \in \mathcal{A}_{human}(s)} \pi_{AV}(a \mid s)\left[C^{g t}(s, a)+\gamma V_{\pi_{human}}\left(s^{\prime}\right)-V_{\pi_{human}}(s)\right] d a \\
= & \underbrace{\int_{a \in \mathcal{A}_{human}(s)} \pi(a \mid s) C^{g t}(s, a) d a}_{\text {(a)}}+\underbrace{\gamma \int_{a \in \mathcal{A}_{human}(s)} \pi(a \mid s) V_{\pi_{human}}\left(s^{\prime}\right) d a}_{\text {(b) }}-\underbrace{\int_{a \in \mathcal{A}_{human}(s)} \pi(a \mid s) V_{\pi_{human}}(s) d a}_{\text {(c) }} 
\end{aligned}
\end{equation}

Based on Assumption 1, we can bound term (a) as follows:
\begin{equation}
\int_{a \in \mathcal{A}_{human}(s)} \pi(a \mid s) C^{g t}(s, a) d a \leq \int_{a \in \mathcal{A}_{human}(s)} C^{g t}(s, a) d a \leq \kappa .
\end{equation}

Utilizing Lemma 2, term (b) can be expressed as follows:
\begin{equation}
\gamma \int_{a \in \mathcal{A}_{human}(s)} \pi(a \mid s) V_{\pi_{human}}\left(s^{\prime}\right) d a \leq \gamma \int_{a \in \mathcal{A}_{human}(s)} V_{\pi_{human}}\left(s^{\prime}\right) d a \leq \frac{\gamma \epsilon}{1-\gamma} \int_{a \in \mathcal{A}_{human}(s)} d a=\frac{\gamma \epsilon}{1-\gamma} K(s)
\end{equation}
where $K(s)$ represents the area of the human-preferred region in the action space, denoted as $K(s)=\int_{a \in \mathcal{A}_{human}(s)} d a$. It is a function that depends on both the human expert and the current state.

The term (c) is always non-negative, so when we subtract term (c), the resulting negative term will always be less than or equal to zero.

By combining the upper bounds of the three terms, we obtain the bound on the advantage as follows:
\begin{equation}
\label{equation:bound-of-advantage}
\underset{a \sim \pi_{mix}}{\mathbb{E}} A_{\pi_{human}}(s, a) \leq \kappa+\frac{\gamma \epsilon}{1-\gamma} K(s)
\end{equation}

Now we put Eq.~\ref{equation:bound-of-advantage} and Lemma 2 into the performance difference lemma (Lemma 1), resulting in:
\begin{equation}
\begin{aligned}
V_{\pi_{mix}} & =V_{\pi_{human}}+\frac{1}{1-\gamma} \underset{s \sim P_{\pi_{mix}}}{\mathbb{E}} \underset{a \sim \pi_{mix}}{\mathbb{E}}\left[A_{\pi_{human}}(s, a)\right] \\
& \left.\leq \frac{\epsilon}{1-\gamma}+\frac{1}{1-\gamma}\left[\kappa+\frac{\gamma \epsilon}{1-\gamma} \max _s K(s)\right]\right] \\
& =\frac{1}{1-\gamma}\left[\epsilon+\kappa+\frac{\gamma \epsilon^2}{1-\gamma} K^{\prime}\right]
\end{aligned}
\end{equation}
where $K' = \max_s K(s) = \max_s \mathop{\int}_{a \in {\mathcal A_{human}}(s)}da \ge 0$  is associated with the tolerance level of the human expert. 

We have now successfully demonstrated the upper bound of the discounted probability of failure for the mixed behavior policy $\pi_{mix}$ in the HAIM's explicit intervention mechanism.

\section{Hyper-parameters}
\label{Appendix:Hyper-parameters}

\begin{table}[H]
\begin{minipage}{0.45\linewidth}
\centering
\caption{PPO/PPO-Lag}
\begin{tabular}{@{}ll@{}}
\toprule
Hyper-parameter             & Value  \\ \midrule
KL Coefficient              & 0.2    \\
$\lambda$ for GAE~\citep{schulman2015high} & 0.95 \\
Discounted Factor $\gamma$   & 0.99  \\
Number of SGD epochs   & 20     \\
Train Batch Size & 4000 \\
SGD mini-batch size & 100 \\
Learning Rate               & 0.00005 \\ 
Clip Parameter $\epsilon$ & 0.2 \\
\midrule
Cost Limit for PPO-Lag & 1 \\
\bottomrule
\end{tabular}
\end{minipage} \hfill
\begin{minipage}{0.45\linewidth}
\centering
\caption{SAC/SAC-Lag/CQL}
\begin{tabular}{@{}ll@{}}
\toprule
Hyper-parameter             & Value  \\ \midrule
Discounted Factor $\gamma$   & 0.99  \\
$\tau$ for target network update & 0.005 \\
Learning Rate               & 0.0001 \\ 
Environmental horizon $T$ & 1500 \\
Steps before Learning start & 10000\\
\midrule
Cost Limit for SAC-Lag & 1 \\
\midrule
BC iterations for CQL & 200000 \\
CQL Loss Temperature $\beta$ & 5 \\
Min Q Weight Multiplier & 0.2 \\
\bottomrule
\end{tabular}
\end{minipage}\hfill
\end{table}

\begin{table}[H]
\begin{minipage}{0.45\linewidth}
\centering
\caption{CPO}
\begin{tabular}{@{}ll@{}}
\toprule
Hyper-parameter             & Value  \\ \midrule
KL Coefficient              & 0.2    \\
$\lambda$ for GAE~\citep{schulman2015high} & 0.95 \\
Discounted Factor $\gamma$   & 0.99  \\
Number of SGD epochs   & 20     \\
Train Batch Size & 8000 \\
SGD mini-batch size & 100 \\
Learning Rate               & 0.00005 \\ 
Clip Parameter $\epsilon$ & 0.2 \\
\midrule
Cost Limit & 1 \\
\bottomrule
\end{tabular}
\end{minipage}\hfill
\begin{minipage}{0.45\linewidth}
\centering
\caption{BC}
\begin{tabular}{@{}ll@{}}
\toprule
Hyper-parameter             & Value  \\ \midrule
Dataset Size & 49,000 \\
SGD Batch Size & 32 \\
SGD Epoch & 200000 \\
Learning Rate & 0.0001\\
\bottomrule
\end{tabular}
\end{minipage}
\end{table}

\begin{table}[H]
\begin{minipage}{0.45\linewidth}
\centering
\caption{GAIL}
\begin{tabular}{@{}ll@{}}
\toprule
Hyper-parameter             & Value  \\ \midrule
Dataset Size & 49,000 \\
SGD Batch Size & 64 \\
Sample Batch Size &  12800 \\
Generator Learning Rate & 0.0001 \\
Discriminator Learning Rate & 0.005 \\
Generator Optimization Epoch & 5 \\
Discriminator Optimization Epoch & 2000 \\
Clip Parameter $\epsilon$ & 0.2 \\
\bottomrule
\end{tabular}
\end{minipage}\hfill
\begin{minipage}{0.45\linewidth}
\centering
\caption{HG-DAgger}
\begin{tabular}{@{}ll@{}}
\toprule
Hyper-parameter             & Value  \\ 
\midrule
Initializing dataset size & 30,000 \\
Number of data aggregation epoch & 4 \\
Interactions per round & 5000 \\
SGD batch size & 256\\
Learning rate & 0.0004\\
\bottomrule
\end{tabular}
\end{minipage}
\end{table}

\begin{table}[H]
\begin{minipage}{0.45\linewidth}
\centering
\caption{IWR}
\begin{tabular}{@{}ll@{}}
\toprule
Hyper-parameter             & Value  \\ 
\midrule
Initializing dataset size & 30,000 \\
Number of data aggregation epoch & 4 \\
Interactions per round & 5000 \\
SGD batch size & 256\\
Learning rate & 0.0004\\
Re-weight data distribution & True \\
\bottomrule
\end{tabular}
\end{minipage}\hfill
\begin{minipage}{0.45\linewidth}
\centering
\caption{HACO}
\begin{tabular}{@{}ll@{}}
\toprule
Hyper-parameter             & Value  \\ \midrule
Discounted Factor $\gamma$   & 0.99  \\
$\tau$ for Target Network Update & 0.005 \\
Learning Rate               & 0.0001 \\ 
Environmental Horizon $T$ & 1000 \\
Steps before Learning Start & 100\\
Steps per Iteration & 100 \\
Train Batch Size & 1024  \\
CQL Loss Temperature & 10.0 \\
Target Entropy & 2.0\\ 
\bottomrule
\end{tabular}
\end{minipage}
\end{table}

\begin{table}[H]
\begin{minipage}{0.45\linewidth}
\centering
\caption{HAIM-DRL}
\begin{tabular}{@{}ll@{}}
\toprule
Hyper-parameter             & Value  \\ \midrule
Number of \(HV\) & 5 \\
Total simulation time & 300 steps \\
Length of the road & 800 m \\
Initial headway distance \(d_{i,0}^{HV}\) & 15 m \\
Safe distance \(s_0\) & 5 m \\
Safe time headway & 1 s \\
Initial velocity of \(HV\) \(v_{i,0}^{HV}\) & 0 m/s \\
Maximum velocity of \(HV\) \(v^{HV*}\) & 15 m/s \\
Minimum velocity of \(HV\) & 0 m/s \\
Initial acceleration of \(HV\) \(acc_{i,0}^{HV}\) & 0 \(m/s^2\) \\
Maximum acceleration \(a_{max}\) & 4 \(m/s^2\) \\
Comfortable deceleration \(b\) & -4 \(m/s^2\) \\
Time step size \(\Delta t\) & 0.1 s \\
\midrule
AV agent’s action persist time $T$ &  10 s \\
Thresholds for calculating disturbance cost $\lambda$ &  -5 \(m/s^2\) \\
Weighting Factor $\psi$ & 1 \\
Entropy Regularization Coefficient $\alpha$ & 10 \\
Weighting Factor $\beta$ & 1 \\
Weighting Factor $\varphi$ & 1 \\
\midrule
Discounted Factor $\gamma$   & 0.99  \\
$\tau$ for Target Network Update & 0.005 \\
Learning Rate               & 0.0001 \\ 
Environmental Horizon & 1000 \\
Steps before Learning Start & 100\\
Steps per Iteration & 100 \\
Train Batch Size & 1024  \\
CQL Loss Temperature & 10.0 \\
Target Entropy & 2.0\\ 
\bottomrule
\end{tabular}
\end{minipage}
\end{table}

\newpage

\subsection*{  } 
\setlength\intextsep{0pt} 
\begin{wrapfigure}{l}{25mm}
    \centering
    \includegraphics[width=0.12\textwidth]{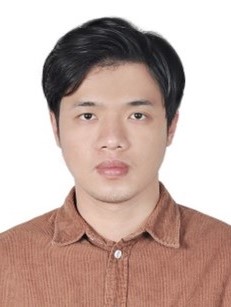}
\end{wrapfigure}
\noindent \textbf{Zilin Huang} received his B.S. degree from School of Electromechanical Engineering, Guangdong University of Technology in 2018. He received his M.S. degree in Communication and Transportation Engineering from South China University of Technology in 2021. He is currently pursuing a Ph.D. degree at the Department of Civil
and Environmental Engineering, University of Wisconsin-Madison, USA. Before joining UW-Madison, he worked at the Center for Connected and Automated Transportation (CCAT), Purdue University, USA. His research interests include human-centered AI, autonomous driving, robotics, human-AI interaction, intelligent transportation. \par

\hspace*{\fill} 

\subsection*{  } 
\setlength\intextsep{0pt} 
\begin{wrapfigure}{l}{25mm}
    \centering
    \includegraphics[width=0.12\textwidth]{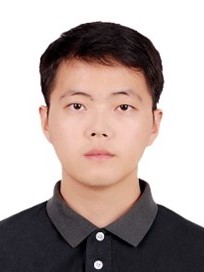}
\end{wrapfigure}
\noindent \textbf{Zihao Sheng} received his B.S. degree in Automation from Xi'an Jiaotong University, Xi’an, China in 2019 and M.S. degree in Control Engineering from Shanghai Jiao Tong University, Shanghai, China in 2022. He is currently pursuing a Ph.D. degree at the Department of Civil and Environmental Engineering, University of Wisconsin-Madison, USA. His main research interests include human-centered AI, autonomous driving, reinforcement learning, and intelligent transportation. \par

\hspace*{\fill} 

\subsection*{  } 
\setlength\intextsep{0pt} 
\begin{wrapfigure}{l}{25mm}
    \centering
    \includegraphics[width=0.12\textwidth]{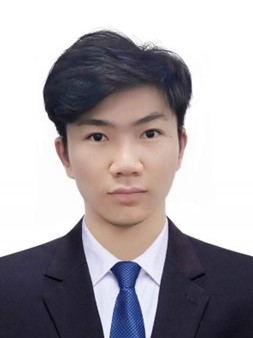}
\end{wrapfigure}
\noindent \textbf{Chengyuan Ma} received the B.S. and Ph.D. degrees in traffic engineering from Tongji University in 2017 and 2023, respectively. He is currently a Postdoctoral Research Associate with the Department of Civil and Environmental Engineering of the University of Wisconsin-Madison. His research interests include traffic control with connected autonomous vehicles and transportation big data analytics. \par

\hspace*{\fill} 

\subsection*{  } 
\setlength\intextsep{0pt} 
\begin{wrapfigure}{l}{25mm}
    \centering
    \includegraphics[width=0.12\textwidth]{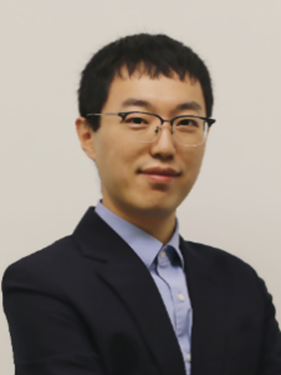}
\end{wrapfigure}
\noindent \textbf{Sikai (Sky) Chen} is an Assistant Professor at the Department of Civil and Environmental Engineering, University of Wisconsin-Madison. He received his Ph.D. in Civil Engineering with a focus on Computational Science \& Engineering from Purdue University. His research centers around three major themes: human users, AI, and transportation. He aims to innovate and develop safe, efficient, sustainable, and human-centered transportation systems using cutting-edge methods and technologies. The focus is on incorporating human behaviors, interactive autonomy, and intelligent control frameworks. In addition, he is a member of two ASCE national committees: Connected \& Autonomous Vehicle Impacts, and Economics \& Finance; and TRB Standing Committee on Statistical Methods (AED60). More information can be found at Sky-Lab: \href{https://sky-lab-uw.github.io/}{https://sky-lab-uw.github.io/}\par

\end{document}